\documentclass[a4paper,american,english,pdftex, a4paper, twoside, BCOR12mm, titlepage, openright, headsepline, chapterprefix, mpexclude,  idxtotoc, liststotoc, cleardoublestandard]{ANUthesisLyx}
\usepackage[latin9]{inputenc}
\usepackage{babel}
\usepackage{array}
\usepackage{float}
\usepackage{textcomp}
\usepackage{multirow}
\usepackage{amsthm}
\usepackage{amsmath}
\usepackage{graphicx}
\usepackage{esint}
\usepackage{hyperref}

\makeatletter

\pdfpageheight\paperheight
\pdfpagewidth\paperwidth

\providecommand{\tabularnewline}{\\}
\floatstyle{ruled}
\newfloat{algorithm}{tbp}{loa}[chapter]
\providecommand{\algorithmname}{Algorithm}
\floatname{algorithm}{\protect\algorithmname}

 \usepackage{amssymb}
 \usepackage{graphicx}
 \usepackage{url}
\numberwithin{equation}{section}
\numberwithin{figure}{section}

\usepackage{algpseudocode}
\usepackage{placeins}
\usepackage{colortbl}
\usepackage{euscript}

\DeclareSymbolFont{rsfscript}{OMS}{rsfs}{m}{n}
\DeclareSymbolFontAlphabet{\mathrsfs}{rsfscript}

\usepackage{titletoc}
\dottedcontents{figure}[4.5em]{}{3em}{0.75em}
\dottedcontents{table}[4.5em]{}{3em}{0.75em}

\usepackage{tikz}
\usetikzlibrary{shapes,decorations,shadows}
\usetikzlibrary{decorations.pathmorphing}
\usetikzlibrary{decorations.shapes}
\usetikzlibrary{fadings}
\usetikzlibrary{patterns}
\usetikzlibrary{calc}
\usetikzlibrary{decorations.text}
\usetikzlibrary{decorations.footprints}
\usetikzlibrary{decorations.fractals}
\usetikzlibrary{shapes.gates.logic.IEC}
\usetikzlibrary{shapes.gates.logic.US}
\usetikzlibrary{fit,chains}
\usetikzlibrary{positioning}
\usepgflibrary{shapes}
\usetikzlibrary{scopes}
\usetikzlibrary{arrows}
\usetikzlibrary{backgrounds}

\tikzset{latent/.style={circle,fill=white,draw=black,inner sep=1pt, 
minimum size=20pt, font=\fontsize{10}{10}\selectfont},
obs/.style={latent,fill=gray!25},
const/.style={rectangle, inner sep=0pt},
factor/.style={rectangle, fill=black,minimum size=5pt, inner sep=0pt},
>={triangle 45}}

\pgfdeclarelayer{b}
\pgfdeclarelayer{f}
\pgfsetlayers{b,main,f}

\newcommand{\plate}[4]{
\begin{pgfonlayer}{b}
\node (invis#1) [draw, color=white, inner sep=1pt,rectangle,fit=#2] {};
\end{pgfonlayer}\begin{pgfonlayer}{f}
\node (capt#1) [ below left=0 pt of invis#1.south east, xshift=2pt,yshift=1pt] {\footnotesize{#3}};
\node (#1) [draw,inner sep=1pt, rectangle,fit=(invis#1) (capt#1),#4] {};
\end{pgfonlayer}
}

\newcommand{\nofactor}[4]{
\node (#1) [factor, #2]  {};
\node (capt#1) [#4 of #1]{\footnotesize{#3}};
}

\newcommand{\gate}[3]{
\node (#1) [rectangle,draw,dashed, inner sep=2pt, fit=#2]{};
}

\makeatother

\addto\captionsamerican{\renewcommand{\algorithmname}{Algorithm}}

\begin{document}

\newtheorem{theorem}{Theorem}

\newcommand{\diff}[2]{\frac{\partial #1}{\partial #2}} 

\title{Multi-GPU Distributed Parallel Bayesian Differential Topic Modelling}
\author{Aaron(Qiaochu) Li}
\degree{Bachelor of Science(Advanced)(Honours)}
\department{Research School of Computer Science}
\chair{Doctor Wray Buntine} 
\numberofmembers{2}
\degreeyear{2012} 
\degreesemester{Semester 2}
\othermembers{Doctor Scott Sanner}
\maketitle
\copyrightpage
\abstract{}{
There is an explosion of data, documents, and other content, and people require tools to analyze and interpret these, tools to turn the content into information and knowledge. Topic modelling have been developed to solve these problems. Bayesian topic models such as Latent Dirichlet Allocation (LDA) \cite{bleinj03} allow salient patterns in large collection of documents to be extracted and analyzed automatically. When analyzing texts, these patterns are called topics, represented as a distribution of words. Although numerous extensions of LDA have been created in academia in the last decade to address many problems, few of them can reliablily analyze multiple groups of documents and extract the similarities and differences in topics across these groups. Recently, the introduction of techniques for differential topic modelling, namely the Shadow Poisson Dirichlet Process model (SPDP) \cite{chen2012spdp} performs uniformly better than many existing topic models in a discriminative setting.\par
\vspace{.1in}
There is also a need to improve the running speed of algorithms for topic models. While some effort has been made for distributed algorithms, there is no work currently done using graphical processing units (GPU). Note the GPU framework has already become the most cost-efficient and popular parallel platform for many research and industry problems. \par
\vspace{.1in}
In this thesis, I propose and implement a scalable multi-GPU distributed parallel framework which approximates SPDP, called MGPU-DP-SPDP, and a version running on a single GPU, Improved-GPU-SPDP. Through experiments, I have shown Improved-GPU-SPDP improved the running speed of SPDP by about 50 times while being almost as accurate as SPDP, with only one single cheap laptop GPU. Furthermore, I have shown the speed improvement of MGPU-DP-SPDP is sublinearly scalable when multiple GPUs are used, while keeping the accuracy fairly comparable to SPDP. Therefore, on a medium-sized GPU cluster, the speed improvement could potentially reach a factor of a thousand. \par
\vspace{.1in}
Note SPDP is just a representative of perhaps another hundred other extensions of LDA. Although my algorithm is implemented to work with SPDP, it is designed to be a general framework that can be extended to work with other LDA extensions and improve their speed, with only a small amount of modification.  The speed-up on smaller collections, typically gained as a result of an exploratory query to a search engine (i.e., 1000s of documents rather than 100,000s), means that these more complex LDA extensions could now be done in real-time, thus opening up a new way of using these LDA models in industry.} 
\begin{frontmatter}
\tableofcontents
\listoffigures
\listoftables
\acknowledgements{}{
I would like to thank Dr. Wray Buntine and Dr. Scott Sanner for their guidance and support throughout the year. The research is a journey exploring the unknown. Their advices and experiences have saved me many times through the journey when I was about to give up, and guided me to the correct path when I thought it was impossible to go ahead. In particular, I would like to thank them for sharing their knowledge for many things, encouraging me to challenge myself with things that others have never done before, inspiring me with their frontier research results, and spending a large amount of their time to help me refine this thesis.  \par
\vspace{.1in}
I would also like to thank Changyou Chen, who patiently explained to me the background knowledge of the field, and generously shared his expertise in differential topic models. Without Changyou's tremendous amount of work previously done on differential topic models, I would not be able to deliver the research with this many interesting results. \par
\vspace{.1in}
Finally, I would like to thank ANU and NICTA, who brought researchers together, created and maintained the environment for researchers to work on things they love to do.  }
\end{frontmatter}

\global\long\def\items{\EuScript{V}}

\global\long\def\tastes{\CMcal{T}}

\global\long\def\nodes{\EuScript{N}}

\global\long\def\edges{\EuScript{E}}

\global\long\def\content{\mathrsfs{C}}

\global\long\def\vocab{\CMcal{V}}

\global\long\def\docs{\CMcal{D}}

\global\long\def\cites{\CMcal{C}}

\global\long\def\up#1{\uppercase{#1}}

\clearpage{}

\chapter{Introduction}

\section{Overview}

\paragraph*{\textmd{There is an explosion of data, documents, and other content,
and people require tools to analyze and interpret these, tools to
turn the content into information and knowledge. Topic modelling is
a research area that has been developed for exploratory information
access, and can be applied to documents in particular to support the
task of understanding content. Bayesian topic models such as Latent
Dirichlet Allocation (LDA) \cite{bleinj03} allow salient patterns
in large collection of documents to be extracted and analyzed automatically.
When analyzing texts, these patterns are called topics, represented
as a distribution of words. }}

\paragraph*{\textmd{As research effort in topic models are getting slowly adapted
to industry practice, there is a need to improve the running speed
of the algorithms. In particular, in exploratory data analysis, it
is usually the case that topic modelling is done in an interactive
environment, where fast response is important. When data analysis
is provided as a service, it is also important to do the computation
cost-efficiently. While some effort has been made for distributed
algorithms, there is no work currently done using graphical processing
units (GPU). Note the GPU framework has already become the most cost-efficient
and popular parallel platform for many research and industry problems.}}

\paragraph*{\textmd{Although numerous extensions of LDA has been created in academia
in the last decade to address many problems, few of them can reliablily
analyze multiple groups of documents and extract the similarities
and differences in topics across these groups. This problem is seen
when businesses want to do comparative analysis, or when political
analysts want to understand opinions across different political groups.
Recently, the introduction of techniques for differential topic modelling,
namely the Shadow Poisson Dirichlet Process model (SPDP) \cite{chen2012spdp}
performs uniformly better than many existing topic models in a discriminative
setting.}}

\paragraph*{\textmd{In this thesis, I propose and implement a scalable multi-GPU
distributed parallel framework which approximates SPDP, called MGPU-DP-SPDP,
and a version running on a single GPU, Improved-GPU-SPDP. Through
experiments, I have shown Improved-GPU-SPDP improved the running speed
of SPDP by about 50 times, with only one single cheap laptop GPU.
Furthermore, I have shown the speed improvement of MGPU-DP-SPDP is
sublinearly scalable when multiple GPUs are used. Therefore, on a
medium-sized GPU cluster, the speed improvement could potentially
reach thousands of times. My experiments have shown when a single
GPU is used, Improved-GPU-SPDP is almost as accurate as SPDP, as measured
by perplexity and intepretability; when multiple GPUs are used, MGPU-DP-SPDP
is fairly comparable.}}

\paragraph*{\textmd{Note SPDP is a representative of perhaps another hundred
LDA extensions. Although my algorithm is implemented to work with
SPDP, it is designed to be a general framework that can be extended
to work with other LDA extensions and improve their speed, with only
a small amount of modification. The speed up on smaller collections,
typically gained as a result of an exploratory query to a search engine
(i.e., 1000s of documents rather than 100,000s), means that these
more complex LDA extensions could now be done in realtime, thus opening
up a new way of using these LDA models in industry.}}

\section{Introduction to Topic Modelling}

\paragraph*{\textmd{Today, the amount of information available to us is far greater
than our capacity to process the information. The explosion of information
has led to the rise of a new research area: Information Access. People
need tools to organize, search, summarize, and understand information,
tools to turn information into knowledge \cite{Wray/ALTA/Talk}. }}

\paragraph*{\textmd{Techniques such as topic modelling were invented to address
these issues. Topic models uncover the underlying patterns in a collection
of documents through analyzing the semantic content. Bayesian topic
models are a class of topic models that assume a document contains
multiple patterns to different extents, represented as Bayesian latent
variables. When analyzing text, these patterns are represented as
a distribution of words, called ``topics''. For example, (``Java''
0.25, ``C++'' 0.3, ``C'' 0.1, ``Python'' 0.1, ``computer''
0.15, ``science'' 0.1) can be interpreted as the topic ``programming
languages''. }}

\paragraph*{\textmd{At present, most search engines and document analysis tools
uses keywords and relationships between documents as fundamental metrics.
While these tools work reasonably well in searching for specific terms
and have been shown to be able to find popular documents with respect
to public opinion, they cannot explore the underlying patterns and
topics among these documents, that someone may want to do in an exploratory
analysis. By comparison, topic models provide insightful analysis
on the topics contained by documents through statistical methods,
represented in probability distributions over topics and words. Today,
there are many proposed applications of topic modelling to different
industries: }}
\begin{itemize}
\item Finance, media, governmental: Public sentiment analysis \cite{Lin:2009:JSM:1645953.1646003}
\item Social network companies: Content based social network recommendation
systems \cite{conf/itng/LeeCM11,Pennacchiotti:2011:ITM:1963192.1963244}
\item Media companies: Trend analysis\cite{conf/wsdm/Kawamae11}, traditional
media bias detection \cite{conf/www/YounusQKSTOP12}
\item Military: Differential discovery on text collections \cite{chen2012spdp},
author detection\cite{Rosen-Zvi:2004:AMA:1036843.1036902}
\item Enterprises, consumers: Search engine and document processing \cite{journals/ws/NewmanBCHKMSZ10}
\item And many others...
\end{itemize}

\paragraph*{\textmd{Here is an intuitive explanation of how Bayesian topic models
work: It is a known fact that a human can quickly skim through a document,
summarize the main topics, and provide a few words to describe each
topic. Humans achieve this by memorizing words appeared in this document,
and implicitly compare the relative frequency of words that have appeared
so far. This process that can be imitated by computers with a few
conditions: The input data, which is a collection of documents, only
contain documents that are short enough for humans to skim through,
but also long enough to contain multiple topics. The process is begun
with feeding integer labeled words and documents to our program, which
contains one topic model. Topic models make a few assumptions on how
topics and words are generated by humans, make guesses on the underlying
topics, then observe and count the words being fed to the program.
Based on the observations, topic models adjust initial guesses on
underlying topics, then make a more accurate estimate. The process
re-iterates, until the topic models determine that the estimate of
underlying topics are accurate enough. Then, the result is read out
and presented for human to analyze.}}

\section{Introduction to Graphical Processing Units (GPUs)}

\paragraph*{\textmd{A few years ago the graphical processing unit (GPU) was only
considered a dedicated device to render images, or to convert digital
images to analog signals for monitors. Most of them are used in high-end
personal computers for gaming, by film companies to create animations
and special effects, or by large organizations to visualize their
data. Through the last few years, people have discovered the potential
computing power of GPUs. Pioneers have developed programming frameworks
that allow programmers to transfer computing instructions to GPU,
to utilize the huge arithmetic computing power inside GPU that hasn't
been properly exploited before (\cite{GPUcomp}). }}

\paragraph*{\textmd{These days the GPU has already become the one of the most
adopted parallel computing devices for many research and industry
problems due to its superior performance, cost-efficiency, and enegery-efficiency
for massive parallelism. According to the top 500 supercomputer list
in June 2012 \cite{Top500SuperComputer}, governments and private
insititutions have already invested a massive amount of resources
to create supercomputers with multi-GPU architecture. As of today,
many companies and public research organizations have specialized
teams in high performance computing dedicated to develop massive GPU
parallel algorithms for their existing applications. }}

\paragraph*{\textmd{Leaders of many industries have started to favor GPU computing
as opposed to old-school CPU computing. In the financial industry,
industry leaders such as Goldman Sachs and Morgan Stanley invested
large amount of money into building GPU-based computing infrastructures,
to run simulations of portofolio, financial market, and many other
applications. In the research project Square-Kilometre-Array project
ran by CSIRO \cite{GPU_SKA}, Australia, GPU is the crucial element
to get peta-bytes of data per second processed. Millions of developers
and computer technology hobbists around the world are also involved
in GPU computing. Bitcoin is the world-first Peer-to-Peer decentralized
virtual currency \cite{Bitcoin}, with more than 10 million US dollar
equivalent of transactions being processed every month. The network
is secured by its users running cryptography algorithm (SHA256) on
the network. Before 2011, the mainstream is to run the cryptography
algorithm on CPU. The trend shifted completely in a short 2 months
after the first GPU version encryption algorithm was developed. Today,
among millions of Bitcoin users, almost no one runs the cryptography
algorithm on the CPU anymore. }}

\section{Motivation}

\paragraph*{\textmd{Since LDA was published in the last decade, hundreds of extensions
have been made for many purposes, appearing in conferences such as
ICML, NIPS, KDD, SIGIR, ICCV, and others. Many use sophisticated non-parametric
statistics, and can be considerably slower than standard LDA. The
problem is, the algorithms are all too slow in practice. When analyzing
millions of documents, supercomputers are needed to get the result
in a reasonable amount of time. When analyzing a small collection
of documents, the running speed and the cost efficiency can still
make a decisive distinction in many real world situations. For example,
when the analysis tool is provided as a service, it is not feasible
to use an expensive computing resource. Furthermore, people prefer
to get the result as quickly as possible, rather than wait in a queue
for the analysis to be scheduled on supercomputers then wait for hours
or days to get the result back. When the analysis is done in an interactive
environment, people expect to get fast response and immediate feedback.}}

\paragraph*{\textmd{Because of these cost-efficiency and running speed issues,
many research results introduced at the beginning of this chapter
cannot be feasibly implemented and used in the real world. Researchers
and industries need solutions that are fast, scalable, cost-effective,
and generalizable.}}

\paragraph*{\textmd{An example of an LDA extension is SPDP (introduced in \ref{sub:Differential-Topic-Modelling}),
designed for differential text analysis. SPDP is a complex extension
of LDA which has longer running time than most LDA extensions. In
this thesis, I intend to use SPDP as a representative to implement
and find a distributed parallel approximation framework, which should
be both conceptually applied to another hundred LDA extensions, and
practically implemented with small amount of modification. To address
the cost-efficiency issue, I focus on the modern multi-GPU architecture,
which has been increasingly popular among both industries and researchers
in the last few years. My goal is to find a way to significantly improve
the running speed and cost-efficiency of SPDP, as a representative
of other extensions of LDA, under contemporary hardware architecture
and framework, so that research results can get truly applied and
adopted in industry and the real world, and provide a more efficient
tool for researchers.}}

\chapter{Background}

\section{Scope}

\paragraph*{\textmd{We assume the reader has graduate level background knowledge
in general areas of computer science, statistics, and some related
mathematics. In addition, to limit the length of our background chapter,
we expect the reader to be familiar with standard machine learning,
computer architecture, distributed and parallel computing, micro-processors,
Bayesian statistics, parameter estimation, statistical inference,
and Bayesian graphical models (see \cite{Bishop06}). The Wikipedia,
for instance, gives a good coverage of these areas. Moreover, we expect
familiarities with basic topic models such as Latent Dirichlet Allocation
(LDA) (see \cite{bleinj03,oai:fraunhofer.de:N-101883}) and common
performance measure such as perplexity (see section \ref{sub:Perplexity}),
as we only provide short explanations for these.}}

\paragraph*{\textmd{While non-Bayesian topic models such as Probabilistic Latent
Semantic Analysis (PLSA) \cite{DBLP:conf/uai/Hofmann99} do exist,
the focus of the topic modelling research field had been mostly shifted
to Bayesian topic models since the advent of Latent Dirichlet Allocation
(LDA) given their superior theoretical basis and good performance.
Therefore, we restrict the scope of our thesis to Bayesian topic models
only, and use LDA as a starting point for discussion. }}

\section{Notation}

\paragraph*{\textmd{Unless otherwise explicitly stated, all random variables
are discrete variables, all variables are non-negative real numbers
or integers, and all probability distributions are discrete distributions.
All Bayesian graphical models such as figure \ref{fig:LDA-graphical-model}
use plate notation, where each rectangle represent repeating entities,
number of repetition and range of repeating variables are specified
in the bottom right corner.}}

\section{Existing Bayesian Topic Models}

\paragraph*{\textmd{In this section I will give a brief overview of some existing
models which are relevant to my research. Technical details such as
model derivation, technical definitions, effects of hyper-parameter,
predictive probability, and inference algorithm will be left to the
next section. }}

\paragraph*{\textmd{A mathematical definition and a conceptional description
are given on the following models, ordered by their simplicity and
the dates they are created:}}
\begin{itemize}
\item Latent Dirichlet Allocation (LDA) \cite{bleinj03}
\item Pitman-Yor Topic Modelling (PYTM) \cite{conf/kdd/SatoN10}
\item Hierarchical Pitman-Yor Topic Modelling (HPYTM) \cite{conf/kdd/SatoN10}
\item Differential Topic Modelling using Shadow Poisson Dirichlet Process
(SPDP) \cite{chen2012spdp}
\end{itemize}

\paragraph*{\textmd{Most topic models are unigram (i.e 1-gram) models - they
only keep the number of occurrence of words appeared in documents,
and completely ignore the order of words. In other areas of linguistic
research, N-gram models are more popular than unigram models. They
consider every N consecutive words as a block while ignore the order
of blocks. }}

\paragraph*{\textmd{The reason that n-gram models are not commonly used in topic
model research is, although n-gram topic models contains richer semantic
information, it is often very difficult to find a large enough collection
of documents that contain multiple occurrence of each n-gram block,
hence posing severe difficulty to allow the topic model to extract
any statistically meaningful information. On the other hand, most
topic modelling algorithms have running time complexity growing linearly
or quadratically with the size of vocabulary. As n-gram models extend
the vocabulary size to $n$-th power of original vocabulary size,
the efficiency of topic modelling with n-gram becomes an efficiency
issue. }}

\paragraph*{\textmd{All models introduced in this section are unigram models.
Semantic structures contained in words and sentence are completely
ignored.}}

\FloatBarrier

\subsection{Latent Dirichlet Allocation (LDA)}

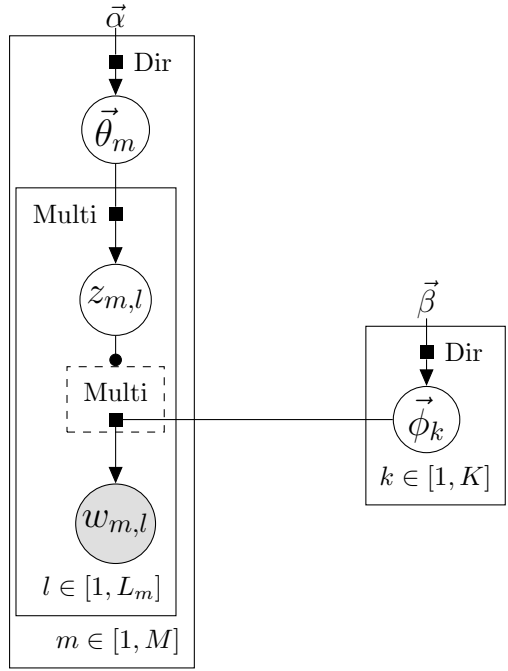
\begin{figure}
\selectlanguage{english}%
\begin{centering}
\begin{tikzpicture}
[latent/.style={circle,fill=white,draw=black,inner sep=1pt, minimum size=20pt, font=\fontsize{14}{14}\selectfont},
obs/.style={latent,fill=gray!25},
const/.style={rectangle, inner sep=0pt}]
\matrix[row sep=4mm,column sep=10mm, matrix anchor=mid] (lda) {
&
\\
&
\\
\node (theta) [latent] {$\vec{\theta}_{m}$}; 
&
\\
\nofactor{dz}{}{Multi}{left=0pt};
&
\\
\node (z) [latent]  {$z_{m,l}$}; 
&
\\
\nofactor{dw}{}{Multi}{above=0pt}; \gate{gw}{(dw) (captdw)}{$t$};
& & &  \node (phi) [latent]  {$\vec{\phi}_k$};
\\
\node (w) [obs]  {$w_{m,l}$}; 
&
\\
};
\nofactor{dtheta}{above=10pt of theta}{Dir}{right=0pt};
\nofactor{dphi}{above=10pt of phi}{Dir}{right=0pt};
\node (alpha) [const, above=10pt of dtheta]  {$\vec{\alpha}$};
\node (beta) [const, above=10pt of dphi]  {$\vec{\beta}$};
\draw [->] (beta) -- (dphi) -- (phi);
\draw [->] (alpha) -- (dtheta) -- (theta);
\draw [->] (theta) -- (dz) -- (z);
\draw [->] (phi) -- (dw) -- (w);
\draw [-*] (z) -- (gw);
\plate{wordplate}{(z)(w)(dz)(dw)(captdz)(captdw)(gw)}{$l \in{[1,L_m]}$}{}
\plate{docplate}{(wordplate)(theta)(dtheta)(captdtheta)}{$m \in{[1,M]}$}{}
\plate{topicplate}{(phi)(dphi)(captdphi)}{$k\in{[1,K]}$}{}
\end{tikzpicture}
\par\end{centering}

\selectlanguage{american}%
\protect\caption{\selectlanguage{english}%
LDA graphical model\label{fig:LDA-graphical-model}\selectlanguage{american}%
}
\end{figure}

\paragraph*{\textmd{Latent Dirichlet Allocation (LDA) \cite{bleinj03} is one
of the most popular models used in topic modelling because of its
simplicity. The graphical model is shown in Figure \ref{fig:LDA-graphical-model}.
The generation process is as follows.}}

\paragraph*{
\begin{alignat*}{2}
\vec{\phi}_{k} & \sim Dirichlet(\vec{\beta}) & \forall k=1...K\protect\\
\vec{\theta}_{m} & \sim Dirichlet(\vec{\alpha}) & \forall m=1...M\protect\\
z_{m,l} & \sim Multi(\vec{\theta}_{m}) & \forall m=1...M,\,l=1...L_{m}\protect\\
w_{m,l} & \sim Multi(\vec{\phi}_{z_{m,l}}) & \forall m=1...M,\,l=1...L_{m}
\end{alignat*}
\textmd{Here $M$ be the total number of documents, $L_{m}$ is total
number of words in document $m$, subscript $m$ denotes a document
index, $k$ denotes a topic index. The roles of other variables are:}}
\begin{itemize}
\item $\vec{\theta}_{m}$ : Topic proportion, a vector used as parameter
of multinomial distribution for document $m$. It determines the likelihood
of each topic in document $m$.
\item $\vec{\alpha}$: A constant hyper-parameter vector of dimension equal
to number of topics. It serves as a prior for the Dirichlet distribution
to determine the likelihood of topic proportions to be generated for
document $m$.
\item $\vec{\phi_{k}}$: Mixture component, a vector used as parameter of
multinomial distribution for topic $t$. It determines the likelihood
of each word in topic $k$.
\item $\vec{\beta}$: A constant hyper-parameter vector of dimension equal
to size of vocabulary, similar to $\vec{\alpha}$. It serves as a
prior for the Dirichlet distribution to determine the likelihood of
mixture components to be generated for topic.
\end{itemize}

\paragraph*{\textmd{Intuitively, the process is as follows: First, $\vec{\theta}_{m}$
and $\vec{\phi_{k}}$ are randomly generated for document $m$ and
topic $k$ according to parameters $\vec{\alpha}$ and $\vec{\beta}$.
Then, for each word index (range from 1 to $L_{m}$) in each document
$m$, a topic $z_{m,l}=k$ is drawn from a bag of topics (represented
as integers from 1 to $K$, where $K$ is the number of topics), where
the probability of each topic corresponds to value of each dimension
in $\vec{\theta}_{m}$. After the topic $k$ is drawn, the word $w_{m,l}$
is drawn from the vocabulary, where the probability for each word
corresponds to each dimension in $\vec{\phi_{k}}$.}}

\selectlanguage{american}%

\selectlanguage{english}%
\FloatBarrier

\subsection{Pitman-Yor Topic Modelling (PYTM)}

\begin{figure}
\begin{centering}
\includegraphics[width=1\textwidth]{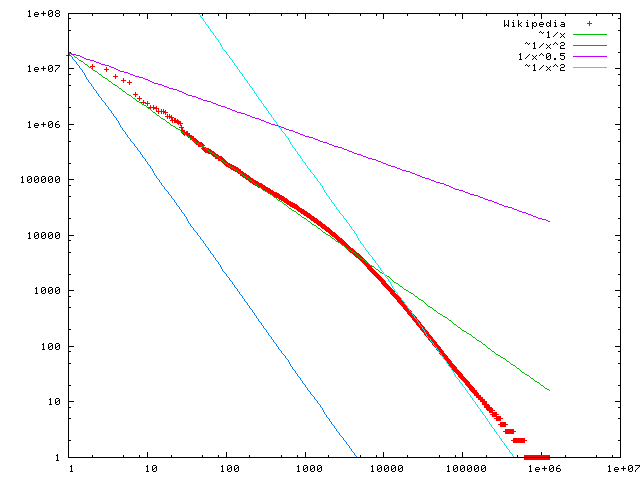}
\par\end{centering}

\protect\caption{Word frequency (y-axis in log scale) v.s Word frequency ranking (x-axis
in log scale) in Wikipedia (extracted from Wikipedia \cite{Wikipedia/ZipfsLaw}
under LGPL license)\label{fig:Word-frequency-v.s}}
\end{figure}

\paragraph*{\textmd{Pitman-Yor Topic Modelling (PYTM) \cite{conf/kdd/SatoN10}
made a few improvements over LDA, hence achieved significantly better
performance when measured in perplexity. The PYTM assumes in each
document the words are sequentially drawn from a distribution generated
by a Poisson Dirichlet Process (PDP)\cite{journals/corr/abs-1007-0296}
(also known as Pitman-Yor Process\cite{ishwaran01}). Topics are not
drawn before words - instead, the generation processes of a word and
a topic are mixed together. The PDP ensures words generated in each
document follow the properties of a power-law, which states words
already appearing before are exponentially more likely to appear again.
The justification is based on Zipf's law developed in linguistics
research, which states that in large text corpora, the number of times
a word appear in the corpora is approximately inversely proportional
to its rank. Figure \ref{fig:Word-frequency-v.s} shows the plot of
word frequency in Wikipedia as of 27 November, 2008 \cite{Wikipedia/ZipfsLaw}
. Each line represents a trend line fit to Zipf's law with some parameters. }}

\paragraph*{\textmd{Because the clustering effect in the PDP different from LDA,
the number of unique words and rare words generated by PDP in each
document is significantly higher than the results from LDA. In contrast
LDA only generates i.i.d words from a multinomial distribution, hence
is unable to capture the properties of a power-law \cite{conf/kdd/SatoN10}.}}

\paragraph*{\textmd{Figure \ref{fig:PYTM-graphical-model} shows the graphical
model of PYTM. Figure \ref{fig:PYTM-graphical-model} shows a more
precise illustration that breaks down each PDP generation step. The
generation process of PYTM is given as follows. In LDA we generate
$z_{m,l}$ and $w_{m,l}$ according to $H(z,w|\vec{\theta}_{m},\Phi)$,
which is a bivariate distribution equivalent to $z\sim\vec{\theta}_{m}$,
$w\sim\vec{\phi}_{z}$, and $\Phi$ is the matrix representation of
the collection of $\vec{\phi}_{k}$ for all $k=1...K$. In PYTM we
modify the process and take a variant of $H(z,w|\vec{\theta}_{m},\Phi)$,
then sample $z_{m,l}$ and $w_{m,l}$ from the variant.}}

\paragraph*{\textmd{
\begin{alignat*}{2}
\vec{\phi}_{k} & \sim Dirichlet(\vec{\beta}) & \forall k=1...K\protect\\
\vec{\theta}_{m} & \sim Dirichlet(\vec{\alpha}) & \forall m=1...M\protect\\
G_{m} & \sim PDP(\gamma,d,H(\cdot,\cdot|\vec{\theta}_{m},\Phi)) & \forall m=1...M\protect\\
(z_{m,l},w_{m,l}) & \sim G_{m} & \forall m=1...M,\,l=1...L_{m}
\end{alignat*}
}}

\paragraph*{\textmd{Note that as $\gamma$ approaches infinity, $G_{m}$ approaches
$H$, so the process collapsed into regular LDA. }}

\paragraph*{\textmd{Below is another way of representing this process, similar
to what is introduced in \cite{conf/kdd/SatoN10}. It breaks down
the sampling part to Chinese Restaurant Process. For each word $w_{m,l}$
with index $l$ in each document $m$:}\protect \\
\textmd{
\begin{alignat*}{1}
r_{m,l} & \sim Bernoulli(\ensuremath{\frac{\gamma+dJ_{m}}{\gamma+l}})\protect\\
x_{m,l} & =\protect\begin{cases}
\mbox{\ensuremath{x_{m,s}}}\mbox{for some \ensuremath{1\le s\le l-1}, each with probability \ensuremath{\frac{n_{m,s}-\gamma}{d+l-1}}} & (r_{m,l}=0)\protect\\
J_{m}+1 & (r_{m,l}=1)
\protect\end{cases}\protect\\
z_{m,l} & =\protect\begin{cases}
z_{m,s} & (r_{m,l}=0)\protect\\
\mbox{Draw from }Multi(\vec{\theta}_{m}) & (r_{m,l}=1)
\protect\end{cases}\protect\\
v_{m,x_{m,l}} & =\protect\begin{cases}
\mbox{unchanged} & (r_{m,l}=0)\protect\\
\mbox{Draw from }Discrete(\vec{\phi}_{z_{m,l}}) & (r_{m,l}=1)
\protect\end{cases}\protect\\
w_{m,l} & =v_{m,x_{m,l}}
\end{alignat*}
where: $\gamma$ is the concentration parameter of PDP, $d$ is the
discount parameter of PDP. $K$ is total number of topics, $M$ is
total number of documents. $J_{m}$ is number of distinct words so
far in document $m$. Through out the process there are two types
of generated words: either (1) when $r_{m,l}=0$, reuse a topic $k$
or word already generated from $\vec{\theta}_{m}$ or $\vec{\phi}_{k}$
before, or (2) when $r_{m,l}=1$, a new topic $z_{m,l}$ and a new
word $w_{m,l}$ is generated from $\vec{\theta}_{m}$ and $\vec{\phi}_{k}$
directly (which can be same as a topic or word generated before, or
completely new). $x_{m,l}$ and $v_{m,x_{m,l}}$ are defined as: If
we put type (2) words into a sequence $\{v_{x_{m,l}}\}_{m}$ for each
document $m$, and label them consecutively, $x_{m,l}$ is then the
label for each word $w_{m,l}$. Since type (1) word only reuses word
that already appeared before, they share the same label with same
type (2) word. New topic $z_{m,l}$ is only generated for type (2)
word.}}

\begin{figure}
\begin{centering}
\begin{tikzpicture}
[latent/.style={circle,fill=white,draw=black,inner sep=1pt, minimum size=20pt, font=\fontsize{14}{14}\selectfont},
obs/.style={latent,fill=gray!25},
const/.style={rectangle, inner sep=0pt}]
\matrix[row sep=4mm,column sep=10mm, matrix anchor=mid] (PYTM2) {
&
\\
&
\\
\node (theta) [latent] {$\vec{\theta}_{m}$}; 
&
\\
\nofactor{dz}{}{$PY(\gamma ,d ,H(\cdot,\cdot|\vec{\theta}_{m},\Phi))$}{left=0pt};
& & \node (phi) [latent]  {$\vec{\phi}_k$};
\\
\node(Gm)[latent]{$G_{m}$}; 

\\
\node (z) [latent]  {$z_{m,l}$}; 

\\
\node (w) [obs]  {$w_{m,l}$}; 
&
\\
};
\nofactor{dtheta}{above=10pt of theta}{Dir}{right=0pt};
\nofactor{dphi}{above=10pt of phi}{Dir}{right=0pt};
\node (alpha) [const, above=10pt of dtheta]  {$\vec{\alpha}$};
\node (beta) [const, above=10pt of dphi]  {$\vec{\beta}$};
\draw [->] (beta) -- (dphi) -- (phi);
\draw [->] (alpha) -- (dtheta) -- (theta);
\draw [->] (theta) -- (dz) -- (Gm) -- (z);
\draw [->] (phi) -- (dz);
\draw [-*] (z) -- (w);
\plate{wordplate}{(z)(w)}{$l \in{[1,L_m]}$}{}
\plate{docplate}{(wordplate)(theta)(dtheta)(Gm)}{$m \in{[1,M]}$}{}
\plate{topicplate}{(phi)(dphi)}{$k\in{[1,K]}$}{}
\end{tikzpicture}
\par\end{centering}

\protect\caption{PYTM graphical model \label{fig:PYTM-graphical-model}}

\end{figure}

\selectlanguage{american}%
\begin{figure}
\selectlanguage{english}%
\begin{centering}
\begin{tikzpicture}
[latent/.style={circle,fill=white,draw=black,inner sep=1pt, minimum size=20pt, font=\fontsize{14}{14}\selectfont},
obs/.style={latent,fill=gray!25},
const/.style={rectangle, inner sep=0pt}]
\matrix[row sep=4mm,column sep=10mm, matrix anchor=mid] (PYTM) {
&
\\
&
\\
&
\node (theta) [latent] {$\vec{\theta}_{m}$}; 
&
\\
&
\nofactor{dz}{}{Multi}{left=0pt};
&
\\
\node(x)[latent]{$x_{m,l}$};
&
\node (z) [latent]  {$z_{m,j}$}; 
&
\\
\nofactor{dw}{}{Table-Lookup}{above=0pt}; \gate{gw}{(dw)(captdw)}{$w$};
&
\nofactor{dv}{}{Discrete}{above=0pt}; \gate{gv}{(dv)(captdv)}{$v$};
& & &  
\node (phi) [latent]  {$\vec{\phi}_k$};
\\
\node (w) [obs]  {$w_{m,l}$}; 
&
\node (v) [latent]  {$v_{m,j}$}; 
&
\\
};
\nofactor{dtheta}{above=10pt of theta}{Dir}{right=0pt};
\nofactor{dphi}{above=10pt of phi}{Dir}{right=0pt};
\nofactor{dx}{above left=15pt of x}{PDP-Table-Draw}{right=0pt};
\node (gamma) [const, above left=30pt of dx] {$\gamma$};
\node (dc) [const, below left=30pt of dx] {$d$};
\node (alpha) [const, above=10pt of dtheta]  {$\vec{\alpha}$};
\node (beta) [const, above=10pt of dphi]  {$\vec{\beta}$};
\draw [->] (beta) -- (dphi) -- (phi);
\draw [->] (alpha) -- (dtheta) -- (theta);
\draw [->] (gamma) -- (dx) -- (x);;
\draw [->] (dc) -- (dx) -- (x);;
\draw [->] (theta) -- (dz) -- (z);
\draw [->] (phi) -- (dv) -- (v);
\draw [-*] (z) -- (gv);
\draw [-*] (x) -- (gw);
\draw [->] (dw) -- (w);
\draw [->] (v) -- (dw);
\plate{wordplate}{(w)(x)(dw)(dx)(captdw)(gw)}{$l \in{[1,L_{m}]}$}{}
\plate{tableplate}{(z)(v)(dz)(dv)(captdv)(gv)}{$j \in{[1,J_{m}]}$}{}
\plate{docplate}{(tableplate)(wordplate)(theta)(dtheta)(captdtheta)}{$m \in{[1,M]}$}{}
\plate{topicplate}{(phi)(dphi)(captdphi)}{$k\in{[1,K]}$}{}
\end{tikzpicture}
\par\end{centering}

\selectlanguage{american}%
\protect\caption{\selectlanguage{english}%
PYTM graphical model breakdown\label{fig:PYTM-graphical-model-breakdown}\selectlanguage{american}%
}
\end{figure}

\selectlanguage{english}%

\paragraph*{\textmd{The generation process of PYTM can also be described by a
much simpler Chinese Restaurant Process (CRP) analogy of Poisson Dirichlet
Process. We will give an overview of Poisson Dirichlet Process and
the CRP analogy in the next section.}}

\FloatBarrier

\subsection{Hierarchical Pitman-Yor Topic Modelling (HPYTM) }

\begin{figure}
\selectlanguage{english}%
\begin{centering}
\begin{tikzpicture}
[latent/.style={circle,fill=white,draw=black,inner sep=1pt, minimum size=20pt, font=\fontsize{14}{14}\selectfont},
obs/.style={latent,fill=gray!25},
const/.style={rectangle, inner sep=0pt}]
\matrix[row sep=4mm,column sep=10mm, matrix anchor=mid] (HPYTM) {
&
\\
&
\\
&
\node (theta) [latent] {$\vec{\theta}_{m}$}; 
&
\\
&
\nofactor{dz}{}{Multi}{left=0pt};
&
\\
\node(x)[latent]{$x_{m,l}$};
&
\node (z) [latent]  {$z_{m,j}$}; 
&
\node (topicgammak) [latent] {$\gamma_{k}$};
&
\node (topicdck)[latent]{$d_{k}$};
&
\node(phi0)[latent]{$\phi_{0}$};
\\
\nofactor{dw}{}{Table-Lookup}{above=0pt}; \gate{gw}{(dw)(captdw)}{$w$};
&
\nofactor{dv}{}{Discrete}{above=0pt}; \gate{gv}{(dv)(captdv)}{$v$};
& &  
\node (phik) [latent]  {$\vec{\phi}_{k}$};
\\
\node (w) [obs]  {$w_{m,l}$}; 
&
\node (v) [latent]  {$v_{m,j}$}; 
&
\\
};
\nofactor{dtheta}{above=10pt of theta}{Dir}{right=0pt};
\nofactor{dphik}{above=10pt of phik}{PDP}{right=0pt};
\nofactor{dtopicgammak}{above=10pt of topicgammak}{Gamma}{right=0pt};
\nofactor{dtopicdck}{above=10pt of topicdck}{Beta}{right=0pt};
\nofactor{dphi0}{above=10pt of phi0}{PDP}{right=0pt};
\nofactor{dx}{above left=15pt of x}{PDP-Table-Draw}{right=0pt};
\node (gamma) [const, above left=30pt of dx] {$\gamma$};
\node (dc) [const, below left=30pt of dx] {$d$};
\node (atopicgammak) [const, above left=10pt of dtopicgammak] {$a_{\gamma}$};
\node (btopicgammak) [const, above right=10pt of dtopicgammak] {$b_{\gamma}$};
\node (atopicdck) [const, above left=10pt of dtopicdck] {$a_{d}$};
\node (btopicdck) [const, above right=10pt of dtopicdck] {$b_{d}$};
\node (topicgamma0) [const, above left=10pt of dphi0] {$\gamma_{0}$};
\node (topicdc0) [const, above=10pt of dphi0] {$d_{0}$};
\node (U) [const, above right=10pt of dphi0] {$U$};
\node (alpha) [const, above=10pt of dtheta]  {$\vec{\alpha}$};
\draw [->] (topicdc0) -- (dphi0) -- (phi0);
\draw [->] (topicgamma0) -- (dphi0) -- (phi0);
\draw [->] (U) -- (dphi0) -- (phi0);
\draw [->] (atopicgammak) -- (dtopicgammak) -- (topicgammak);
\draw [->] (btopicgammak) -- (dtopicgammak) -- (topicgammak);
\draw [->] (atopicdck) -- (dtopicdck) -- (topicdck);
\draw [->] (btopicdck) -- (dtopicdck) -- (topicdck);
\draw [->] (phi0) -- (dphik);
\draw [->] (topicgammak) -- (dphik);
\draw [->] (topicdck) -- (dphik);
\draw [->] (dphik) -- (phik);
\draw [->] (alpha) -- (dtheta) -- (theta);
\draw [->] (gamma) -- (dx) -- (x);
\draw [->] (dc) -- (dx) -- (x);
\draw [->] (theta) -- (dz) -- (z);
\draw [->] (phik) -- (dv) -- (v);
\draw [-*] (z) -- (gv);
\draw [-*] (x) -- (gw);
\draw [->] (v) -- (dw) -- (w);
\plate{wordplate}{(w)(x)(dw)(dx)(captdw)(gw)}{$l \in{[1,L_{m}]}$}{}
\plate{tableplate}{(z)(v)(dz)(dv)(captdv)(gv)}{$j \in{[1,J_{m}]}$}{}
\plate{docplate}{(tableplate)(wordplate)(theta)(dtheta)(captdtheta)}{$m \in{[1,M]}$}{}
\plate{topicplate}{(phik)(dphik)(topicgammak)(topicdck)(dtopicgammak)(dtopicdck)}{$k\in{[1,K]}$}{}
\end{tikzpicture}
\par\end{centering}

\selectlanguage{american}%
\protect\caption{\selectlanguage{english}%
HPYTM graphical model breakdown\label{fig:HPYTM-breakdown}\selectlanguage{american}%
}
\end{figure}
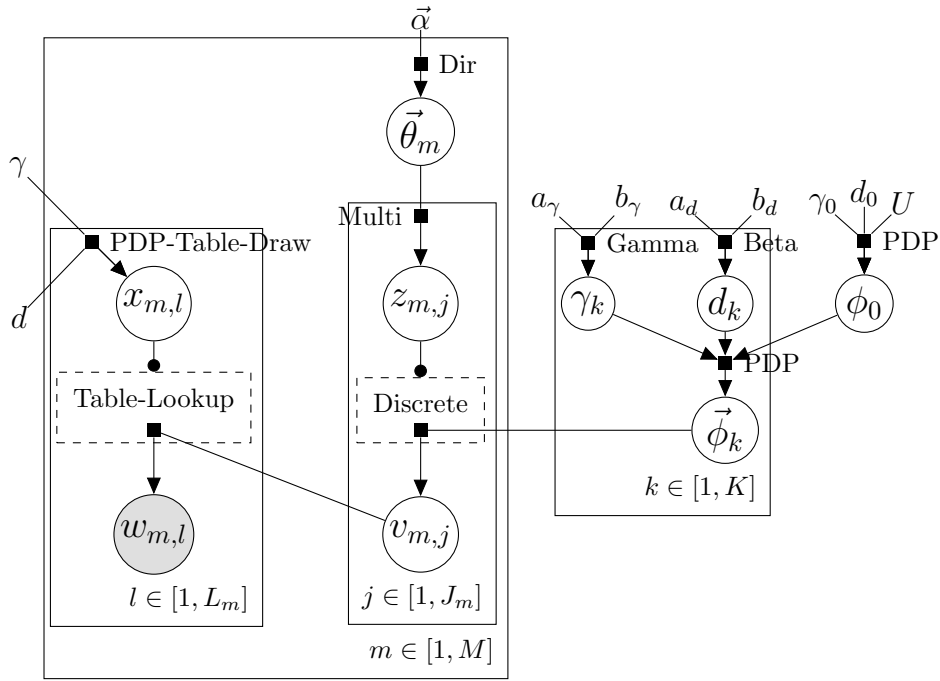

\paragraph*{\textmd{The Hierarchical Pitman-Yor Topic Modelling (HPYTM) \cite{conf/kdd/SatoN10}
made one extension to PYTM by assuming the power-law phenomenon not
only exists in each document but also within each topic. PDP word
generation is now document-topic specific instead of only document-specific
as it is in PYTM. In this setup, new words (type (2) words) are no
longer drawn from $Discrete(\vec{\phi}_{z_{m,l}})$, instead they
are drawn from a distribution generated by PDP for a specific topic.
The Hierarchical Bayesian Language Model \cite{conf/acl/Teh06} replaces
some parts of PYTM still inheriting some features of LDA with a more
complicated structure, as illustrated in Figure \ref{fig:PYTM-graphical-model}
and \ref{fig:HPYTM-breakdown}. Note $\vec{\beta}$ and $\vec{\phi}$
in PYTM have been replaced by a two-tier hierarchical model.}}

\paragraph*{\textmd{The break-down generation process is same as the generation
process in PYTM, except for $v_{m,x_{m,l}}$:}}

\paragraph*{\textmd{
\[
v_{m,x_{m,l}}=\protect\begin{cases}
\mbox{unchanged} & (r_{m,l}=0)\protect\\
\mbox{Draw from }Discrete(\vec{\phi}_{z_{m,l}}) & (r_{m,l}=1)
\protect\end{cases}
\]
}}

\paragraph*{\textmd{And in addition:}}

\paragraph*{\textmd{
\begin{alignat*}{2}
d_{k} & \sim Beta(a_{d},b_{d}) & \forall k=1,...,K\protect\\
\gamma_{k} & \sim Gamma(a_{\gamma},b_{\gamma}) & \forall k=1,...,K\protect\\
\vec{\phi}_{k} & \sim PDP(\gamma_{k},d_{k},\vec{\phi}_{0}) & \forall k=1,...,K\protect\\
\vec{\phi}_{0} & \sim PDP(\gamma_{0},d_{0},U)
\end{alignat*}
}}

\paragraph*{\textmd{Where $a_{d},b_{d},a_{\gamma},b_{\gamma},\gamma_{0},d_{0}$
are all hyper-parameters, $U$ is a discrete uniform distribution
($\forall w:p(w)=1/V$). }}

\begin{figure}
\begin{centering}
\begin{tikzpicture}
[latent/.style={circle,fill=white,draw=black,inner sep=1pt, minimum size=20pt, font=\fontsize{14}{14}\selectfont},
obs/.style={latent,fill=gray!25},
const/.style={rectangle, inner sep=0pt}]
\matrix[row sep=4mm,column sep=10mm, matrix anchor=mid] (HPYTM2) {
&
\\
&
\\
\node (theta) [latent] {$\vec{\theta}_{m}$}; 
& & & 
\node (topicgammak) [latent] {$\gamma_{k}$};
&
\node (topicdck)[latent] {$d_{k}$};
& 
\node (phi0) [latent] {$\phi_{0}$};
\\
\nofactor{dz}{}{$PY(\gamma ,d, H(\cdot,\cdot|\vec{\theta}_{m},\Phi))$}{left=0pt};
& & & & \node (phik) [latent]  {$\vec{\phi}_k$};
\\
\node(Gm)[latent]{$G_{m}$}; 

\\
\node (z) [latent] {$z_{m,l}$}; 

\\
\node (w) [obs] {$w_{m,l}$}; 
&
\\
};
\nofactor{dtheta}{above=10pt of theta}{Dir}{right=0pt};
\nofactor{dphik}{above=10pt of phik}{PDP}{right=0pt};
\nofactor{dtopicgammak}{above=10pt of topicgammak}{Gamma}{right=0pt};
\nofactor{dtopicdck}{above=10pt of topicdck}{Beta}{right=0pt};
\nofactor{dphi0}{above=10pt of phi0}{PDP}{right=0pt};
\node (alpha) [const, above=10pt of dtheta]  {$\vec{\alpha}$};
\node (atopicgammak) [const, above left=10pt of dtopicgammak] {$a_{\gamma}$};
\node (btopicgammak) [const, above right=10pt of dtopicgammak] {$b_{\gamma}$};
\node (atopicdck) [const, above left=10pt of dtopicdck] {$a_{d}$};
\node (btopicdck) [const, above right=10pt of dtopicdck] {$b_{d}$};
\node (topicgamma0) [const, above left=10pt of dphi0] {$\gamma_{0}$};
\node (topicdc0) [const, above=10pt of dphi0] {$d_{0}$};
\node (U) [const, above right=10pt of dphi0] {$U$};
\draw [->] (alpha) -- (dtheta) -- (theta) -- (dz);
\draw [->] (phik) -- (dz);
\draw [->] (dz) -- (Gm);
\draw [->] (Gm) -- (z);
\draw [-*] (z) -- (w);
\draw [->] (topicdc0) -- (dphi0) -- (phi0);
\draw [->] (topicgamma0) -- (dphi0) -- (phi0);
\draw [->] (U) -- (dphi0) -- (phi0);
\draw [->] (atopicgammak) -- (dtopicgammak) -- (topicgammak);
\draw [->] (btopicgammak) -- (dtopicgammak) -- (topicgammak);
\draw [->] (atopicdck) -- (dtopicdck) -- (topicdck);
\draw [->] (btopicdck) -- (dtopicdck) -- (topicdck);
\draw [->] (phi0) -- (dphik);
\draw [->] (topicgammak) -- (dphik);
\draw [->] (topicdck) -- (dphik);
\draw [->] (dphik) -- (phik);
\plate{wordplate}{(z)(w)}{$l \in{[1,L_m]}$}{}
\plate{docplate}{(wordplate)(theta)(dtheta)(dz)(Gm)}{$m \in{[1,M]}$}{}
\plate{topicplate}{(topicdck)(phi0)(topicgammak)(dtopicdck)(dtopicgammak)(dphik)(phik)(dphi0)}{$k\in{[1,K]}$}{}
\end{tikzpicture}
\par\end{centering}

\centering{}\protect\caption{HPYTM graphical model \label{fig:HPYTM-graphical-model}}
\end{figure}
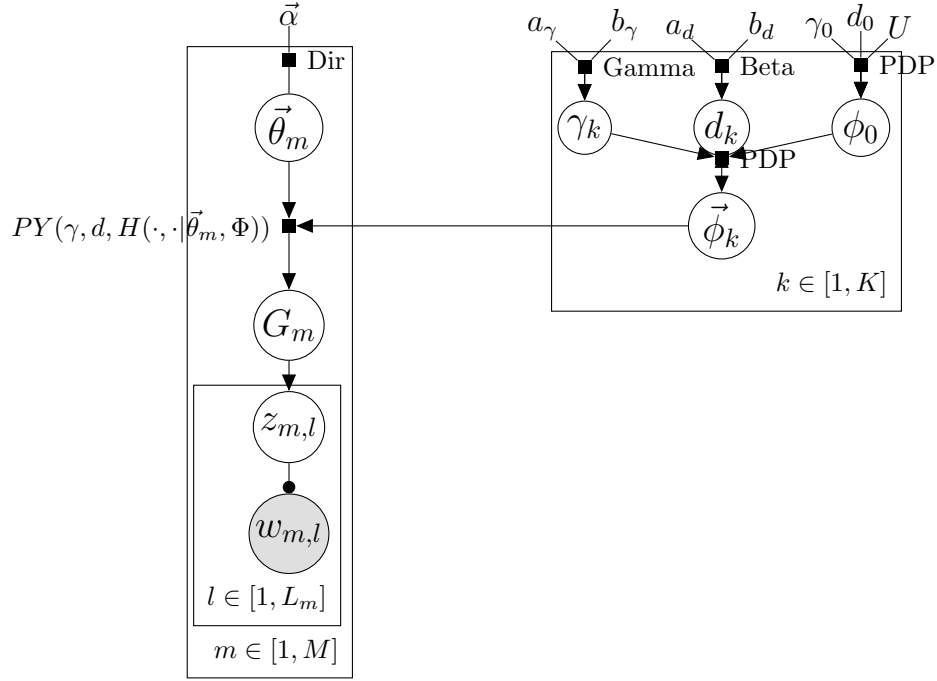

\selectlanguage{american}%

\selectlanguage{english}%
\FloatBarrier

\subsection{Differential Topic Modelling Using Shadow Poisson Dirichlet Process\label{sub:Differential-Topic-Modelling}}

\paragraph*{\textmd{This differential model \cite{chen2012spdp} addresses the
problem of comparing multiple groups of documents. Differential topic
models extend standard topic models by giving the model the ability
to find similarities and differences in topics among multiple groups
of documents. In this setup, the input documents are organized into
multiple groups sharing the same vocabulary. Topics are shared across
all groups but each group has their own representation of each topic.
In addition, each group is allowed to have their unique topics. }}

\paragraph*{\textmd{In standard topic modelling, the sources of documents are
not differentiated. In other words, all documents are assumed to be
in the same group. This assumption simplifies the mathematical formulation
of the topic model and the predictive probabilities, but with such
assumption in place, topic models are unable to extract information
on the variation in popularity of topics and words among multiple
sources of documents. The differential information is important because
it allows us to analyze subtle differences in opinions and perspectives
across multiple collection of documents.}}

\paragraph*{\textmd{Here is an example to illustrate the power of differential
topic models: consider a situation where we need to analyze articles
gathered from two media outlets, one from Israel and one from Palestine.
Apparently, Israeli editors are more likely to use the word ``terrorism''
to describe the Israeli-Palestinian conflict, because of some extreme
measures used by some Palestine. In contrast, Palestinians editors
are likely to use ``aggression'' to describe this topic, because
they see Israeli as invaders to their homeland. When standard topic
models are applied to both groups of documents individually, there
is no guarantee that the same topic can be extracted from both groups.
When standard topic models are applied to the whole collection of
two groups of documents, they are more likely to mix Israeli-Palestinian
issues into one topic, hence unable to provide differential analysis.
Differential topic models are able to find both the shared topics
among both groups, and different descriptions to such topics should
they exist.}}

\paragraph*{\textmd{Combining the essence of all models introduced above and
results from other mathematics research and topic modelling research,
especially the theoretical results in \cite{journals/corr/abs-1007-0296},
the improved PDP table-configuration sampler in \cite{chen2011sampling},
and the Hierarchical Dirichlet Process model \cite{teh06}, the Differential
Topic Model with Shadow Poisson Dirichlet Process (SPDP) is born.
This model outperforms many existing models in differential topic
modelling context when the performance is measured in perplexity. }}

\paragraph*{\textmd{Although the superiority of this model has already been demonstrated
in experiments, \cite{chen2012spdp} does not provide an in depth
explanation of the intuition of the model and derivation of the model.
Since the rest of the thesis is entirely based on this model, in the
following discussions I give an step-by-step explanation of this model.}}

\begin{figure}
\selectlanguage{english}%
\begin{centering}
\begin{tikzpicture}
[latent/.style={circle,fill=white,draw=black,inner sep=1pt, minimum size=20pt, font=\fontsize{14}{14}\selectfont},
obs/.style={latent,fill=gray!25},
const/.style={rectangle, inner sep=0pt}]
\matrix[row sep=4mm,column sep=10mm, matrix anchor=mid] (lda) {
&
\\
&
\\
\node (theta) [latent] {$\vec{\theta}_{i,d}$}; 
&
\\
\nofactor{dz}{}{Multi}{left=0pt};
&
\\
\node (z) [latent]  {$z_{i,d}^{l}$}; 
&
\\
\nofactor{dw}{}{Discrete}{above=0pt}; \gate{gw}{(dw) (captdw)}{$t$};
& &  \node (phi) [latent]  {$\vec{\phi}^{i}_{k}$};
& &  \node (phi0) [latent] {$\vec{\phi}^{0}_{k}$};
\\
\node (w) [obs]  {$w_{i,d}^{l}$}; 
\\
};
\nofactor{dtheta}{above=10pt of theta}{Dir}{right=0pt};
\nofactor{dphi}{right=10pt of phi}{PDP}{below=3pt};
\nofactor{dphi0}{right=10pt of phi0}{Dir}{below=0pt};
\node (alpha) [const, above=20pt of dtheta]  {$\vec{\alpha}$};
\node (beta) [const, right=10pt of dphi0]  {$\vec{\beta}$};
\node (gammak) [const, above=20pt of phi0]  {$\gamma_{k}$};
\node (dk) [const, below=20pt of phi0]  {$d_{k}$};
\node (P) [const, above=30pt of dphi]{$P_i$};
\draw [->] (beta) -- (dphi0) -- (phi0);
\draw [->] (phi0) -- (dphi) -- (phi);
\draw [->] (P) -- (dphi) -- (phi);
\draw [->] (gammak) -- (dphi) -- (phi);
\draw [->] (dk) -- (dphi) -- (phi);
\draw [->] (alpha) -- (dtheta) -- (theta);
\draw [->] (theta) -- (dz) -- (z);
\draw [->] (phi) -- (dw) -- (w);
\draw [-*] (z) -- (gw);
\plate{wordplate}{(z)(w)(dz)(dw)(captdz)(captdw)(gw)}{$l \in{[1,L_{i,d}]}$}{}
\plate{docplate}{(wordplate)(theta)(dtheta)(captdtheta)}{$d \in{[1,D_i]}$}{}
\plate{topicplate}{(phi)}{$k\in{[1,K]}$}{}
\plate{groupplate}{(docplate)(topicplate)(dphi)(P)}{$i \in{[1,I]}$}{}
\plate{topicplate2}{(gammak)(dk)(phi0)(dphi0)}{$k\in{[1,K]}$}{}
\end{tikzpicture}
\par\end{centering}

\selectlanguage{american}%
\protect\caption{\selectlanguage{english}%
SPDP graphical model\label{fig:SPDP-graphical-model}\selectlanguage{american}%
}
\end{figure}
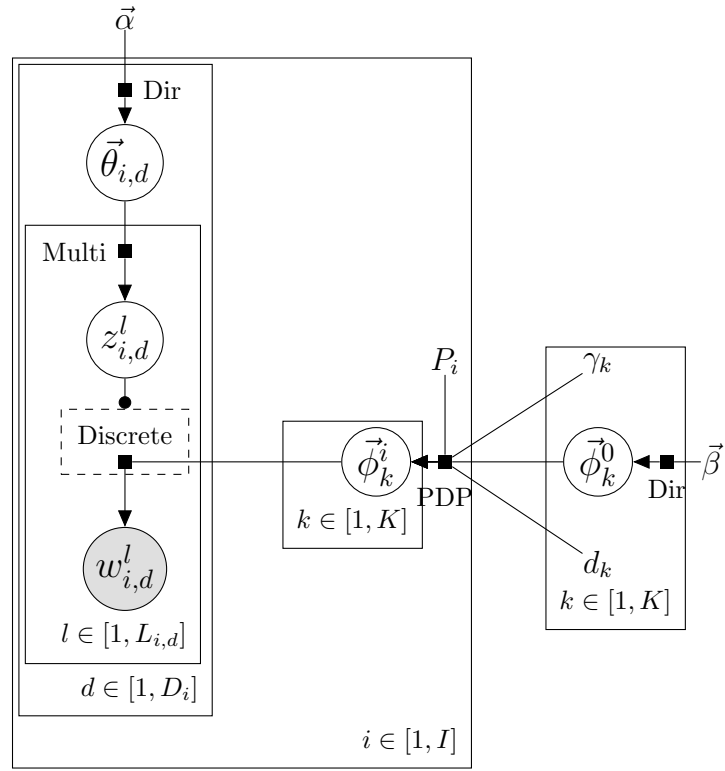

\paragraph*{\textmd{Similar to the structure in HPYTM, each word-topic distribution
is assumed to be generated from a base distribution $\vec{\phi}_{k}^{0}$.
Assume the total vocabulary size is $V$. Each group is attached with
a $V\times V$ transformation matrix $P^{i}$ that transforms the
shared base distribution $\vec{\phi}_{k}^{0}$, represents the similarity
between each pair of words, such that the sum of each row or column
in $P^{i}$ is $1$. As a consequence, different transformation matrices
introduce different word correlations for each group, so when words
are generated, each group produces slightly different words for a
common topic. The graphical model is given in \ref{fig:SPDP-graphical-model}.
The generation process is as follows. The technical details of this
model are left to next section, Model Derivation and Gibbs Sampler.}}

\paragraph*{\textmd{
\begin{alignat*}{2}
\vec{\phi}_{k}^{0} & \sim Dirichlet(\vec{\beta}) & \forall k=1,...,K\protect\\
\vec{\phi}_{k}^{i} & \sim PDP(\gamma_{k},d_{k},P^{i}\vec{\phi}_{k}^{0}) & \forall k=1,...,K,\,\forall i=1,...,I\protect\\
\vec{\theta}_{i}^{d} & \sim Dirichlet(\vec{\alpha}_{i}) & \forall i=1,...,I,\,\forall d=1,...,D_{i}\protect\\
z_{id}^{l} & \sim Discrete(\vec{\theta}_{i}^{d}) & \forall i=1,...,I,\,\forall d=1,...,D_{i},\,\forall l=1,...,L_{i,d}\protect\\
w_{id}^{l} & \sim Discrete(\vec{\phi}_{z_{id}^{l}}^{i}) & \forall i=1,...,I,\,\forall d=1,...,D_{i},\,\forall l=1,...,L_{i,d}
\end{alignat*}
}}

\paragraph*{\textmd{Where $I$ is number of groups, $P^{i}$ is the transformation
matrix for group $i$. $D_{i}$ is number of documents in group $i$.
$L_{i,d}$ is document length of document $d$ in group $i$.}}

\selectlanguage{american}%

\selectlanguage{english}%
\FloatBarrier

\paragraph*{\textmd{The differential topic model SPDP is one particular representative
of perhaps another hundred extensions of LDA. When combined with other
algorithms and models, it has even more potential applications in
practice, For instance, in many real world situations, group labels
on collections of document are not accurately given. Many blogs and
articles are shared around many websites on the Internet, without
providing any reliable label of originality, category, or perspective.
The majority of documents on the Internet (including social network
messages such as tweets) are not tagged. Furthermore, tags do not
always provide accurate information. Multiple documents sharing the
same tag may belong to different categories or be written in different
perspectives, hence do not necessarily belong to the same group. For
example, a document tagged with ``machine learning'' could be in
``reinforcement learning'', ``topic modelling'', or other categories;
a document tagged with ``politics'' could be either ``left'' or
``right'' depending on its perspective. It is desirable to have
an algorithm that automatically classifies documents into different
categories and perspectives, free from human influence and judgements.
Because the differential topic model SPDP provide a fundamental framework
based on groups, it is a very suitable candidate for this task. For
example, a simple and naive solution is to iteratively use the topic
probabilities and word probabilities across multiple groups generated
by SPDP as features for machine learning algorithms such as Expectation
Maximization (EM) to classify documents belonging to multiple unknown
groups.}}

\clearpage{}

\section{Model Derivation and Gibbs Sampler}

\paragraph*{\textmd{All topic models mentioned in last section share some similarities
in their derivations and inference processes, because fundamentally
all of them are extensions of LDA. The collapsed Gibbs sampler is
most frequently chosen by authors of above models to make inference
on the multivariate latent variables. Although different models have
different latent variables, for simplicity, we denote all of them
by one latent parameter $\vec{\theta}\in\Theta_{latent}$. The goal
is to make inference on the latent parameter (which include topic
information) by estimating the Bayesian posterior of latent parameters
(for example in LDA, word distributions $\phi_{k}$, and topic distributions
$\theta_{m}$) given some data, according to Bayes' rule:}}

\paragraph*{\textmd{
\[
p(\vec{\theta}|data)=\frac{p(\vec{\theta},data)}{p(data)}
\]
}}

\paragraph*{\textmd{Simply applying this formula often gives an intractable probability
distribution for $p(\vec{\theta}|data)$. To see this, consider the
simplest LDA model. To infer the latent topic assignment variable
$\vec{z}$:}}

\paragraph*{\textmd{
\[
p(\vec{z}|\vec{w})=\frac{p(\vec{z},\vec{w})}{p(\vec{w})}=\frac{\prod_{l=1}^{L}p(z_{l},w_{l})}{\prod_{l=1}^{L}\sum_{k=1}^{K}p(z_{l}=k,w_{l})}
\]
}}

\paragraph*{\textmd{There are $K^{L}$ terms in the denominator, making the probability
mathematically intractable. }}

\paragraph*{\textmd{However, if we are given a large number of samples from this
probability distribution, we may find a number of ways to estimate
the latent variables $\vec{\theta}$. Suppose we have samples $\vec{x}_{1},..\vec{x}_{n}$
from $p(\vec{\theta}|data)$. The easiest way to estimate it is to
simply calculate the average occurrence of each possible outcome:}}

\paragraph*{\textmd{
\[
\hat{p}(\vec{\theta}|data)=\frac{1}{n}\sum_{i=1}^{n}\delta_{\vec{\theta}}(\vec{x}_{i})
\]
}}

\paragraph*{\textmd{Where $\vec{x}_{i}$ are the observed samples. The collapsed
Gibbs sampler is used for this computation. The collapsed Gibbs sampler
is an iterative sampler that generate random samples of a joint multivariate
probability distribution $p(\vec{x})$ given $p(x_{i}|\vec{x}_{-i})$
is known. Here $\vec{x}_{-i}$ denotes $\vec{x}$ with element $x_{i}$
deleted. The collapsed Gibbs sampler requires a large number iterations
for convergence. It starts with an arbitrary initial value for each
$x_{i}$, sample each element $x_{i}$ from $p(x_{i}|\vec{x}_{-i})$
in each iteration, and update the value of $x_{i}$ as soon as it
is sampled. The probability of samples $\vec{x}$ drawn in each iteration
will almost surely converge to $p(\vec{\theta}|data)$. The proof
can be found in most advanced statistics textbooks, such as \cite{Robert2004}.}}

\paragraph*{\textmd{In our settings, usually the quantity $p(\theta_{i}|\theta_{-i},data)$
can be easily deduced by computing:}}

\paragraph*{\textmd{
\begin{equation}
p(\theta_{i}|\theta_{-i},data)=\frac{p(\vec{\theta},data)}{p(\theta_{-i},data)}\label{eq:joint-to-predictive}
\end{equation}
}}

\paragraph*{\textmd{In topic modelling this is often referred as the predictive
probability. To make our method work, we need a neat way to compute
the predictive probability which is derived from the the joint distribution
of the latent parameter and the data. In the rest of this section,
we will show the derivation process of joint distribution and predictive
probability for most models we described in last section. }}

\paragraph*{\textmd{For most of these models, we have written a step by step
derivation. For the basic LDA model, we only explain the important
steps and provided references to existing publications, to help readers
find more detailed explanation.}}

\subsection{Notation}

\paragraph*{\textmd{Across this section, we use $V$ to denote the size of the
vocabulary, $K$ to denote the total number of topics, $M$ to denote
total number of documents, $n_{mk:m,k}$ to denote number of times
topic $k$ observed in document $m$, where the first part of subscript
is a label to distinguish it from other counting variables that may
also be named with $n$, and $n_{kw:k,w}$ to denote number of times
word $w$ associated with topic $k$. We use $n_{kw:k,.}$ to denote
the sum over dotted variables in the subscript, for example, $n_{kw:k,.}=\sum_{w=1}^{V}n_{kw:k,w}$,
and similarly $n_{mk:m,.}=\sum_{k=1}^{K}n_{mk:m,k}$.}}

\subsection{Latent Dirichlet Allocation (LDA)}

\paragraph*{\textmd{The joint distribution $p(\vec{\theta},data)$ of LDA can
be represented as}}

\paragraph*{\textmd{
\begin{equation}
p(\vec{w},\vec{z}|\vec{\alpha},\vec{\beta})=p(\vec{w}|\vec{z},\vec{\beta})p(\vec{z}|\vec{\alpha})\label{eq:LDA-joint-step1}
\end{equation}
}}

\paragraph*{\textmd{Since $\vec{z}$ is not dependent on $\vec{\beta}$. The
first term $p(\vec{w}|\vec{z},\vec{\beta})$ in \ref{eq:LDA-joint-step1}
can be derived as:}}

\paragraph*{\textmd{
\begin{eqnarray}
p(\vec{w}|\vec{z},\vec{\beta}) & = & \int_{\vec{\phi}}p(\vec{w}|\vec{z},\vec{\phi})p(\vec{\phi}|\vec{\beta})d\vec{\phi}\nonumber \protect\\
 & = & \int_{\vec{\phi}}\prod_{k=1}^{K}\frac{1}{\Delta(\vec{\beta})}\prod_{w=1}^{V}\phi_{k,w}^{n_{k,w}+\beta_{w}-1}d\vec{\phi}\nonumber \protect\\
 & = & \prod_{k=1}^{K}\frac{\Delta(\vec{\beta}+\vec{n}_{k})}{\Delta(\vec{\beta})}\label{eq:LDA-step2}
\end{eqnarray}
}}

\paragraph*{\textmd{Where $n_{k,w}$ is a counting variable representing number
of terms in all documents that have been assigned to topic $k$ and
word $w$. $\Delta(\vec{x})$ is the Dirichlet delta function, the
normalizing constant, as defined in \cite{oai:fraunhofer.de:N-101883}:}}

\paragraph*{\textmd{
\[
\Delta(\vec{x})=\frac{\prod_{k=1}^{\dim\vec{x}}\Gamma(x_{k})}{\Gamma(\prod_{k=1}^{\dim\vec{x}}(x_{k}))}
\]
}}

\paragraph*{\textmd{Similarly, the second term $p(\vec{z}|\vec{\alpha})$ in
\ref{eq:LDA-joint-step1} can be derived as:}}

\paragraph*{\textmd{
\begin{eqnarray}
p(\vec{z}|\vec{\alpha}) & = & \int_{\vec{\theta}}p(\vec{z}|,\vec{\theta})p(\vec{\theta}|\vec{\alpha})d\vec{\theta}\nonumber \protect\\
 & = & \int_{\vec{\theta}}\prod_{d=1}^{M}\frac{1}{\Delta(\vec{\alpha})}\prod_{k=1}^{K}\theta_{m,k}^{n_{m,k}+\alpha_{k}-1}d\vec{\theta}\nonumber \protect\\
 & = & \prod_{k=1}^{M}\frac{\Delta(\vec{\alpha}+\vec{n}_{m})}{\Delta(\vec{\beta})}\label{eq:LDA-step3}
\end{eqnarray}
}}

\paragraph*{\textmd{Where $n_{m,k}$ is a counting variable representing number
of terms in document $m$ that have been assigned with topic $k$.
Putting equations \ref{eq:LDA-joint-step1}\ref{eq:LDA-step2}\ref{eq:LDA-step3}
together:}}

\paragraph*{\textmd{
\begin{eqnarray}
p(\vec{w},\vec{z}|\vec{\alpha},\vec{\beta}) & = & p(\vec{w}|\vec{z},\vec{\beta})p(\vec{z}|\vec{\alpha})\nonumber \protect\\
 & = & \prod_{k=1}^{K}\frac{\Delta(\vec{\beta}+\vec{n}_{k})}{\Delta(\vec{\beta})}\prod_{k=1}^{M}\frac{\Delta(\vec{\alpha}+\vec{n}_{m})}{\Delta(\vec{\beta})}\label{eq:LDA-joint}
\end{eqnarray}
}}

\paragraph*{\textmd{Substitutes \ref{eq:LDA-joint} into \ref{eq:joint-to-predictive}
:}}

\paragraph*{\textmd{
\begin{eqnarray*}
p(z_{i}=k|\vec{w},\vec{z}_{-i},\vec{\alpha},\vec{\beta}) & = & \frac{p(\vec{w},\vec{z})}{p(\vec{w},\vec{z}_{-i})}\protect\\
 & = & \frac{p(\vec{w}|\vec{z})}{p(\vec{w}_{-i},\vec{z}_{-i})}\frac{p(\vec{z})}{p(w_{i})}\protect\\
 & \propto & \frac{\Delta(\vec{\beta}+\vec{n}_{k})}{\Delta(\vec{\beta}+\vec{n}_{k,-i})}\frac{\Delta(\vec{\alpha}+\vec{n}_{m})}{\Delta(\vec{\alpha}+\vec{n}_{m,-i})}\protect\\
 & \propto & \frac{\Gamma(n_{k,w}+\beta_{w})\Gamma(\sum_{w=1}^{V}(n_{k,w,-i}+\beta_{w}))}{\Gamma(n_{k,w,-i}+\beta_{w})\Gamma(\sum_{w=1}^{V}(n_{k,w}+\beta_{w}))}\frac{\Gamma(n_{m,k}+\alpha_{k})\Gamma(\sum_{k=1}^{K}(n_{m,k,-i}+\alpha_{k}))}{\Gamma(n_{m,k,-i}+\alpha_{k})\Gamma(\sum_{k=1}^{K}(n_{m,k}+\alpha_{k}))}\protect\\
 & \propto & \frac{n_{k,w,-i}+\beta_{w}}{\sum_{w=1}^{V}(n_{k,w,-i}+\beta_{w})}\frac{n_{m,k,-i}+\alpha_{k}}{\sum_{k=1}^{K}(n_{m,k,-i}+\alpha_{k})-1}
\end{eqnarray*}
}}

\paragraph*{\textmd{Where a variable with subscript $-i$ denotes the value of
such variable with element $i$ (word $i$) is removed. Above formula
gives the proportional predictive probability for the collapsed Gibbs
sampling. For more detailed explanation on LDA, readers should refer
to \cite{bleinj03}.}}

\subsection{Pitman-Yor Topic Modelling (PYTM)}

\subsubsection{Poisson Dirichlet Process}

\paragraph*{\textmd{We first give an overview of the Poisson Dirichlet Process
(PDP) as it is the foundation of PYTM. As mentioned in the last section,
PDP generates a probability distribution from a base distribution
$H(.)$, concentration parameter $\gamma$, and discount parameter
$d$. The process is denoted by $PDP(\gamma,d,H(.))$. The Chinese
Restaurant analogy is as follows: In a strange Chinese restaurant
which has infinite number of tables, each table serves only one dish,
and only when at least one customer is sitting on that table. Waiters
will arrange each incoming customer to either sit in a table served
with dish $j$, share it among other people who are already sitting
there, or lead the customer to an empty table and immediately serve
a dish. Waiters keep a record of the number of distinct dishes $1,...,J$
served to customers, the number of customers served with dish $j$,
denoted by $n_{j}$, and the total number of customers, denoted by
$N$. They use the following procedure to make seating arrangements
for each incoming customer:}}
\begin{itemize}
\item Take the customer to an empty table with probability $\frac{\gamma+dJ}{\gamma+N}$,
and serve some dish $j$ drawn from $H(.)$
\item Otherwise, take the customer to some table already serving dish $j$,
with probability $\frac{n_{j}-d}{\gamma+N}$ in total for all these
tables
\end{itemize}

\paragraph*{\textmd{Samples drawn from the distribution generated by the PDP
can be understood as the dishes served to each customer. In the PYTM
model (Figure \ref{fig:PYTM-graphical-model} ), a restaurant is created
for each document. Each word in the document is a customer coming
to the restaurant. Each different word in the vocabulary is an unique
type of dish. $x_{m,l}$ records which table the customer sat in.
$v_{m,s}$ records the dish being served at table $s$ in restaurant
$m$. $z_{m,s}$ records the topic associated with table $s$ in restaurant
$m$. The observed words $w_{m,l}=v_{m,x_{m,l}}$ are the samples
drawn from the distribution generated from PDP.}}

\paragraph*{\textmd{The formal definition of the Poisson Dirichlet Process is
a sequence of draws from the base distribution $H(.)$ coupled with
probability weighting vector $\vec{p}$ drawn from Poisson Dirichlet
Distribution as stated in \cite{pitman-yor-97}. A detailed Bayesian
analysis of Poisson Dirichlet Process is given in \cite{journals/corr/abs-1007-0296}
by Buntine and Hutter. Because their works are highly technical, far
above the level of this thesis, and they are not directly related
to topic modelling, we only use only some of their results and only
show the derivation when necessary.}}

\paragraph*{\textmd{Following the analogy we can immediately get the predictive
probability of which dish would be served to an incoming customer:}}

\[
p(x_{i}=j|x_{1},x_{2},...,x_{i-1},\gamma,d,H(.))=\frac{n_{j}-d}{\gamma+N}+\frac{\gamma+dJ}{\gamma+N}H(j)
\]

\subsubsection{Predictive Probability and Inference}

\paragraph*{\textmd{Predictive probability can be derived by removing a word,
similar to LDA. Here we need both topic predictive probability and
word predictive probability to proceed with inference, because when
we remove a word, we add it back later, and we need to reconsider
the sitting arrangement when add it back. Luckily the word predictive
probability is directly given by CRP analogy because in above definition
each word drawn is dependent on all previous words drawn : 
\begin{equation}
p(w_{m,l}=j|\boldsymbol{W},\boldsymbol{Z},\boldsymbol{X})=\frac{n_{mw:m,j}-d}{\gamma+n_{mw:m.}}+\frac{\gamma+dJ_{m}}{\gamma+n_{mw:m.}}\sum_{k=1}^{K}\frac{n_{m,k}+\alpha_{m}}{n_{m}+\alpha_{m:.}}\frac{n_{kw:k,j}+\beta_{k}}{n_{kw:k,.}+\beta_{k:.}}
\end{equation}
}}

\paragraph*{\textmd{Where $n_{mj:m,j}$ is number of times word $w$ appeared
in document $m$ (not including $w_{m,l}$), $n_{mw:m.}$is number
of words observed so far in document $m$ without $w_{m,l}$, $\boldsymbol{W},\boldsymbol{Z},\boldsymbol{X}$
are words, topics, sitting arrangements not including $w_{m,l}$ or
anything associated with $w_{m,l}$. $H(.)$ is replaced by the summation
term, which is borrowed from LDA word predictive probability as for
this part they share the same word generation procedure (see figure
\ref{fig:PYTM-graphical-model-breakdown} and the breakdown generation
illustration.)}}

\paragraph*{\textmd{Similarly, the topic predictive probability is given by}}

\paragraph*{\textmd{
\[
p(z_{m,l}=k|\boldsymbol{W},\boldsymbol{Z},\boldsymbol{X},w_{m,l}=j,x_{m,l}=s)=\frac{n_{mk:m,k}+\alpha_{m}}{n_{mk:m,.}+\alpha_{m:.}}\frac{n_{kw:k,j}+\beta_{k}}{n_{kw:k,.}+\beta_{k:.}}
\]
}}

\paragraph*{\textmd{as derived in LDA.}}

\subsection{Hierarchical Pitman-Yor Topic Modelling (HPYTM) }

\paragraph*{\textmd{The derivation process is almost same as described in PYTM,
except probabilities are computed recursively. We will skip this section
as the detail is not particularly related to SPDP. Readers should
refer to \cite{conf/kdd/SatoN10} if they are interested in the details.}}

\subsection{Shared Topic Modelling Using Shadow Poisson Dirichlet Process}

\paragraph*{\textmd{As mentioned in the last section the Shadow Poisson Dirichlet
Process (SPDP) is in fact a Poisson Dirichlet Process coupled with
linearly transformed base measure. One important property we used
to derive the predictive probability in LDA is Dirichlet distribution
is conjugate to Discrete (categorical, or multinomial) distribution.
The same property is used in PYTM and its extension HPYTM as they
use LDA as a foundation. In this model the same method does not apply
because the transformed base measure is no longer conjugate to a Discrete
(categorical, or multinomial) distribution. }}

\paragraph*{\textmd{To overcome this, first we introduce auxiliary variable $t_{i,k,w}$,
which we refer as multiplicity, represents number of tables served
with dish $w$ in restaurant $i,k$ (group $i$, topic $k$). It is
shown by Corollary 17 in \cite{journals/corr/abs-1007-0296} that
in one ``restaurant'':}}

\paragraph*{\textmd{
\[
p(\vec{w},\vec{t}|\gamma,d,H(.))=\frac{(\gamma|d)_{t_{.}}}{(d)_{n_{.}}}\prod_{j=1}^{J}(H(w_{j}^{\star})^{t_{j}}S_{t_{j},\gamma}^{n_{j}})
\]
}}

\paragraph*{\textmd{Where $J$ is the number distinct dishes, $\vec{w}$ is dish
served to each customer and $\vec{t}$ is the multiplicity of each
dish. $(w_{1}^{\star},...,w_{j}^{\star})$ is the sequence of distinct
dishes. $t_{.}$ is the sum of multiplicities, equivalent to total
number of non-empty tables. $n_{j}$ is number of customers having
dish $j$, and $n_{.}$ is equivalent to total number of customers.
$(x|y)_{N}$ and $(x)_{N}$ are Pochhammer symbol, where $(x|y)_{N}=\prod_{n=0}^{N-1}(x+ny)$,
and $(x)_{N}=(x|1)_{N}$. $S_{M,a}^{N}$ is a generalized Sterling
number, given by linear recursion $S_{M,a}^{N+1}=S_{M-1,a}^{N}+(N-Ma)S_{M,a}^{N}$
and $S_{M,a}^{N}=0$ for $M>N$, $S_{0,a}^{N}=\delta_{N,0}$. Both
generalized Sterling numbers and Pochhammer symbols can be computed
and cached efficiently before the Gibbs sampling process starts.}}

\paragraph*{\textmd{In our settings $H(w_{j}^{\star})$ is replaced by the probability
$\phi_{v}^{0}$. After transformation the base distribution becomes
$P^{i}\vec{\phi}^{0}$, and $\phi_{w}^{0}=\sum_{v}p_{w,v}\phi_{v}^{0}$
where $p_{w,v}$ denotes element of the matrix. Therefore:}}

\paragraph*{\textmd{
\begin{equation}
p(\vec{w},\vec{t}|\gamma,d,\vec{\phi}^{0})=\frac{(\gamma|d)_{t_{.}}}{(d)_{n_{.}}}\prod_{j=1}^{J}S_{t_{j},\gamma}^{n_{j}}(\sum_{v}p_{w,v}\phi_{v}^{0})^{t_{j}}\label{eq:SPDP-nested-joint-prob}
\end{equation}
}}

\paragraph*{\textmd{It is clear the summation term inside the product has to
be simplified. Chen et al. introduced two solutions for this problem:
blocked Gibbs sampling, and hybrid Gibbs sampling with variational
method. }}

\paragraph*{\textmd{The hybrid Gibbs with variational method approximates the
above equation by deriving an inequality for above equation with $\vec{\phi}$
integrated out, then introduces another variable $q_{w,v}=\frac{p_{w,v}\phi_{v}^{0}}{\sum_{v'}p_{w,v'}\phi_{v'}^{0}}$
that can be substituted to the lower bound. By using Jensen's inequality
and a Lagrange multiplier, it can be shown that $q_{w,v}$ maximize
previously derived inequality after substitution. As the summation
term is simplified, one then derive the sampling predictive probabilities
as usual with equation \ref{eq:joint-to-predictive}. The latent variables
such as word probability $\phi_{k,v}^{0}$ , can be computed by}}

\paragraph*{\textmd{
\begin{equation}
\phi_{k,v}^{0}=e^{\psi(\gamma_{v}+\sum_{i}\sum_{w}q_{k,w,v}^{i}t_{i,k,w})}/\sum_{v}e^{\psi(\gamma_{v}+\sum_{i}\sum_{w}q_{k,w,v}^{i}t_{i,k,w})}\label{eq:spdp-phi0estimate}
\end{equation}
}}

\paragraph*{\textmd{where $\psi$ is the digamma function. However, experiements
have shown this approximation is not very accurate, as error accumulates
the performance of the algorithm degrades significantly. Therefore
in the rest of this section we will be concentrating on the first
method: blocked Gibbs sampling.}}

\paragraph*{\textmd{In equation \ref{eq:SPDP-nested-joint-prob} suppose for
each word $w$ we have another auxilliary variable $\vec{v}$ which
has dimension $t_{w}$, the multiplicity of the word in one PDP process.
We need $\vec{v}$ to separate the power term $(\sum_{v}p_{w,v}\phi_{v}^{0})^{t_{j}}$
into this form $(\sum_{v_{1}}p_{w,v_{1}}\phi_{v}^{0})(\sum_{v_{2}}p_{w,v_{2}}\phi_{v_{2}}^{0})...(\sum_{v_{t_{w}}}p_{w,v_{t_{w}}}\phi_{v_{t_{w}}}^{0})$,
such that the terms inside each bracket is dependent on $v_{i}$ and
its effect is marginalized out in \ref{eq:SPDP-nested-joint-prob}.
If we compute the joint probability as in equation \ref{eq:SPDP-nested-joint-prob}
with some $\vec{v}$ we will get:}}

\paragraph*{\textmd{
\begin{equation}
p(\vec{w},\vec{t},\vec{v}|\gamma,d,\vec{\phi}^{0})=\frac{(\gamma|d)_{t_{.}}}{(d)_{n_{.}}}\prod_{j=1}^{J}S_{t_{j},\gamma}^{n_{j}}\prod_{s=1}^{t_{w}}p_{w,v_{s}}\phi_{v_{s}}^{0}\label{eq:SPDP-joint-prob-with-v}
\end{equation}
}}

\paragraph*{\textmd{Which is much simpler than \ref{eq:SPDP-nested-joint-prob},
simple enough to be efficiently computed in logarithm space.}}

\paragraph*{\textmd{Because of the dimensionality difference between $\vec{w}$
and $\vec{v}$, to make use of the auxilliary variable $\vec{v}$
we need another auxilliary variable $\vec{r}$ called table indicator
that has same dimension as $\vec{w}$ , constructed as follows: for
each word (customer) $w_{i,d,l}$ we set $r_{i,d,l}=1$ if the word
has created a new table (the customer is arranged to an empty table
and served a new dish), otherwise we set $r_{i,d,l}=0$. For every
word $w_{i,d,l}$ such that $r_{i,d,l}=1$ we associate the word with
$v_{i,k,w,t}$ where $t$ is the table index of the word. $\vec{r}$
coupled with $\vec{t}$ provides more information than $\vec{t}$
alone because $\vec{r}$ in a specific seating configuration to each
word in the document and $\vec{t}$ disregards the information of
which word is creator of the table. For each table configuration specified
by $\vec{t}$, there are $\prod_{v=1}^{V}\begin{pmatrix}n_{v}\protect\\
t_{v}
\end{pmatrix}$ different configurations of $\vec{r}$, thus}}

\begin{equation}
p(\vec{w},\vec{t}|\gamma,d,\vec{\phi}^{0})=\prod_{v=1}^{V}\begin{pmatrix}n_{v}\\
t_{v}
\end{pmatrix}p(\vec{r},\vec{t}|\gamma,d,\vec{\phi}^{0})\label{eq:SPDP-table-to-head}
\end{equation}

\paragraph*{\textmd{Now we are ready to deduce a joint likelihood of these variables,
which afterwards can be easily transformed into a predictive probability
using \ref{eq:joint-to-predictive} : 
\begin{eqnarray}
p(\mathbf{W},\mathbf{Z},\mathbf{V},\mathbf{R}|\vec{\gamma},\vec{d},\vec{\alpha_{1:I}},\vec{\beta},\mathbf{P}) & = & \int_{\Phi}p(\mathbf{W},\mathbf{Z},\mathbf{V},\mathbf{R}|\vec{\phi}_{1:K}^{0},\vec{\gamma},\vec{d},\vec{\alpha}_{1:I},\mathbf{P})p(\vec{\phi}_{1:K}^{0}|\vec{\beta})d\Phi\nonumber \protect\\
 & = & p(\mathbf{Z}|\vec{\alpha}_{1:I})\int_{\Phi}p(\mathbf{W},\mathbf{V},\mathbf{R}|\mathbf{Z},\vec{\phi}_{1:K}^{0},\vec{\gamma},\vec{d},\mathbf{P})p(\vec{\phi}_{1:K}^{0}|\vec{\beta})d\Phi\nonumber \protect\\
\label{eq:SPDP-Blocked-Gibbs-Joint-Prob}
\end{eqnarray}
}}

\paragraph*{\textmd{Where $\Phi$ is the collection of $\vec{\phi}_{1:K}^{0}$.
Let $\Theta$ be the collection of $\vec{\theta}_{1:I,1:D}$, then}}

\paragraph*{\textmd{
\begin{eqnarray*}
p(\mathbf{Z}|\vec{\alpha}_{1:I}) & = & \int_{\Theta}p(\mathbf{Z}|\Theta)p(\Theta|\vec{\alpha}_{1:I})d\Theta\protect\\
 & = & \int_{\Theta}\prod_{i=1}^{I}\prod_{d=1}^{D}\frac{1}{Beta_{K}(\vec{\alpha}_{i})}\prod_{k=1}^{K}\theta_{i,d,k}^{n_{idk:i,d,k}}\theta_{i,d,k}^{\alpha_{i,k}-1}d\Theta\protect\\
 & = & \prod_{i=1}^{I}\prod_{d=1}^{D}\frac{Beta_{K}(\vec{\alpha}_{i}+\vec{n}_{idk:i,d})}{Beta_{K}(\vec{\alpha}_{i})}
\end{eqnarray*}
}}

\paragraph*{\textmd{Where $\vec{n}_{idk:i,d}=(n_{idk:i,d,1},n_{idk:i,d,2},...,n_{idk:i,d,K})$.
Substitute equation \ref{eq:SPDP-joint-prob-with-v} and \ref{eq:SPDP-table-to-head}
into the terms inside the integral in \ref{eq:SPDP-Blocked-Gibbs-Joint-Prob}:}}

\paragraph*{\textmd{
\begin{eqnarray}
p(\mathbf{W},\mathbf{V},\mathbf{R}|\mathbf{Z},\vec{\phi}_{1:K}^{0},\vec{\gamma},\vec{d},\mathbf{P}) & = & \prod_{i=1}^{I}\prod_{k=1}^{K}\frac{(\gamma|d)_{t_{ikw:i,k,.}}}{(d)_{n_{ikw:i,k,.}}}\nonumber \protect\\
 &  & \bigg(\prod_{w=1}^{V}\begin{pmatrix}n_{w}\protect\\
t_{w}
\end{pmatrix}^{-1}S_{t_{ikw:i,k,j},\gamma_{k}}^{n_{ikj:i,k,j}}\prod_{s=1}^{t_{ikw:i,k,w}}p_{i,w,v_{s}}\phi_{k,v_{s}}^{0}\bigg)\nonumber \protect\\
p(\vec{\phi}_{1:K}^{0}|\vec{\beta}) & = & \prod_{k=1}^{K}\frac{1}{Beta(\vec{\beta})}\prod_{w=1}^{V}(\phi_{k,w}^{0})^{\beta_{v}-1}\label{eq:SPDP-joint-derivation-step0}
\end{eqnarray}
}}

\paragraph*{\textmd{With respect to $\Phi$ the integral in \ref{eq:SPDP-Blocked-Gibbs-Joint-Prob}
can be easily evaluated as a multinomial probability density function.
To simplify the result more, we introduce an auxilliar statistic $q_{i,k,w,v}=\sum_{t=1}^{t_{ikw:i,k,w}}1_{v_{i,k,w,t=v}}$,
the number of tables associated (as defined by $v_{ikwt}$) with one
particular word $v$ in a PDP process. By doing so, $\prod_{s=1}^{t_{ikw:i,k,w}}p_{i,w,v_{s}}\phi_{k,v_{s}}^{0}$
can be neatly rewritten as $\prod_{v=1}^{V}(p_{w,v}\phi_{k,v}^{0})^{q_{i,k,w,v}}$,
thus:}}

\paragraph*{\textmd{
\begin{alignat}{1}
 & \int_{\Phi}p(\mathbf{W},\mathbf{V},\mathbf{R}|\mathbf{Z},\vec{\phi}_{1:K}^{0},\vec{\gamma},\vec{d},\mathbf{P})p(\vec{\phi}_{1:K}^{0}|\vec{\beta})d\Phi\nonumber \protect\\
= & \int_{\Phi}(\prod_{i=1}^{I}\prod_{k=1}^{K}\frac{(\gamma|d)_{t_{ikw:i,k,.}}}{(d)_{n_{ikw:i,k,.}}}\prod_{w=1}^{V}\begin{pmatrix}n_{w}\protect\\
t_{w}
\end{pmatrix}^{-1}S_{t_{ikw:i,k,j},\gamma_{k}}^{n_{ikj:i,k,j}}\prod_{s=1}^{t_{ikw:i,k,w}}p_{i,w,v_{s}}\phi_{v_{s}}^{0})\nonumber \protect\\
 & \prod_{k=1}^{K}\frac{1}{Beta(\vec{\beta})}\prod_{w=1}^{V}(\phi_{k,w}^{0})^{\beta_{w}-1}d\Phi\label{eq:SPDP-joint-derivation-step1-substitution}\protect\\
= & \int_{\Phi}\prod_{k=1}^{K}\frac{1}{Beta(\vec{\beta})}(\prod_{i=1}^{I}\frac{(\gamma|d)_{t_{ikw:i,k,.}}}{(d)_{n_{ikw:i,k,.}}}\prod_{w=1}^{V}\begin{pmatrix}n_{w}\protect\\
t_{w}
\end{pmatrix}^{-1}S_{t_{ikw:i,k,j},\gamma_{k}}^{n_{ikj:i,k,j}}\nonumber \protect\\
 & \prod_{v=1}^{V}(\phi_{k,v}^{0})^{q_{i,k,w,v}+\beta_{w}-1}(p_{w,v})^{q_{i,k,w,v}})d\Phi\label{eq:SPDP-joint-derivation-step2-use-q}\protect\\
= & \prod_{k=1}^{K}\frac{1}{Beta(\vec{\beta})}\bigg((\prod_{i=1}^{I}\frac{(\gamma|d)_{t_{ikw:i,k,.}}}{(d)_{n_{ikw:i,k,.}}}\prod_{w=1}^{V}\begin{pmatrix}n_{w}\protect\\
t_{w}
\end{pmatrix}^{-1}S_{t_{ikw:i,k,j},\gamma_{k}}^{n_{ikj:i,k,j}}\bigg)\nonumber \protect\\
 & \bigg(\int_{\Phi}\prod_{i=1}^{I}\prod_{w=1}^{V}\prod_{v=1}^{V}(\phi_{k,v}^{0})^{q_{i,k,w,v}+\beta_{w}-1}(p_{w,v})^{q_{i,k,w,v}})d\Phi)\bigg)\label{eq:SPDP-joint-derivation-step3-rearranging}
\end{alignat}
}}

\paragraph*{\textmd{Evaluate the integral, we get }}

\paragraph*{\textmd{
\begin{alignat}{1}
 & \prod_{k=1}^{K}\int_{\Phi}\prod_{i=1}^{I}\prod_{w=1}^{V}\prod_{v=1}^{V}(\phi_{k,v}^{0})^{q_{i,k,w,v}+\beta_{w}-1}(p_{w,v})^{q_{i,k,w,v}})d\Phi\nonumber \protect\\
= & (\prod_{i=1}^{I}\prod_{w=1}^{V}\prod_{v=1}^{V}(p_{w,v})^{\sum_{k=1}^{K}q_{i,k,w,v}})(\prod_{k=1}^{K}\frac{\prod_{i=1}^{I}Beta_{V}(\vec{\beta}+\sum_{i}\sum_{w}\vec{q}_{i,k,w})}{Beta_{V}(\vec{\beta})})\label{eq:SPDP-joint-derivation-step4-integral evaluation}
\end{alignat}
}}

\paragraph*{\textmd{Where in \ref{eq:SPDP-joint-derivation-step1-substitution}
we simply substituted values from equation \ref{eq:SPDP-joint-derivation-step0}.
In \ref{eq:SPDP-joint-derivation-step2-use-q} we re-wrote the terms
in $q_{ikwv}$. In \ref{eq:SPDP-joint-derivation-step3-rearranging}
we rearranged the equation. In \ref{eq:SPDP-joint-derivation-step4-integral evaluation}
we took out $p_{w,v}$ and evaluated the integral as multinomial probability
density function.}}

\paragraph*{\textmd{Finally, put all terms together:}}

\paragraph*{\textmd{
\begin{alignat*}{1}
 & p(\mathbf{W},\mathbf{Z},\mathbf{V},\mathbf{R}|\vec{\gamma},\vec{d},\vec{\alpha_{1:I}},\vec{\beta},\mathbf{P})\protect\\
= & \prod_{i=1}^{I}\prod_{d=1}^{D}\frac{Beta_{K}(\vec{\alpha}_{i}+\vec{n}_{idk:i,d})}{Beta_{K}(\vec{\alpha}_{i})}\protect\\
 & \bigg(\prod_{k=1}^{K}(\prod_{i=1}^{I}\frac{(\gamma|d)_{t_{ikw:i,k,.}}}{(d)_{n_{ikw:i,k,.}}}\prod_{w=1}^{V}\begin{pmatrix}n_{w}\protect\\
t_{w}
\end{pmatrix}^{-1}S_{t_{ikw:i,k,j},\gamma_{k}}^{n_{ikj:i,k,j}})\bigg)\protect\\
 & \bigg(\prod_{i=1}^{I}\prod_{w=1}^{V}\prod_{v=1}^{V}(p_{w,v})^{\sum_{k=1}^{K}q_{i,k,w,v}})(\prod_{k=1}^{K}\frac{\prod_{i=1}^{I}Beta_{V}(\vec{\beta}+\sum_{i}\sum_{w}\vec{q}_{i,k,w})}{Beta_{V}(\vec{\beta})}\bigg)
\end{alignat*}
}}

\paragraph*{\textmd{And simplify a bit:}}

\paragraph*{\textmd{
\begin{alignat*}{1}
 & p(\mathbf{W},\mathbf{Z},\mathbf{V},\mathbf{R}|\vec{\gamma},\vec{d},\vec{\alpha_{1:I}},\vec{\beta},\mathbf{P})\protect\\
= & (\prod_{i=1}^{I}\prod_{w=1}^{V}\prod_{v=1}^{V}(p_{w,v})^{\sum_{k=1}^{K}q_{i,k,w,v}})(\prod_{i=1}^{I}\prod_{d=1}^{D}\frac{Beta_{K}(\vec{\alpha}_{i}+\vec{n}_{idk:i,d})}{Beta_{K}(\vec{\alpha}_{i})})\protect\\
 & \bigg(\prod_{k=1}^{K}\big(\prod_{i=1}^{I}\frac{(\gamma|d)_{t_{ikw:i,k,.}}}{(d)_{n_{ikw:i,k,.}}}Beta_{V}(\vec{\beta}+\sum_{i}\sum_{w}\vec{q}_{i,k,w})\big)\bigg)\protect\\
 & \bigg(\prod_{k=1}^{K}\frac{1}{Beta_{V}(\vec{\beta})}\prod_{i=1}^{I}\prod_{w=1}^{V}\begin{pmatrix}n_{w}\protect\\
t_{w}
\end{pmatrix}^{-1}S_{t_{ikw:i,k,j},\gamma_{k}}^{n_{ikj:i,k,j}})\bigg)
\end{alignat*}
}}

\paragraph*{\textmd{Although the derivation of joint probability is a bit complicated
and requires a significant amount of work, the inference steps are
quite simple and efficient. In blocked Gibbs sampling, we sample multiple
variables together in one step, one element each. In our settings
the best choice is to sample one element each from $z,r,v$ together:
for each word $w$, if $r=0$, we can simply ignore the value of $v$
and compute the probability of $(z=k,r=0)$, otherwise we compute
the probability of $(z=k,r=1,v_{w}=v)$. Use the joint probability
computed above and the formula \ref{eq:joint-to-predictive}:}}

\paragraph*{\textmd{
\begin{alignat}{1}
 & p(w_{i,d,l}=w,z_{i,d,l}=k,r_{i,d,l}=0|\mbox{ the rest with word \ensuremath{w} removed\ensuremath{)}}\nonumber \protect\\
\propto & \frac{\alpha_{i,k}+n_{idk:i,d,k}}{b_{k}+m_{ikw:i,k\cdot}}\frac{m_{ikw:i,k,w}-t_{ikw:i,k,w}+1}{m_{ikw:i,k,w}+1}\frac{S_{t_{ikw:i,k,w},a_{k}}^{m_{ikw:i,k,w}+1}}{S_{t_{ikw:i,k,w},a_{k}}^{m_{ikw:i,k,w}}}\label{eq:SPDP-sampling-w-z-r0}
\end{alignat}
}}

\paragraph*{\textmd{
\begin{alignat}{1}
 & p(w_{i,d,l}=w,z_{i,d,l}=k,r_{i,d,l}=1,v_{i,k,w,t}=v|\mbox{ the rest with word \ensuremath{w} removed\ensuremath{)}}\nonumber \protect\\
\propto & p_{i,w,v}(\alpha_{i,k}+n_{idk:i,d,k})\frac{b_{k}+a_{k}t_{ikw:i,k,\cdot}}{b_{k}+m_{ikw:i,k\cdot}}\frac{t_{ikw:i,k,w}+1}{m_{ikw:i,k,w}+1}\frac{\gamma_{v}+\sum_{i}\sum_{w}q_{i,k,w,v}}{\sum_{v'}\gamma_{v'}+\sum_{i}\sum_{w}t_{ikw:i,k,w}}\frac{S_{t_{ikw:i,k,w}+1,a_{k}}^{m_{ikw:i,k,w}+1}}{S_{t_{ikw:i,k,w},a_{k}}^{m_{ikw:i,k,w}}}\label{eq:SPDP-sampling-w-z-r1}
\end{alignat}
}}

\paragraph*{\textmd{The algorithm is illustrated in Algorithm \ref{alg:SPDP-Full-Gibbs}.}}

\begin{algorithm}
\begin{algorithmic}[1]
\ForAll{$w_{i,d,l}$ in all documents in all groups}
	\State $k=z_{i,d,l}$, $w=w_{i,d,l}$
	\State Sample $r_{i,d,l}\sim Bernoulli(\frac{t_{ikw:i,k,w}}{m_{ikw:i,k,w}})$
	\State Decrement $n_{idk:i,d,k}$, $m_{ikw:i,k,w}$
	\If{$r_{i,d,l} \equiv 1$}
		\State Decrement $t_{i,k,w}$
		\State Sample $ t \sim Uniform(1,...,t_{ikw:i,k,w}+1)$
		\State Remove t-th element from linked list $\vec{v}_{ikw:i,k,w}$
		\State Decrement $q_{i,k,w,v}$
	\EndIf
	\For{each topic k in 1...K}
		\State Compute following proportionalities for each v in 1...V:
		\State $z_{i,d,l}=k$, $r_{i,d,l}=0$:
		\State \indent $\frac{\alpha_{i,k}+n_{idk:i,d,k}}{b_{k}+m_{ikw:i,k\cdot}}\frac{m_{ikw:i,k,w}-t_{ikw:i,k,w}+1}{m_{ikw:i,k,w}+1}\frac{S^{m_{ikw:i,k,w}+1}_{t_{ikw:i,k,w},a_{k}}}{S^{m_{ikw:i,k,w}}_{t_{ikw:i,k,w},a_{k}}}$
		\State $z_{i,d,l}=k$, $r_{i,d,l}=1$, $v_{i,k,w,t}=v$: 
\State \indent $p_{i,w,v}(\alpha_{i,k}+n_{idk:i,d,k})\frac{b_{k}+a_{k}t_{ikw:i,k,\cdot}}{b_{k}+m_{ikw:i,k\cdot}}\frac{t_{ikw:i,k,w}+1}{m_{ikw:i,k,w}+1}\frac{\gamma_{v}+\sum_{i}{\sum_{w}{q_{i,k,w,v}}}}{\sum_{v'}{\gamma_{v'}}+\sum_{i}{\sum_{w}{t_{ikw:i,k,w}}}}\frac{S^{m_{ikw:i,k,w}+1}_{t_{ikw:i,k,w}+1,a_{k}}}{S^{m_{ikw:i,k,w}}_{t_{ikw:i,k,w},a_{k}}}$
	\EndFor
	\State Jointly sample $z_{i,d,l}$, $r_{i,d,l}$, and $v_{i,k,w,t}$ according to proportionalities above.
	\State Increment $n_{idk:i,d,k}$, $m_{ikw:i,k,w}$
	\If{$r_{i,d,l} \equiv 1$}
		\State Increment $t_{i,k,w}$ and $q_{i,k,w,v}$
		\State Add $v$ to linked list $\vec{v}_{ikw:i,k,w}$
	\EndIf
\EndFor
\end{algorithmic}

\protect\caption{\label{alg:SPDP-Full-Gibbs}SPDP Full Gibbs Sampling}
\end{algorithm}

\paragraph*{\textmd{After convergence, the probability of a topic in a document
can be estimated in the same way as in LDA, by taking expectation
of the Dirichlet distribution:}}

\paragraph*{\textmd{
\begin{equation}
\tilde{\theta}_{i,d,k}=\frac{n_{idk:i,d,k}+\alpha_{k}}{\sum_{k=1}^{K}(n_{idk:i,d,k}+\alpha_{k})}\label{eq:spdp-topic-doc-estimate}
\end{equation}
}}

\paragraph*{\textmd{And the probability of a topic can be estimated by taking
the weighted sum of $\theta_{i,d,k}$ with document length as weight.
The probability of a word in a topic can be esitmated by using the
approximation of \ref{eq:spdp-phi0estimate} with $q_{ikwv:i,k,w,v}t_{ikw:i,k,w}$
replaced by the exact count $q_{ikwv:i,k,w,v}$, and the posterier
of PDP:}}

\paragraph*{\textmd{
\begin{eqnarray}
\tilde{\phi}_{k,v}^{0} & = & \frac{\gamma_{v}+\sum_{i}\sum_{w}q_{ikwv:i,k,w,v}}{\sum_{v}\gamma_{v}+\sum_{v}\sum_{i}\sum_{w}q_{ikwv:i,k,w,v}}\label{eq:spdp-word-topic-estimate}\protect\\
\tilde{\phi}_{k,w}^{i} & = & \frac{m_{ikw:i,k,w}-a_{k}t_{ikw:i,k,w}}{b_{k}+m_{ikw:i,k,\cdot}}+\frac{a_{k}t_{ikw:i,k,\cdot}}{b_{k}+m_{ikw:i,k,\cdot}}(\sum_{v}p_{w,v}^{i}\tilde{\phi}_{k,v}^{0})\label{eq:spdp-group-word-topic-estimate}
\end{eqnarray}
}}

\chapter{Problems and Solutions}

\section{SPDP Computational Time Analysis\label{sec:SPDP-Computational-Time}}

\subsection{Theoretical Running Time Analysis}

\paragraph*{\textmd{Assume there are $N$ words in total across all documents
in all groups. Let $K$ represents the total number of topics to be
extracted, and $V$ represents total number of words in the vocabulary.
Furthermore, assume Gibbs sampler only converges after $T$ iterations.
Algorithm \ref{alg:SPDP-Full-Gibbs} shows for each word, all topics
and all possible word-associations must be sampled. Although Stirling
numbers in the algorithm need to be computed recursively, it is possible
to cache almost all of them before the algorithm begins, allowing
constant-time retrieval complexity. To summarize, the theoretical
worst case complexity of algorithm \ref{alg:SPDP-Full-Gibbs} is $\Theta(TNKV)$.
As the transofrmation matrices $P^{i}$ define all possible word-associations
to be sampled in the nested inner loop of algorithm \ref{alg:SPDP-Full-Gibbs},
it is a significant influential factor of the total running time.
To reduce the worst case complexity, the transformation matrices $P^{i}$
need to be sparse. Once it is guaranteed that there is no more than
a small constant number of non-zero elements per row or column, the
average running time of \ref{alg:SPDP-Full-Gibbs} can be lowered
to $\Theta(TNK)$.}}

\subsection{Practical Issues}

\paragraph*{\textmd{In practice, a small collection of documents contains over
2000 documents in each group, consisting of over 500000 words, and
a vocabulary size over 20000. Number of topics we expect to retrieve
from such a collection is ranging between approximately 30 to 100,
and the collapsed Gibbs sampler requires about 2000 iterations to
converge. Multiplying these numbers together, we can see that to analyze
such a collection of documents may require at least a constant times
$10^{15}$ cycles if transformation matrix $P$ is not sparse. }}

\paragraph*{\textmd{Suppose the matrix is sparse and each row or column contains
at most $20$ non-zero entries. Compared to non-sparse transformation
matrix, the number of operations required is now reduced to a constant
times over $10^{12}$ cycles. The constant factor, which accounts
for memory instructions and arithmetic computations at line 14,16
in algorithm \ref{alg:SPDP-Full-Gibbs}, could be very large. Complicated
data structure could make the issue even worse, e.g. word-association
linked lists $v_{ikw}$, the sparse matrices $P^{i}$ , and the sparse
counts $q_{ikwv}$. Accessing or modifying them require frequent pointer
chasing operations and a tradeoff between memory usage and access
efficiency. If these variables are implemented as full-size arrays
instead, the amount of memory being allocated would exceed a constant
times $V^{2}$. In practice, we observed that the amount of memory
required by full array implementation could reach tenth of gigabytes
on a small document collection. This magnitude of memory consumption
is far more than what is offered in consumer-grade computers, severely
limits the scalability of this algorithm. }}

\paragraph*{\textmd{In the experiments, 2000 Gibbs iterations over a small collection
of documents took several hours to complete, even when the transformation
matrices $P^{i}$ are set to identity matrices. }}

\subsection{Parallelization Issues}

\paragraph*{\textmd{As explained in Chapter 1, algorithm \ref{alg:SPDP-Full-Gibbs}
is under the framework of the collapsed Gibbs sampler. Mathematically,
the collapsed Gibbs sampler requires all words to be sampled sequencially.
The state of the collapsed Gibbs sampler (in our case they are the
vector $p(\mathbf{W,Z|}\mbox{ the rest\ensuremath{)}}$ at each step
when a word is sampled) form a Markov chain. In other words, the probability
distribution of topic and word-association for the sample of the current
word is dependent on the sample generated for previous words. Sampling
multiple words in parallel with old state information breaks the rule
of dependency. Consequently, it is not mathematically guaranteed that
the collapsed Gibbs sampler running in parallel would converge.}}

\section{The Goal }

\paragraph*{\textmd{We are looking for a solution that not only addresses the
above issues, but also satisfies a set of properties. It is preferred
that the solution has flexible memory usage, is able to process large
collection of documents without being slowed down by memory access;
the solution should take advantage of parallelism, so it can be made
scalable enough to be executed on multiple devices. The solution does
not necessarily need to be exact. A good approximation which sacrifice
a bit of accuracy but greatly improve running speed is good enough
for practical purposes. To keep the balances between accuracy and
speed, we also need a set of performance and quality requirements
to measure the fitness of my approximation. They could include:}}
\begin{itemize}
\item Scalability
\item Perplexity: a measure of fitness of the model to the data
\item Topic quality and intepretability: how good the topics are in terms
of human understanding
\end{itemize}

\paragraph*{\textmd{In addition, it will also be discussed why the popular PMI-score
is not an appropriate measure for performance in our problem. Note
also since the goal is not to measure the quality of topics but to
measure how well the approximation matches the original (sequential)
SPDP, perplexity is adequate enough. }}

\subsection{Scalability}

\paragraph*{\textmd{As the algorithm is expected to process hundreds to millions
of documents in real world applications, it becomes necessary to design
a memory access architecture that allows efficient access to both
dynamic count varaibles $t,n,m,q,v$, and static constant variables
$P^{i}$ (transformation matrices), $S_{m}^{n}$ (Stirling numbers).
When the amount of memory required to store these variables becomes
too large, these variables can no longer be physically stored in main
memory. As Algorithm \ref{alg:SPDP-Full-Gibbs} accesses the variable
$n$ and the transformation matrices $P^{i}$ in a linear, consecutive
pattern, they are the easiest ones to cache. In comparison, there
is no common pattern in how words and topics should appear across
a document, causing more or less random access to count variables
$t,m,q,v$.}}

\paragraph*{\textmd{A good memory access architecture should divide variables
into multiple regions and multiple levels of hierarchy, putting the
current demand to priority, and take historical access frequency into
consideration. In a distributed or parallel system, the hierarchy
can be global memory, device memory, local memory, constant memory,
cache, and buffers. Should the algorithm be executed in a distributed
or parallel manner, redundant copies are unavoidably created across
multiple levels in the memory hierarchy. }}

\paragraph*{\textmd{Suppose the current memory consumption of the algorithm is
$M$ and we are running a distributed or parallel version of this
algorithm on $S$ devices. A naive architecture that creates a redundant
copy on every device require $O(MS)$ amount of memory. This is problematic
because $M$ could exceed the total amount of memory available on
each device. A good architecture should limit the amount of redundancy,
ideally making the total memory consumption $O(M)$, independent to
$S$, or indepedent except for a few variables that only consume a
small amount of memory.}}

\subsection{Topic Performance Measure}

\paragraph*{\textmd{Perplexity and pointwise mutual information score (PMI score)
based on Wikipedia corpus\cite{conf/naacl/NewmanLGB10} are the two
most popular measures used in the research field to judge the quality
of generated topic models. }}
\begin{itemize}
\item PMI score based on Wikipedia corpus calculates pointwise relevancy
of top ten words in each topic with respect to frequency of co-occurence
between corresponding words in the Wikipedia corpus.
\item Perplexity measures how well the generated model fits the test data. 
\end{itemize}

\paragraph*{\textmd{A good algorithm is expected to give results with high PMI
score and low perplexity. In practice, we found in many situations
the PMI score could be unreliable, as we will soon illustrate. We
believe it is also important to manually check the intepretability
of generated topics, and whether these topics make sense to humans.
This is a time comsuming process highly subjective to human knowledge
and intepretation, but this guarantees we get a sensible result.}}

\subsubsection{PMI-Score Based on Wikipedia Corpus}

\paragraph*{\textmd{Out of many evaluation methods proposed in \cite{conf/naacl/NewmanLGB10},
the PMI-score based on the Wikipedia corpus is the consistent best
performer with respect to intrinsic semantic quality of learned topics. }}

\paragraph*{\textmd{In \cite{chen2012spdp} Chen et al.$\ $defined PMI-score
as $PMIScore(\vec{w})=\frac{1}{45}\sum_{i<j}PMI(w_{i},w_{j})$, where
$i,j\in\{1,...,10\}$, $PMI(w_{i},w_{j})=\log\frac{P(w_{i},w_{j})}{P(w_{i})P(w_{j})}$,
and $P(w_{i},w_{j})$ is defined as word $w_{i}$ and $w_{j}$ appears
in the same 10 word window, $P(w_{i})$ and $P(w_{j})$ are the probability
of occurrence of word $w_{i},w_{j}$ respectively, estimated from
word frequency as in the April 2011 Wikipedia dump. Overall PMI-score
is computed by summing PMI-score of top ten words in each topic over
all topics. }}

\paragraph*{\textmd{However, the PMI-score measure in SPDP is susceptable to
the influence of transformation matrices $P^{i}$. The entries of
transformation matrices $P^{i}$ determine word-association, make
associated words more likely to appear in the same topic. It is possible
to manipulate entries of $P^{i}$ to artificially increase the PMI-score.
Furthermore, there are many situations that the PMI-score cannot accurately
judge the semantic relevancy between two words. When the collection
of documents is focused on a specific area, very often there are technical
phrases such as ``machine learning'', ``group theory'', ``topic
quality'' appear everywhere across the documents. These phrases don't
make sense to people who are not specialized in machine learning.
They may appear very infrequently in Wikipedia except only in very
few technical articles. In contrast individual terms ``machine''
``learning'' ``group'' ``theory'' ``topic'' ``quality''
may appear very frequently across everywhere in Wikipedia. Should
we analyze a machine learning journal and produce a topic consists
words such as ``machine learning topic modelling score function ...'',
we would obtain a low PMI-score, indicating poor topic quality, which
is apparently not the case.}}

\paragraph*{\textmd{Therefore, we choose not to use PMI-score to measure the
quality of the result.}}

\subsubsection{Perplexity\label{sub:Perplexity}}

\paragraph*{\textmd{Perplexity represents a scaled likelihood of test data given
the parameters trained by the training data. When measuring the quality
of a topic model, perplexity is usually defined as: 
\begin{eqnarray*}
P(\mathbf{W}|\mbox{ the rest}) & = & \prod_{i=1}^{I}\prod_{d=1}^{D_{i}}p(\vec{w}_{i,d}|\mbox{ the rest})^{-\frac{1}{N}}\protect\\
 & = & exp(-\frac{\sum_{i=1}^{I}\sum_{d=1}^{D}log(p(\vec{w}_{i,d}|\mbox{ the rest\ensuremath{))}}}{\sum_{i=1}^{I}\sum_{d=1}^{D}N_{i,d}})
\end{eqnarray*}
}}

\paragraph*{\textmd{Where $\mathbf{W}$ represents all words among all test documents
in all groups. Test documents are documents held out in each group
during training phase. $D_{i}$ is the number of test documents in
group $i$, $N$ is total number of words, $N_{i,d}$ is total number
of words in test document $d$ group $i$, $I$ is total number of
groups, $\vec{w}_{d}$ is the words in test document $d$ of group
$i$. Similar to the definition in \cite{oai:fraunhofer.de:N-101883},
we define $p(\vec{w}_{i,d}|\mbox{ the rest\ensuremath{)}}$ as: 
\begin{eqnarray*}
p(\vec{w}_{i,d}|\mbox{ the rest\ensuremath{)}} & = & \prod_{n=1}^{N_{i,d}}\sum_{k=1}^{K}p(w_{n}=t|z_{i,d,l}=k)p(z_{i,d,l}=k)\protect\\
 & = & \prod_{n=1}^{N_{d}}\sum_{k=1}^{K}(\tilde{\phi}_{k,w}^{i}\tilde{\theta}_{i,d,k})^{n_{i,d,w}}
\end{eqnarray*}
}}

\paragraph*{\textmd{Where $N_{i,d}$ is length of test document $i,d$, and $n_{i,d,w}$
is number of times word $w$ appears in test document $i,d$. The
variables $\tilde{\phi}_{k,w}^{i}\mbox{, }\tilde{\theta}_{i,d,k}$
are as defined in Equations \ref{eq:spdp-topic-doc-estimate} and
\ref{eq:spdp-word-topic-estimate}.}}

\subsubsection{Topic Quality and Intepretability}

\paragraph*{\textmd{Good performance in perplexity and PMI score is not sufficient
to indicate a good topic model. If the produced topics do not deliver
coherent intepretable information to humans, they would not be useful
in practice, even when they perform well in both PMI-score and perplexity. }}

\paragraph*{\textmd{For instance, the topic ``barack obama apple iphone ipad
health insurance'' could have high PMI score because some word-pairs
in this topic have frequent appearences in the Wikipedia corpus ,but
apparently this is not a good topic because it is a mixture of three
topics. Similarly, low perplexity only shows that the topic model
has good ability to predict words in test documents, which does not
neccesarily mean the topics are of good quality. The problem is best
illustrated with an example in american polital blog document collection,
where a topic model simply puts highest weight into the most frequent
words such as ``obama republican democrats said just'', or simply
computes word frequency and evenly spread word across all topics.
The perplexity of such topic model can be even lower than good topic
models but they do not give any useful topic information. }}

\paragraph*{\textmd{In differential topic modelling, we are also interested in
the coherency of topics shared among different groups. One important
distinction of SPDP is its ability to find subtle differences in topics
shared among multiple groups. Rather than relying on a single measure
produced by an automated algorithm, this ability is better to be judged
by a human as it involves understanding the background knowledge and
complicated semantic analysis. }}

\section{The Innovation: Speeding Up SPDP }

\paragraph*{\textmd{In this section, I propose a number of ways to improve the
running speed of SPDP. First, I give a discussion of a basic trivial
parallelization on line 11-17 of Algorithm \ref{alg:SPDP-Full-Gibbs}.
I show that when the number of topics and the number of effective
entries in the transformation matrices $P^{i}$ are small, such parallelization
may raise significant thread-creation and synchronization overhead,
contrary to what people would expect in the first place. I address
the challenge that exact collapsed Gibbs sampling require words to
be sequencially sampled, hence any parallelization may incur considerable
risk in convergence and loss of accuracy. I give a discussion on possible
ways to overcome this obstacle, then propose a parallel approximation
that not only can be justified to work in theory, but also can be
implemented and tested to work reasonably well in practice. }}

\paragraph*{\textmd{In addition to this, I also propose a method to significantly
increase the accuracy of the approximation algorithm. After this,
I introduce some existing state-of-art distributed and parallel models
for LDA. I argue that the same model can be applied to SPDP after
modification. Finally, I combine all these ideas together, and create
an all-in-one multi-GPU distributed parallel approximation of SPDP.
I show that this all-in-one model not only works in theory, but also
address the practical issues illustrated previously in the thesis.}}

\paragraph*{\textmd{Throughout this section, I stick to one major principle:
we are designing things for real world application. For this reason,
I make notes on how these proposals can be adopted into multiple architectures,
namely the conventional CPU architecture, and the novel GPU architecture,
as the title of the thesis suggests. Of these two architectures, the
primary one I focus on is commercially available consumer-grade GPU,
especially the multi-GPU distributive architecture. I also give a
brief discussion about the performance of the algorithms on traditional
CPU architecture wherever it is applicable. Details on these two different
architectures at framework and hardware level are left to section
\ref{sec:Implmentation}.}}

\subsection{Distributed Parallelization Proposals}

\subsubsection{Basic Parallelism: Over Topics and Word-Associations\label{sub:Basic-Parallelization:-Over}}

\paragraph*{\textmd{Line 11-17 of algorithm \ref{alg:SPDP-Full-Gibbs} contain
only independent operations. Assuming sparse transformation matrices
$P^{i}$ are used, $K\times S$ threads can be issued in parallel
to compute sampling probabilities of $(k,v)$ pairs, where $S$ is
the maximum number of non-zero entries in each row and column of $P^{i}$,
$k$ represents a topic, and $v$ represents a word-association. In
most systems (especially on a CPU architecture), thread creation is
a very expensive operation. The cost of creating a thread can be greater
than the benefit of having multiple threads computing these probabilities
in parallel. }}

\paragraph*{\textmd{Typically the cost of creating a thread ranges from $10^{5}$
to $10^{7}$ CPU cycles, depending on system architecture. Regardless,
the value is far greater than the number of cycles required to compute
the probabilities for each $(k,v)$ pair, which is in between $10^{2}$
to $10^{3}$ cycles. Unless the value of $K\times S$ is in the order
$10^{4}$ and the system has an architecture that supports fast hardware
thread creation, it is not a good idea to create threads inside the
loop around line 11-17 of Algorithm \ref{alg:SPDP-Full-Gibbs}.}}

\paragraph*{\textmd{The next thing worth trying is to have all threads created
before the execution of the algorithm, and pre-allocate the threads
to $(k,v)$ pairs. Multiple synchronization points are required at
different parts of the algorithm. Parts of the algorithm that cannot
be parallelized (everything except line 11-17) must be designated
to a single thread, while the rest of the threads are kept idle as
they wait for this single thread to finish.}}

\paragraph*{\textmd{The effectiveness of this method is not as simple as it looks
like. How much does it cost to synchronize all threads at multiple
synchronization points? How many threads should be created so the
algorithm can achieve best performance? The first question cannot
be answered without doing experiments. The answer to the latter question
is simple if we only consider an architecture based on CPUs - simply
create as many as the CPU could support at hardware level. However,
as we look further into GPU architecture, it can be quite complicated
to determine the number of threads that should be created at this
level to achieve best performance. }}

\subsubsection{Parallel Word Sampling\label{sub:Parallelization-Over-Sampling}}

\paragraph*{\textmd{We mentioned in previous sections that any parallel word
sampling breaks the mathematical rule required by the collapsed Gibbs
sampler. The risk of divergence and loss of accuracy are not avoidable,
but with appropriate parallelization methods, the loss can be minimized.}}

\paragraph*{\textmd{Let us look into the convergence process of the collapsed
Gibbs sampling in SPDP. In each iteration, the topic of a particular
word is sampled based on the current counts with respect to all words
except the word itself. Frequent topics and words have their counts
accummulated quickly, while uncommon topics and words also lose their
counts gradually. As the algorithm progresses through many iterations,
changes are slowed down and counts are stablized, until they converge
to a stationary state. This behavior is similar to many chaotic systems,
where convergence is not dependent on initial values, and the final
stationary state is not sensitive to small external change in positions
of each body during early phase of convergence. This suggests that
a relatively small error in counts probably does not matter much to
the overall accuracy. The convergence of the collapsed Gibbs sampler
is determined by statistics on words, not the dynamic topic assignments,
so a small error in topic assignments should not affect the overall
convergence.}}

\paragraph*{\textmd{Suppose there are $N$ parallel threads sampling words on
$P$ processors in parallel. If all count varaibles are stored in
global memory, and all threads are allowed to modify them directly,
a rather large amount of inconsistency would be introduced, causing
high risk of divergence and significantly loss in accuracy. }}
\begin{enumerate}
\item Counts relevant to all words being sampled concurrently in $P$ processors
are removed at the beginning of Algorithm \ref{alg:SPDP-Full-Gibbs},
provide incorrect information to all threads at the beginning of execution.
The large discrepancy in count variables can be critical since the
probability formulas in Algorithm \ref{alg:SPDP-Full-Gibbs} are computed
based on the assumption that only one word is removed. 
\item A second level of inconsistency is introduced while multiple threads
modify the same count variable at the same time, especially if modification
happens while some threads are still in the process of computing topic
and word-association probabilities.
\item If spinlocks are used to prevent conflict in accessing the same count
variable, a huge amount of execution time has to be wasted on waiting
for memory access. Spinlocks could also introduce a large variance
on execution time, adding large extra cost to synchronization operations,
which can be fatal if $P$ gets large. 
\end{enumerate}

\paragraph*{\textmd{On the other hand if a local copy of count variables is created
for each thread, and only get synchronized after each sampling thread
finishes execution, the lack of between-thread communication would
inevitably cause count variables in all threads to be delayed by $N$
steps. This is a critical issue when $N$ is large, the collapsed
Gibbs sampler cannot make use of any information that is too old.}}

\paragraph*{\textmd{To minimize the error while keeping execution speed as fast
as possible, and the degree of parallelism as high as possible, we
need a mechanism to keep count variables in each thread up-to-date
as much as possible, with minimum amount of conflict in memory access,
and smallest amount of global synchronization pointw. The solution
has to be sought separately for GPU and CPU architectures. On a CPU
architecture, the number of processors per device (machine) is typically
small. All processors share a single global memory with very low access
latency. Each processor has a large amount of cache, advanced instruction
sets, and great arithmetic processing power. On the GPU architecture,
global synchronization is not possible. Memory resource is scarce,
access is differentiated into multiple levels in hierarchy. The number
of processors on a GPU is very large but each processor is much slower
and much less advanced than a CPU processor. Massive parallelization
is effective under a GPU architecture, but not as effective on a CPU
architecture. Memory resource is not much of an issue under a CPU
architecture, but it has to be carefully measured under a GPU architecture. }}

\paragraph*{\textmd{The solution we propose is to create a minimum local copy
of count variables while keeping global count variables updated at
the same time. The process is as follows: }}
\begin{enumerate}
\item At the start of the algorithm, a thread is created for each word in
all documents.
\item At the beginning of each thread a local copy of count variables relevant
to this word is created. 
\item Following this, the global count variables relevant to this word is
immediately updated as in line 2-10 in Algorithm \ref{alg:SPDP-Full-Gibbs}. 
\item After a new sample is drawn, the global count variable is immediately
updated so any subsequent reading to this global counter is up-to-date. 
\end{enumerate}

\paragraph*{\textmd{Since local count variables are created at approximately
the same time across all threads before computing probabilities, the
second level of inconsistency we mentioned previously no longer exists.
Spinlocks or semaphores on global variables are totally optional.
Error may accumulate if multiple threads modify the same count variable
at the same time, but the chance of having multiple threads in the
same batch accessing the same word and same topic is very low. Inconsistency
between count variables and auxillary variables (e.g. sum of some
count variable in some dimension) can be manually corrected regularly
during the iteration.}}

\paragraph*{\textmd{Access to each variable is not accompanied with a lock, therefore
it is safe to have the number of threads $N$ far larger than number
of available processors $P$, and have threads scheduled in batches
to fit $P$ processors. Not only subsequent batches can take advantage
of updates in previous batches, multiple batches can also be pipelined
on both CPU and GPU to take advantage of SIMD features on the processor
or multiprocessors. Most parallel programming frameworks and hardware
architectures can do this implicitly, if such features are supported.}}

\paragraph*{\textmd{When pipelining is used, the time between creating local
copy of count variables and updating global variables should be reduced
to minimal, so the update could propagate faster. Because the only
major operations between these two steps are to comput probabilities
for each topic and word-association pair, an additional level of parallelization
can be combined with the solution to minimize the delay. }}

\paragraph*{\textmd{Similar to what we discussed in Section \ref{sub:Basic-Parallelization:-Over},
for each word we assign a group of at most $K\times S$ threads to
compute the probabilities of topic and word-association pairs. We
call this a workgroup}\label{par:workgroup}.\textmd{ Denote the number
of threads we use to compute probabilities of topic and word-association
pairs by $Q$, which divides $K\times S$. Since we only have $P$
processors, the total number of workgroups being executed concurrently
on hardware is no more than $\lceil\frac{P}{Q}\rceil$. As we discussed,
it is safe to schedule more threads than number of available processors,
which also implies it is safe to schedule more workgroups than number
of available processors. Therefore, we can let the total number of
workgroups equal to total number of words across all documents in
all groups, and create them all at the beginning of each Gibbs iteration.
As it will be soon revealed in Section \ref{sec:Implmentation}, this
structure perfectly fits GPU architecture. }}

\subsubsection{Parallelism With Improved Accuracy: Word Order Rearrangement\label{sub:Parallelization-With-Better}}

\paragraph*{\textmd{When multiple words in the same document are sampled in parallel,
the risk of conflicting memory access to count variable $n$ is higher.
To reduce the risk, we can change the order of picking words when
we sample them. Since our model is unigram, we are allowed to sample
words among all documents in whatever order we wish. At one time,
we only want one word from each document sampled in parallel. Since
the number of processors is limited, we can simply rearrange the order
of words before scheduling them to processors. Words in the same document
should be kept apart as much as possible, so at one time each batch
of workgroups being concurrently processed mostly consists of only
words from different documents. }}

\paragraph*{\textmd{This can be done as follows. }}
\begin{enumerate}
\item Before execution of the algorithm, create an empty array $A$ to store
the words. 
\item Create a variable $w_{ind}\gets0$ to store the current word index. 
\item Create $D_{max}\gets\max\{L_{i,d}\mbox{ }\forall i=1..I,d=1..D_{i}\}$,
representing the length of the longest document. $L_{i,d}$ is the
length of document $d$ in group $i$.
\item Until $w_{ind}\ge D_{max}$, (randomly) pick up word at index $w_{ind}$
from each document $d$ in each group $i$ which $L_{i,d}>w_{ind}$,
and append to array $A$. Increment $w_{ind}$ each time all documents
and all groups are traversed.
\end{enumerate}

\paragraph*{\textmd{This technique is applicable to both CPU and GPU architecture
as it does not change anything inside original algorithm.}}

\subsubsection{Traditional Distributed Model: Dividing Documents}

\paragraph*{\textmd{Many distributed models have already been proposed for LDA.
The most related and influencial ones are AD-LDA (Approximate Distributed
LDA) proposed by Newman et al. in \cite{journals/jmlr/NewmanASW09},
and an improved version proposed by Smola and Narayanamurthy in \cite{journals/pvldb/SmolaN10}. }}

\paragraph*{\textmd{Newman's model distributes $D$ documents and counts related
to these $D$ documents into $P$ processors (not necessarily on the
same machine), and only synchronize count variables after each iteration
of the collapsed Gibbs sampling. Newman argued this model is a good
approximation because the sampling process on multiple processors
barely touch the same word and same topic, hence error accumulated
in count variables is insignificant. }}

\paragraph*{\textmd{Smola and Narayanamurthy's made an improvement over Newman's
distributed framework. They proposed an architecture which assigns
a dedicated processor to update and synchronize count variables globally
and locally for each thread, so remaining threads can keep on with
sampling and never get interrupted. Since updates are more frequent,
count variables are more up-to-date, hence further reducing the amount
of accumulated error. }}

\paragraph*{\textmd{SPDP has a much more complicated structure than LDA. Effectively,
LDA only use two types of count variable: $n_{kw}$ and $n_{dk}$.
In contrast, SPDP has $n,t,m,v,q$. Both types of count variable $n_{kw}$
and $n_{dk}$ in LDA can be accurately re-generated directly from
topic assignments, whereas in SPDP count variables $t,v,q$ cannot
be re-generated or verified. Newman's distributed framework can be
applied to SPDP, but keep count variables $t,v,q$ approximately correct
can be a challenge. }}

\paragraph*{\textmd{To enhance the model's ability to keep count variables approximately
correct, two levels of error correction should be implemented. One
level is inside each processor (or GPU), and the other is a global
correction done without massive parallelism. Neither Newman nor Smola
and Narayanamurthy discussed GPU architecture in their work. Rather
than dividing documents to multiple processors, they can be divided
to multiple GPUs under a GPU architecture. Smola and Narayanamurthy's
improvement over Newman's model can only be applied if communication
between GPUs is possible. However, the only two GPU programming frameworks
available, namely OpenCL and CUDA, do not provide any method to achieve
this. The OpenCL Specification \cite{opencl08} also explicitly stated
that the behavior of modifying the content of one memory object while
another device is accessing the same memory object is undefined. However,
through experiments we found at the hardware level NVIDIA and AMD
both support implicit weak synchronization across multiple GPUs, though
such synchronization is explicitly declared ``unspecified'' in their
official guide.}}

\subsection{All-in-one: Putting Everything Together\label{sub:All-in-one:-Putting-everything}}

\FloatBarrier

\paragraph*{\textmd{The previous ideas can be combined into a single three-layer
distributed parallel model. We call the processor which initiates
Algorithm \ref{alg:SPDP-Full-Gibbs} as the host. The combined model
is as follows:}}
\begin{enumerate}
\item At the beginning of each iteration of collapsed Gibbs sampling, rearrange
words to maximize the distance between words in the same document,
as illustrated in \ref{sub:Parallelization-With-Better}. 
\item Randomly divide all documents in all groups to $G$ devices (a device
can either be a machine, a processor, or a GPU). Distribute re-arranged
words to corresponding devices. By so doing, each device should have
approximately an equal amount of load.
\item For each device $g$, dispatch count variables relevant to words and
documents assigned to $g$ to the device.
\item Sample words in parallel on each device with the proposed two-level
framework described in Section \ref{sub:Parallelization-With-Better}.
Depending on values of $K$ (the number of topics) and $S$ (the number
of non-zero entries per row and column in transformation matrix $P^{i}$),
degree of parallelism should be tailored to hardware architecture
and specification.
\end{enumerate}

\paragraph*{\textmd{Under the CPU architecture, the degree of parallelism is
based on the number of processors available and the maximum number
of threads supported on each device. The process is basically as same
as what is described in \ref{sub:Parallelization-With-Better}.}}

\paragraph*{\textmd{Under the GPU architecture, hardware vendors specify the
number of computing units available on device, as well as the optimal
workgroup size. The process described in Section \ref{sub:Parallelization-With-Better}
needs to be adjusted to fit these specifications. When GPU memory
is insufficient to store all count variables and words at once, they
have to be scheduled in multiple waves. This could be both beneficial
and problematic. Each time a wave of words is sampled, count variables
and auxillary variables can be validated and corrected efficiently
on host, before the next wave is dispatched for sampling. On the other
hand memory transfer between global memory on host and GPU memory
on device has much higher latency than internal memory operations.
The optimal number of waves can only be found through experiments
and trial and error. The underlying pricinple is that the time spent
on memory transfer must be far less than time spent on GPU kernel
execution.}}

\paragraph*{\textmd{After a wave of words is sampled on the device, new samples
and new count variables are read back to host. As words are sampled
in parallel, errors are inevitably accumulated in count variables,
and they must be validated and corrected before doing anything else.
An mentioned before, count variables $n,m$ can be re-generated from
topic assignments. If synchronization between devices is done in-place,
other count variables can be simply read back from an arbitrary device,
and auxillary variables can be corrected based on values of count
variables. Otherwise, other count variables have to be corrected by
adding up the differences between their original values and the new
values returned from all devices. However, as we discussed in the
last section, count variables can change gradually at early stages
of collapsed Gibbs sampling. Under the distributed framework, all
devices operate independently. The sum of all differences is an exaggeration
of the actual amount of change in one variable. We are yet to find
a well-justified formula to update count variables based on the differences,
and in practice we found this method does not work well. Instead,
through the experiments we rely on the implicit synchronization of
GPU memory between multiple devices sharing the same context on the
same platform. Although the official hardware vendor programming guide
does not guarantee any update to one variable is immeidately synchronized
to all devices, in practice this method works very well. We do not
really need updates to be reflected to all devices, as long as the
delay can be tolerated. }}

\paragraph*{\textmd{\label{par:When-implicit-between-device}When implicit between-device
synchronization is not reliable enough, it is possible to create multiple
duplicates of the training data to mitigate the loss of accuracy.
Random errors with respect to one word in particular document can
be averaged out through its duplications, effectively prevented them
to be accumulated toward one direction after a few iterations. This
technique is also useful when the amount of data available is too
small to be distributed to multiple devices. Nonetheless, duplication
of the training data should be considered as a last resort only when
the quality of the result is highly inaccurate, and the number of
duplicates should be far smaller than number of devices. In practice,
I did not encounter any situation which the training data have to
be duplicated to get sensable result, though I found by duplicating
the training data once, the quality of the result can always be slightly
improved.}}

\begin{figure}
\includegraphics[angle=90,width=1\textwidth,height=1\textheight,keepaspectratio]{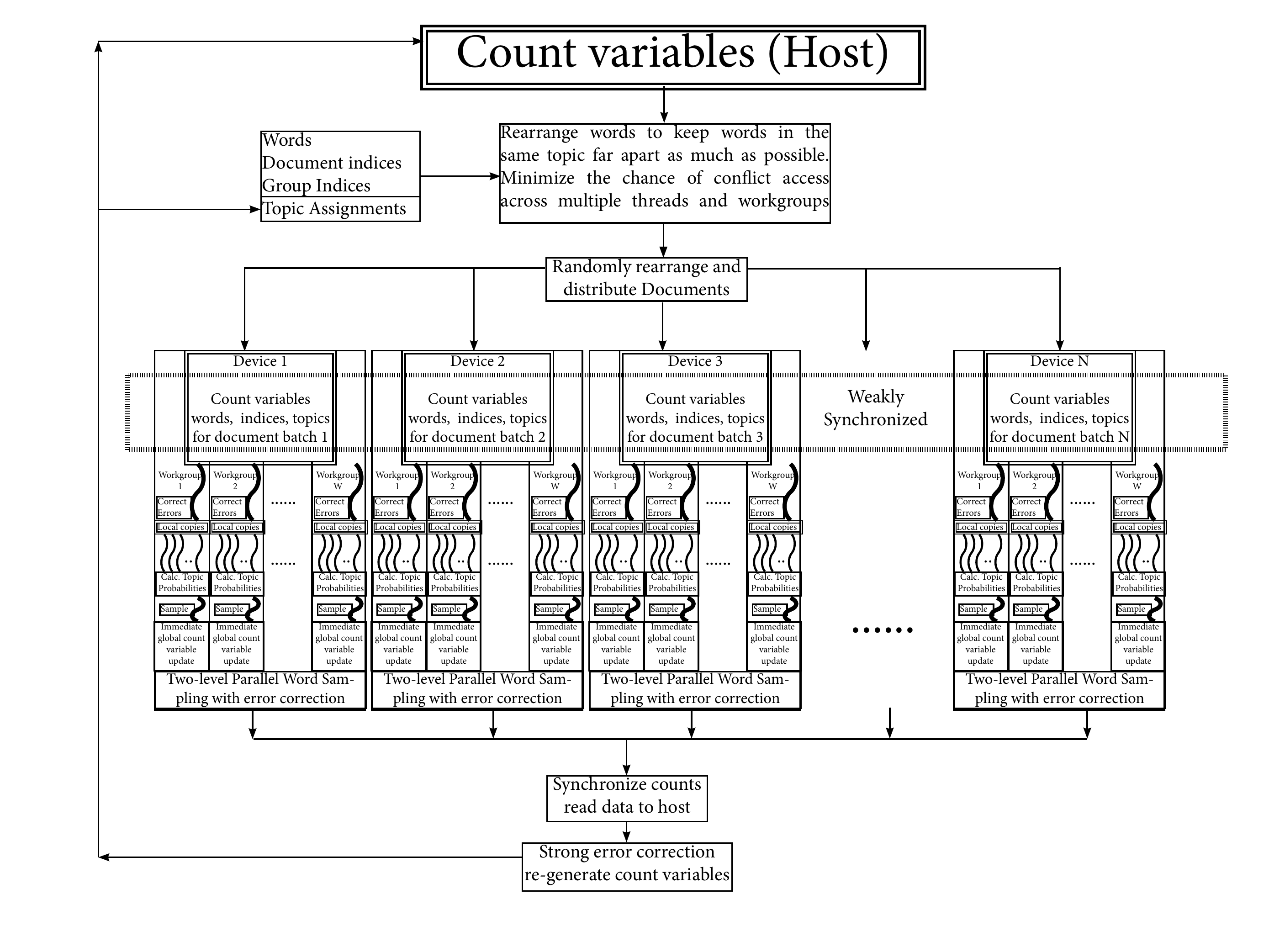}

\protect\caption{\label{fig:SPDP-Multi-GPU-Distributed}SPDP Multi-GPU Distributed
Parallel Sampling Proposal}

\end{figure}

\paragraph*{\textmd{Figure \ref{fig:SPDP-Multi-GPU-Distributed} illustrates
the big picture of my proposal.}}

\FloatBarrier

\section{Implementation\label{sec:Implmentation}}

\paragraph*{\textmd{Previously the issues of SPDP, along with the solutions,
have been discussed. A generalized framework for speeding up has been
created, a few distributed and parallel techniques have been proposed,
and their fitness under both GPU and CPU architectures has been examined.
This section is exclusively focused on GPU architecture and implmentation
issues. I show that for my algorithm, GPU architecture has superior
performance and superior cost-efficiency when compared to CPU architecture.
A discussion is given on programming frameworks, namely OpenCL and
CUDA, the two dominating GPU programming frameworks, and I argue that
OpenCL is a better framework to start with. The architectural differences
between two major GPU vendors, namely NVIDIA and AMD, are discussed,
and a conclusion has been made that for my algorithm there is no evidence
which one is better than another before any optimization. Practical
issues such as memory constraints, memory transfer rate, clock cycle
rate, work group size, and number of processors, are discussed in
detail. An implementation is given at the end of the section, but
it is not optimized for any particular type of GPU.}}

\paragraph*{\textmd{For simplicity it is assumed that across the whole section
the transformation matrices $P^{i}$ are identity matrices. As a consequence,
linked lists $v$ are reduced to arrays storing the number of associations,
as the only possible word-association is the word itself, and count
variable $q$ is always equal to the corresponding $t$ count. The
lowest level of parallelism, the topics and word-associations, has
effectively reduced to two types of possibilities for each topic:
topics with no word associations, and topics with identical word-association.
This is done for two reasons:}}
\begin{enumerate}
\item OpenCL is a subset of C99 language with a small number of extensions.
Non-primitive data structures like linked list are not easy to be
natively implemented on GPU, and even harder to be implemented in
a way to support parallel accesss. This is not the primary interest
of this thesis, so we decide to leave this out for now. At the end
of Chapter 4, a brief discussion is given on this issue. 
\item An effective error correction method for $v,q$ counts when transformation
matrices $P^{i}$ are not identity matrices has not been found. A
few proposals have been made but none has been thoroughly tested and
analyzed. To avoid confusing the readers, I decided not to present
these partial solutions. 
\end{enumerate}

\subsection{Parallelization Framework Comparisons}

\subsubsection{CPU v.s. GPU}

\paragraph*{\textmd{The key winning factors for GPU over CPU are overall higher
computing power and higher energy efficiency. This is a combined result
from higher memory bandwidth and a higher number of simple processors
at lower operating frequency. Table \ref{tab:CPU-v.s-GPU} shows a
comparison between a modern CPU and GPU, both designed by Advanced
Micro Devices, Inc. (AMD), available on the consumer market at approximately
the same price ( \$100USD per piece, as of 30 September 2012). A comparison
between NVIDIA GPU and Intel CPU is given in Table \ref{tab:Intel-CPU-v.s-NVIDIA-GPU}
and \ref{tab:Intel-CPU-v.s-NVIDIA-GPU-mem}.}}

\begin{table}
\begin{tabular*}{1\textwidth}{@{\extracolsep{\fill}}|>{\centering}p{0.25\textwidth}|>{\centering}p{0.25\textwidth}|>{\centering}p{0.25\textwidth}|>{\centering}p{0.15\textwidth}|}
\hline 
\multirow{1}{0.25\textwidth}{} & CPU & GPU & Winner Ratio\tabularnewline
\hline 
Example Device & AMD Phenom\texttrademark{} II X4 & AMD Radeon\texttrademark{} HD 7770 & \tabularnewline
\hline 
Core Frequency & \cellcolor{cyan}2800 MHz & 1 GHz & 3 X\tabularnewline
\hline 
Compute Units & 4 & \cellcolor{cyan}10 & 2.5 X\tabularnewline
\hline 
Approx. Power & 95W & \cellcolor{cyan}80W & 1.2 X\tabularnewline
\hline 
Approx. Power/Compute Unit & 19W & \cellcolor{cyan}8W & 2.4 X\tabularnewline
\hline 
Peak Single-Precision Billion Floating-Point Ops/Sec & 90 & \cellcolor{cyan}1280 & 14 X\tabularnewline
\hline 
Approx GFLOPS/Watt & 0.9 & \cellcolor{cyan}16 & 18 X\tabularnewline
\hline 
Max In-flight HW Threads & 4 & \cellcolor{cyan}25600 & 6400 X\tabularnewline
\hline 
Simultaneous Executing Threads & 4 & \cellcolor{cyan}640 & 160 X\tabularnewline
\hline 
Memory Bandwidth & 26 GB/s & \cellcolor{cyan}72 GB/s & 2.8 X\tabularnewline
\hline 
Int Add latency & \cellcolor{cyan}0.4ns & 4ns & 10 X\tabularnewline
\hline 
FP Add Latency & \cellcolor{cyan}1.4ns & 4ns & 2.9 X\tabularnewline
\hline 
Approx DRAM Latency & \cellcolor{cyan}50ns & 270ns & 5.4 X\tabularnewline
\hline 
L2+L3 (GPU only L2) cache capacity & \cellcolor{cyan}8192KB & 128KB & 64 X\tabularnewline
\hline 
Approx Kernel Launch Latency & \cellcolor{cyan}25$\mbox{\ensuremath{\mbox{\ensuremath{\mu}s}}}$ & 50$\mbox{\ensuremath{\mu}s}$ & 2 X\tabularnewline
\hline 
\end{tabular*}

\protect\caption{CPU v.s GPU comparison (extracted from \cite{AMDSDKGuide})\label{tab:CPU-v.s-GPU}}
\end{table}

\begin{table}
\includegraphics[width=1\textwidth]{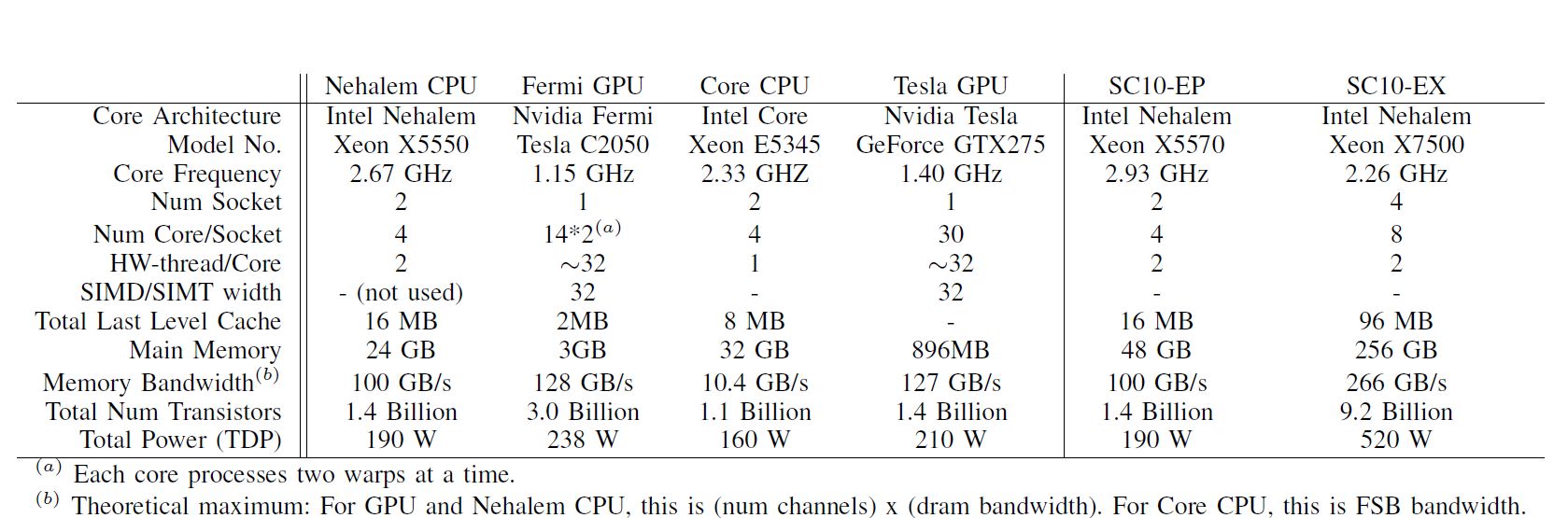}

\protect\caption{Intel CPU v.s NVIDIA GPU comparison (from \cite{conf/IEEEpact/HongOO11})\label{tab:Intel-CPU-v.s-NVIDIA-GPU}}

\end{table}

\begin{table}
\begin{centering}
\includegraphics[width=0.5\textwidth]{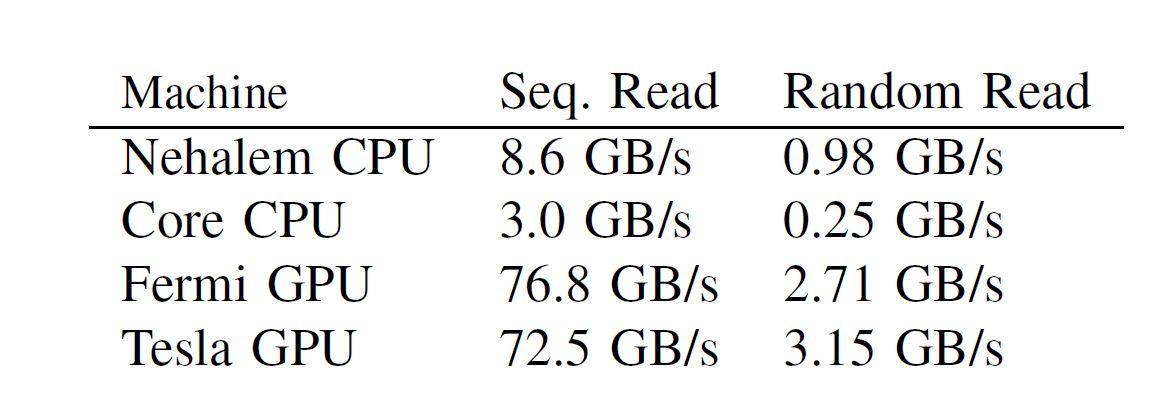}
\par\end{centering}

\protect\caption{Intel CPU v.s NVIDIA GPU memory bandwidth comparison (from \cite{conf/IEEEpact/HongOO11})
\label{tab:Intel-CPU-v.s-NVIDIA-GPU-mem}}

\end{table}

\paragraph*{\textmd{The tables show the key distinctions between the CPU and
the GPU are core frequency, instruction latency, and the capacity
of executing and scheduling large amount of hardware threads in parallel.
The GPU has an overwhelming advantage in the last measurement, which
compensentates for the minor deficiency in the first two measurements.
Overall, the GPU is a much superior computing device for massive parallel
tasks. }}

\paragraph*{\textmd{In my algorithm, the lowest parallel layer executes massive
amount of threads in parallel (corresponding to line 11-17 of Algorithm
\ref{alg:SPDP-Full-Gibbs}), thus taking advantage of massive parallel
execution and scheduling power of the GPU. The next parallel layer,
where words are being sampled in parallel in multiple of workgroups,
could take advantage of the large amount of computing units on the
GPU. As the size of count variables related to a single word is relevatively
small, the small but fast L2 cache of each computing unit perfectly
fits the need of carrying single or multiple workgroups. In the upper
most layer of the distributed parallel framework, where all documents
and words are randomly distributed to multiple GPU devices, the large
memory bandwidth of the GPU becomes useful. We can dispatch as many
words as possible to fill the global memory of each GPU, keeping the
GPU as busy as possible, ensure the GPU kernel execution time is far
greater than the time wasted on memory transfer. Multiple GPUs may
finish kernel execution at different times, but data can be asynchronously
read back to the CPU and the host memory for correction. The error
correction process has a high accuracy requirement. Each topic assignment
must be processed individually and many count variables must be incremented
atomically, making the CPU the most suitable device for this part
of the task.}}

\paragraph*{\textmd{To summarize, under the proposed three-level distributed
parallelism, the combined use of the GPU and the CPU surely improve
the efficiency of execution, given features of the GPU and the CPU
are used appropriately.}}

\FloatBarrier

\subsubsection{OpenCL v.s CUDA}

\paragraph*{\textmd{The two dominating programming frameworks for general purpose
GPU computing are CUDA, developed by NVIDIA, and OpenCL, developed
by Khronous Group. Khronous Group is led by Apple Inc, with members
including AMD, IBM, Intel, NVIDIA, and many other industry leaders.
While CUDA is optimized for NVIDIA GPUs, it is a closed ecosystem
controlled by NVIDIA which does not support any other device than
NVIDIA GPU. In contrast, OpenCL is designed to be an open and portable
framework which is supported by many types of devices including both
CPU and GPU. At present, OpenCL is supported on AMD GPU, AMD CPU,
Intel CPU, Intel GPU, NVIDIA GPU, Apple CPU, IBM PowerPC, and most
ARM CPU. An OpenCL program could run on all above types of devices
without any modification, as long as the program is not using any
device-type specific extensions provided by specific vendors. These
extensions are typically provided for optimization and debugging purposes.}}

\paragraph*{\textmd{Many performance analyses have been done for comparing these
two frameworks \cite{journals/corr/abs-1005-2581,conf/icpp/FangVS11}.
Throughout these analysis I did not find any convincing evidence which
shows one framework is better than another in terms of performance
and usability. Therefore, I decided to choose OpenCL as the programming
framework to implement my algorithm and conduct the experiments, because
of its openness and portability.}}

\begin{figure}
\includegraphics[width=1\textwidth]{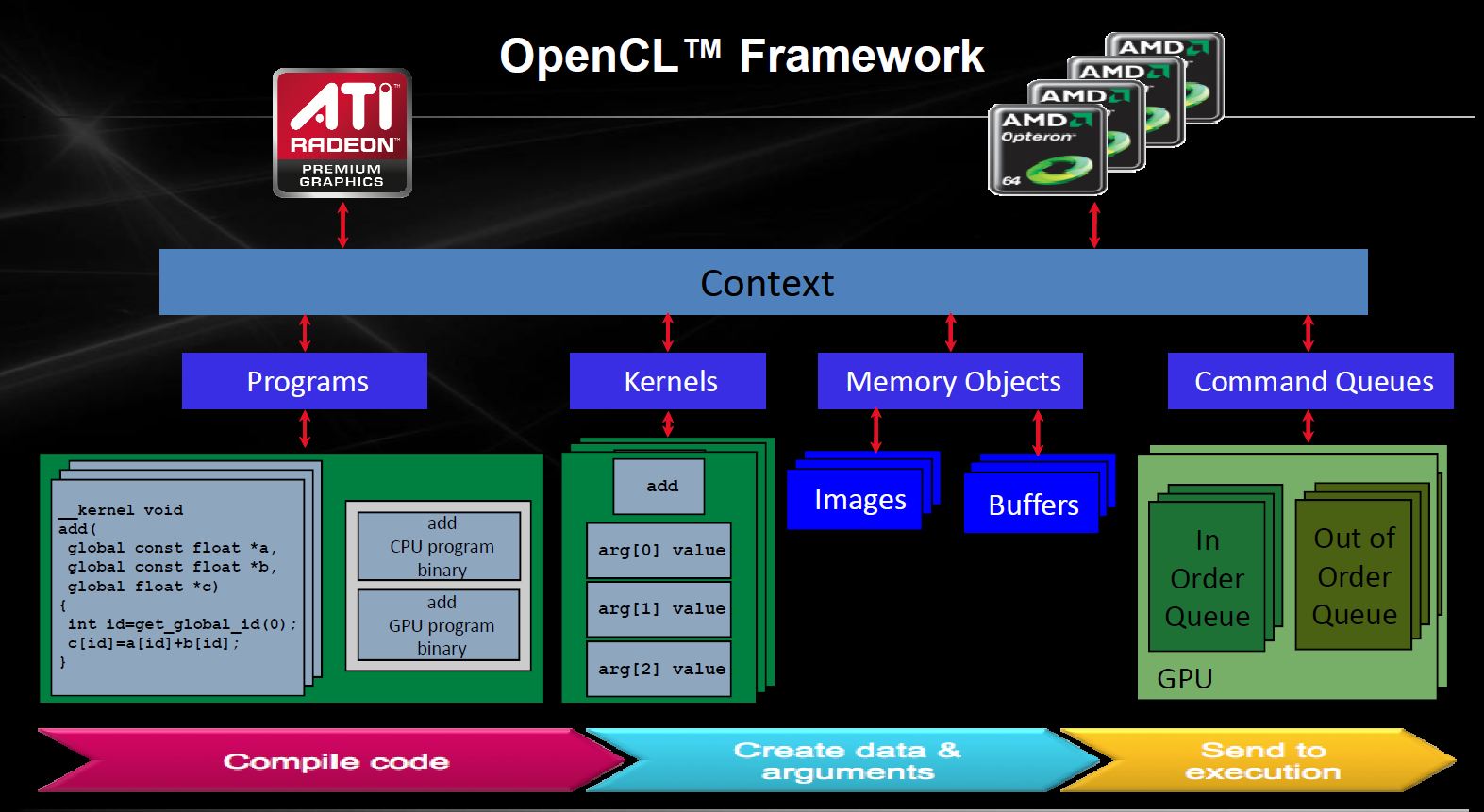}

\protect\caption{OpenCL Framework (from \cite{AMDOpenCLSlide})\label{fig:OpenCL-Framework}}

\end{figure}

\paragraph*{\textmd{Figure \ref{fig:OpenCL-Framework} shows an overview of the
OpenCL framework, taken from \cite{AMDOpenCLSlide}. }}

\paragraph*{\textmd{OpenCL defines three types of memory on GPU: global memory,
local memory, and private memory. Global memory is a shared memory
region on GPU, accessible to all computing units and all threads.
Local memory is located on each computing device, only accessible
to the computing device itself, but can be accessed much faster than
global memory. Private memory is small regions of memory only accessible
by each thread itself, often implemented as invidual registers on
hardware. The details of memory structure is looked at in section
\ref{sub:Hardware-and-Architectures}, where it is shown that my algorithm
fits well into this structure. }}

\paragraph*{\textmd{A context, roughly speaking, is a shared information structure
among devices in the same platform. Data belonging to the same context
can be communicated and exchanged among multiple devices, but only
devices in the same platform can share the same context. For example,
AMD CPUs and AMD GPUs belong to the same platform so they can share
the same context, while NVIDIA GPUs belong to another platform because
they use a different structure to store information, hence they cannot
share the same context with AMD devices. Intel CPUs, though implemented
OpenCL, have a totally different structure to any other type of devices,
thus cannot share the same context with either NVIDIA device or AMD
device.}}

\paragraph*{\textmd{A program is a piece of OpenCL code that can be compiled
at run-time. OpenCL uses a subset of C99 language as fundamental programming
framework, but it also allows programmers to use specific vendor extensions
for optimization and debugging. For example, at present double precision
floating point numbers are not officially supported by OpenCL specification.
Khronos group made an extension for it, and programmers can call ``\#pragma
OPENCL EXTENSION cl\_khr\_fp64 : enable'' to enable this feature.
AMD also have their own extension of double precision floating point
numbers, which can be enabled by calling ``\#pragma OPENCL EXTENSION
cl\_amd\_fp64 : enable''. More examples can be found in \cite{AMDSDKGuide,opencl08}.}}

\paragraph*{\textmd{A program can define multiple kernel functions and multiple
utility functions. Kernel functions have a specific keyword ``\_\_kernel''
as a signature, acting as entry points to the program to be executed
on OpenCL devices. When the program compiles, OpenCL automatically
optimize the program, at the same time changing all utility functions
into inline functions. }}

\paragraph*{\textmd{A program can define constant program-wide global variables,
but cannot have non-constant global variables. A program can allocate
fixed size arrays, but not dynamic arrays similar to what is created
by malloc() function in C. Varaibles in a program are allocated to
private memory by default, but with keywords ``\_\_global'' or ``\_\_local'',
a variable can be allocated to device global memory or local memory,
respectively. }}

\paragraph*{\textmd{A kernel is an entry point at execution of the program. Multiple
programs and multiple kernels are allowed for the same context, but
a kernel is not allowed to call another kernel in the same program,
as it must be scheduled on host. Each kernel may have a set of parameters
that can be allocated to private memory, local memory, or global memory.
Since kernel arguments are passed from the host, it is possible to
allocate a dynamic array as kernel arguments. Besides C99 primitive
types, a kernel argument can also be a struct, but the struct cannot
contain pointers, where the only exception when the pointers represent
fixed size arrays. The reason is it does not make sense pointing to
a region of memory other than fixed size arrays given OpenCL devices
have their own memory management. Similarly it doesn't make sense
to have pointers as kernel arguments except when these pointers represent
pre-allocated arrays. }}

\paragraph*{\textmd{A memory object represents a region of memory residing on
the device. From the perspective of the kernel, a memory object can
be read-only, write-only, or allowing both reading and writing, depending
on how the memory object is created on host. Instead of residing on
device, memory object can also be a pointer mapping itself to a region
in host memory, if the memory object is created with flag ``CL\_MEM\_USE\_HOST\_PTR''. }}

\paragraph*{\textmd{Each device has its own command queue to accept a sequence
of command instructions from the host. Typical command instructions
include launching kernel, reading memory object, writing memory object,
mapping memory object to a buffer, and many others. When the host
issue these commands, most of the time the programmer could choose
to either block the execution of host code for synchronization or
safety purposes, or not to block the execution of host code to allow
more flexible and efficient scheduling. An optional notification can
be sent back to the host when a command finishes execution on the
kernel.}}

\FloatBarrier

\subsubsection{Hardware and Architectures\label{sub:Hardware-and-Architectures}}

\paragraph*{\textmd{Many types of GPUs are available today, but the two major
ones, NVIDIA and AMD, who use slightly different architectures, have
almost all the market share in both the consumer and the professional
market. Programmers need to estimate the actual memory access pattern
and level of parallelism required for their applications, to find
the right type of GPU which suits their purposes. A siginificant amount
of analysis has been done in academia and industry to compare the
performance of these two types of GPUs, unanimously reaching the same
conclusion that each type of GPU has their own favourable applications.
For example, long time ago in the Bitcoin community \cite{BitcoinMiningHardwareComparison}
people have found AMD GPUs outperform NVIDIA GPUs for everything they
need, whereas in the scientific community people are often in favour
to NVIDIA GPUs because of better memory effiency, greater performance
in double precision computation, and better cluster energy efficiency.
A comperhensive academical study comparing a modern AMD GPU and NVIDIA
GPU can be found in \cite{NVIDIAvsAMDGPUArch}. }}

\begin{figure}
\begin{centering}
\includegraphics[width=0.5\textwidth]{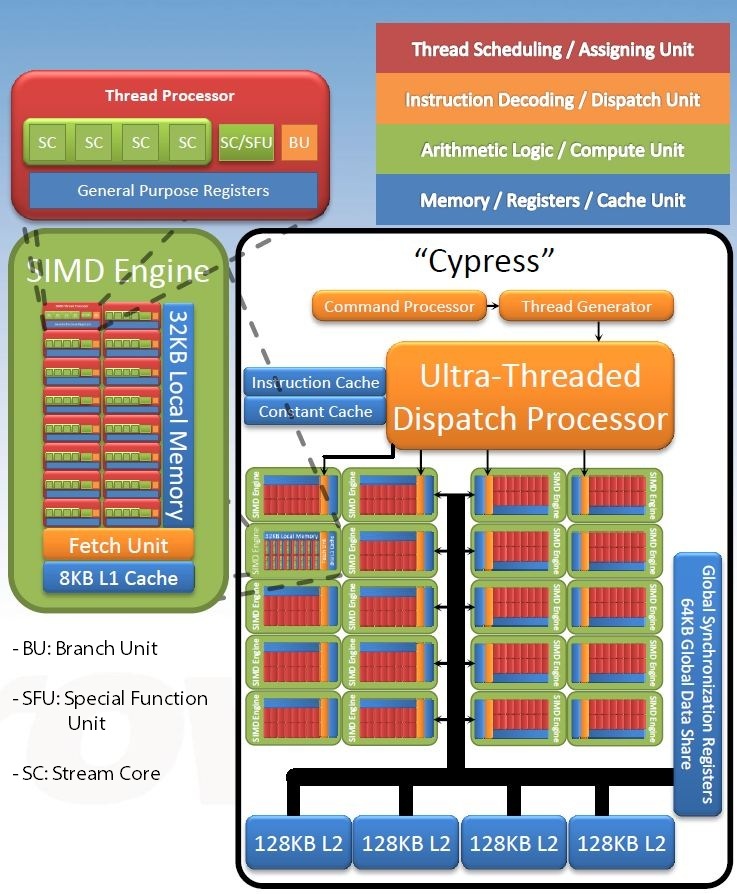}
\par\end{centering}

\protect\caption{AMD HD5870 ``Cypress'' Architecture (from \cite{MicrowayGPGPUcomparison})\label{fig:AMD-HD5870-Cypress}}

\end{figure}

\begin{figure}
\begin{centering}
\includegraphics[width=0.5\textwidth]{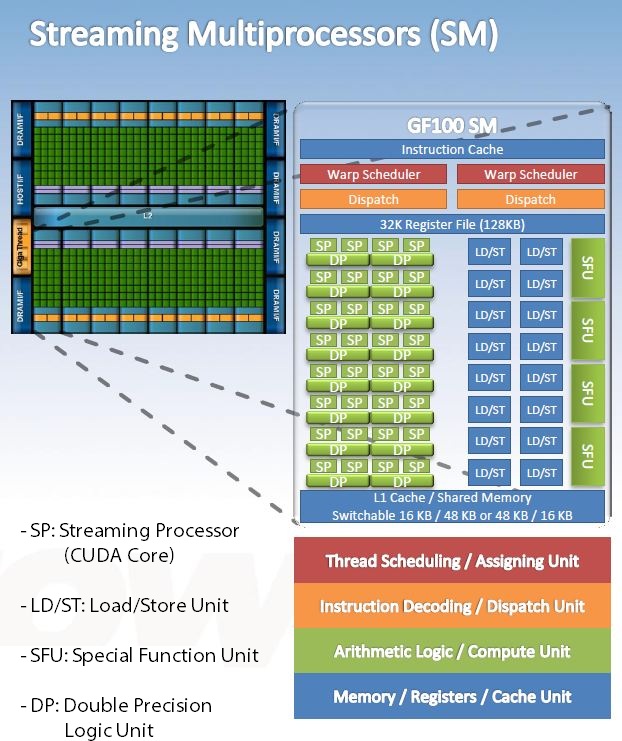}
\par\end{centering}

\protect\caption{NVIDIA GTX480 ``Fermi'' Architecture (from \cite{MicrowayGPGPUcomparison})
\label{fig:NVIDIA-GTX480-Fermi}}

\end{figure}

\paragraph*{\textmd{Figure \ref{fig:AMD-HD5870-Cypress} and \ref{fig:NVIDIA-GTX480-Fermi}
shows the architecture differences between AMD GPUs and NVIDIA GPUs.
This particular example compares AMD Radeon HD5870 ``Cypress'' GPU
and NVIDIA GeForce GTX480 ``Fermi'' GPU. These two GPUs are released
at around the same time in consumer market at approximately the same
price, and they have been chosen quite often to benchmark AMD GPUs
against NVIDIA GPUs for a period of time. In these figures, a warp
scheduler is a thread dispatching and scheduling unit. A warp is a
batch of simutaneous threads being dispatched and executed together
in a Single Instruction, Multiple Data (SIMD) style. Table \ref{tab:GTX480-v.s.-HD5870}
shows their differences in a few key parameters. All processors of
the Fermi GPU can operate at the shader frequency (see Table) except
the schedulers and texture units. }}

\paragraph*{\textmd{From the table and the figures it is not hard to see that
Fermi GPU has more advanced multi-thread dispatching architecture,
greater core frequency, larger memory, faster memory operations, and
faster memory exchange speed with host memory, benefited from its
multiple dedicated scheduler units, greater memory bandwidth, and
greater memory bus width. In contrast Cypress provide greater maximum
computing power measured in GFLOPs, higher number of processors (cores),
higher number of threads, higher number of computing devices, and
better energy efficiency. }}

\paragraph*{\textmd{Fermi's greater memory bandwidth is not particularly useful
feature to the task at hand as long as we can keep the device busy
by pushing maximum amount of data per iteration. However, in my proposed
algorithm, the two-level parallelism involves considerable amount
of global memory operations. Some features provided by Fermi, namely
higher core frequency, higher memory frequency, and greater cache
size, are greatly beneficial in this perspective. On the other hand
if we could manage to reduce the amount of global memory access operations,
and leverage the high throughput provided by AMD GPU to compensate
its mild disadvantage in core frequency, AMD GPUs may perform much
better than NVIDIA GPUs. However, achieving this may require complicated
optimization at the code level for specific hardware. }}

\begin{table}
\begin{centering}
\begin{tabular}{|c|c|c|}
\hline 
GPU Name & GTX480 ``Fermi'' & HD5870 ``Cypress''\tabularnewline
\hline 
\hline 
Core Frequency & 700MHz(Core) /1400MHz(Shader) & 850MHz\tabularnewline
\hline 
Memory Frequency & 3696MHz & 1200MHz\tabularnewline
\hline 
Memory Bandwidth & 177.4GB/s & 153.6GB/s\tabularnewline
\hline 
Number of Cores & 480 & 1600\tabularnewline
\hline 
Number of Computing Devices & 15 Streaming Multiprocessors & 20 SIMD Engines\tabularnewline
\hline 
Memory Bus width & 384-bit & 256-bit\tabularnewline
\hline 
GFLOPs & 1345 & 2720\tabularnewline
\hline 
Max. Power Consumption & 250W & 188W\tabularnewline
\hline 
\end{tabular}
\par\end{centering}

\protect\caption{GTX480 v.s. HD5870\label{tab:GTX480-v.s.-HD5870}}

\end{table}

\FloatBarrier

\subsection{Implementation}

\paragraph*{\textmd{Although an OpenCL kernel program must be written in C, the
host program can be written in other languages. The official OpenCL
specification use C++ for native host code language but wrappers have
been already created for multiple languages including Java, Objective-C,
Python, and many others. To achieve maximum portability without sacrifice
in efficiency we have chosen Java for the implementation and the experiments.
When properly implemented, Java has comparable efficiency to C and
C++. }}

\paragraph*{\textmd{Algorithm \ref{alg:Multi-GPU-SPDP-Host} and \ref{alg:Multi-GPU-SPDP-Kernel}
outlines the implementation pseudo code for my algorithm MGPU-DP-SPDP
(Multi-GPU Distributed Parallel SPDP). As mentioned as the beginning
of this section, we have transformation matrices $P^{i}$ set to identity,
which limits the number of word-associations to 1, and maximum number
of threads needed at topic sampling level to number of topics times
2,. Typically, two times number of topics is far less than maximum
numbers of threads in a local workgroup supported by a GPU device,
hence for simplicity we assume this is the case in algorithm \ref{alg:Multi-GPU-SPDP-Kernel},
and let each thread handle exactly one probability calculation. If
number of topics times 2 is more than maximum numbers of threads in
a local workgroup supported by a GPU device, the only thing needs
to be done is to adjust the number of probability calculations one
thread is required to handle.}}

\paragraph*{\textmd{Readers who intend to implement the algorithm by themselves
to reproduce the result should be aware that Line 22 and 24 of the
pseudo code in Algorithm \ref{alg:Multi-GPU-SPDP-Kernel} should be
computed in logarithm space to avoid overflow, especially when the
required Stirling numbers can be an extremely large number at runtime.
In my experiments it is observed in the original SPDP algorithm, a
medium size document collection could require Stirling numbers as
large as to the magnitude of $e$ to the power of hundreds or thousands.
The probability formulas provided in Equation \ref{eq:SPDP-sampling-w-z-r0},
Equation \ref{eq:SPDP-sampling-w-z-r1}, and Algorithm \ref{alg:SPDP-Full-Gibbs}
are proportionalities. The result must be normalized before being
used to generate a sample, and appropriate technique must be used
to avoid overflow. One common way is to use the identity below, where
$p_{n}$ denotes the probability and $q_{n}$ denotes corresponding
proportionalities: 
\[
p_{n}=exp(\frac{\log q_{n}-\log\max\{q_{0},q_{1},q_{2},...q_{2K}\}}{\sum_{n}(\log q_{n}-\log\max\{q_{0},q_{1},q_{2},...q_{2K}\})})
\]
}}

\paragraph*{\textmd{This implementation is not optimized. The number of threads
in each workgroup is not fine-tuned to suit GPU's preference, and
global memory operations are not reduced to increase its efficiency
on AMD GPUs (see Section \ref{sub:Hardware-and-Architectures}). An
optimized implementation may further improve the running speed of
my algorithm by many times, but this is not in the scope of this thesis.
A basic version (which is used in the experiments) is given here to
allow readers with any generic equipment able to reproduce the result.
\label{par:implementation-why-we-did-not-optimize}}}

\begin{algorithm}
\begin{algorithmic}[1]
\State Initialize OpenCL. Read data from file.
\State Randomly initialize topic assignments $z$
\State Initialize counting variables $m,n,t,q,v$ from $z$.
\State Run CPU SPDP algorithm for 1 iteration
\State Create shared device memory for common parametres
\While{Gibbs not converged}
	\State Create shared device memory $n,m,q,v$
	\ForAll{GPU $g$}
		\State Create device specific memory for topic assignments $z_g$ from $z$, counts $t_g$ from $t$
		\State Create kernel with shared parameters, including device specific memory, shared device memory, and common parametres
		\State Enque execution of kernel
	\EndFor
	\State Wait until all kernels have finished execution
	\ForAll{GPU $g$}
		\State Read back $t_g$ from $g$
		\State Read back topic assignments $z_g$ from $g$
	\EndFor
	\State Synchronize $t_g$ and merge into $t$
	\State Re-generate counts from $z,t$ 
\EndWhile
\end{algorithmic}

\protect\caption{\label{alg:GPU-SPDP-host}MGPU-DP-SPDP Host Code\label{alg:Multi-GPU-SPDP-Host}}
\end{algorithm}

\begin{algorithm}
\begin{algorithmic}[1]
\Function{ParallelSampleWord}{Global Counts, Shared SPDP Parametres, Local Memory Variables ...}
	\State Load shared parametres into private memory
	\State $K \gets $ number of topics
	\State $gid \gets$ Workgroup Group ID, $lid \gets$ Workgroup Local ID
	\State $d \gets$ document ID, $i \gets$ group ID, $w \gets$ word ID of workgroup $gid$, $k \gets$ topic assigned to word
	\If{$lid \le 2K$ and $lid \equiv 0 mod 2$}
		\State $kid \gets lid/2$ (topic represented by $lid$)
		\State Copy global counts of topic $kid$ word $w$ group $i$ into local memory
		\State Copy global counts of topic $kid$ document $d$ group $i$ into local memory
		\State Copy auxillary global counts, e.g sums, related to topic $kid$ into local memory
		\State Correct local counts copied by $kid$ to ensure they are in valid range
		\State Correct global counts copied by $kid$ to ensure they are in valid range.
	\EndIf
	\State Synchronize all threads in local workgroup
	\If{$lid \equiv 0$}
		\State Decrement word counts in local memory in the same way as in non-parallel algorithm
		\State Decrement word counts in global memory atomically
	\EndIf
	\If{$lid \le 2K$}
		\State $kid \gets lid/2$ (topic represented by $lid$)
		\If{$lid \equiv 0 mod 2$}
			\State $p[lid] \gets$ probability for $z_{gid}=kid$, $r_{gid}=1$, $v_{i,k,w,t}=w$
		\Else
			\State $p[lid] \gets$ probability for $z_{gid}=kid$, $r_{gid}=0$:
		\EndIf
	\Else
		\State $p[lid] \gets 0$ 
	\EndIf
	\State Synchronize all threads in local workgroup
	\If{$lid \equiv 0$}
		\State $z_{gid} \gets$ new topic assignment sampled from $p$
		\State Increment counts in global memory atomically
	\EndIf
\EndFunction
\end{algorithmic}

\protect\caption{\label{alg:GPU-SPDP-kernel}MGPU-DP-SPDP Kernel\label{alg:Multi-GPU-SPDP-Kernel}}
\end{algorithm}

\FloatBarrier

\chapter{Experiments}

\section{Experimental Setup}

\paragraph*{\textmd{Three datasets \footnote{The Reuters RCV1 corpus, Volume 1: English Language, 1996-08-20 to 1997-08-19, \cite{Lewis04}.}
are used for the experiments to show the effectiveness of the three-level
distributed-parallel framework introduced in Chapter 2. For all documents,
words that do not carry any semantic meaning are discarded, such as
``I'' ``you'', ``she'', ``he'', ``are'', ``is'', ``it'',
``am'', the so called stop words. Apostrophes, comma, and other
symbols which do not carry semantic meanings are also removed, and
words are tokenized into integers. We consider the plural form of
a word is distinct from its singular form, as they may carry different
meanings in some situations:}}
\begin{enumerate}
\item Dataset ``RedState v.s DailyKos'': A blog dataset that includes
2276 blogs from American political blogs RedState (\href{http://www.redstate.com}{www.redstate.com})
and 2386 blogs from DailyKos (\href{http://www.dailykos.com}{www.dailykos.com})
to the end of 2011. Both blogs are focused on current political issues
and government policies in the United States, and related international
issues, but their political positions are opposing each other. It
is well known that RedState is in association with Republican politicians,
and DailyKos is an active supporter of Democratic Party in the United
States. I train the algorithm with 2045 and 2146 blogs from RedState
and DailyKos respectively, and evaluate the perplexity on the rest
of the data. After words and symbols are processed and discarded,
this yields a collection of documents that has vocabulary size 14724,
average effective document length of 157 words in the RedState group,
and average document length of 103 words in the DailyKos group.
\item Dataset ``Reuters Disaster'': A news article dataset collected from
worldwide disasters news articles between 1996 to 1997, written by
journalists from different regions of the news agency ``Reuters''.
Articles are divided into four regions: (1) Middle East and South
East Asia (1508 articles), (2) European (1834 articles), (3) Asia
Pacific (1580 articles), (4) America (mainly the United States, 2418
articles). For each group, 10\% of the data is held out from training
for computing the perplexity. After words and symbols are processed
and discarded, this yields a collection of documents that has vocabulary
size 51377, average effective document length of 128 words, 134 words,
132 words, and 151 words in group 1, 2, 3, 4 respectively.
\item Dataset ``International Political News'': A news article dataset
collected from worldwide political news sources between 1996 to 1997,
containing both government releases and private news agencies. Data
is divided into three groups according to location where the news
article is written: (1) European and Africa (4034 articles) (2) North
America and South America (5042 articles) (3) Asia Pacific (6897 articles).
For each group, 10\% of the data is held out from training for computing
perplexity. After words and symbols are processed and discarded, this
yields a collection of documents that has vocabulary size 156080,
average effective document length of 184 words, 177 words, and 151
words in group 1, 2, 3 respectively.
\end{enumerate}

\paragraph*{\textmd{The three datasets differ significantly in size. Dataset
1,2,3 contain about 500000, 900000, and 3000000 words respectively.
In addition, the vocabulary size of dataset 2 is about 3.5 times as
large as dataset 1, and the vocabulary size of dataset 3 is about
10 times as large as dataset 1.}}

\paragraph*{\textmd{I use the following parametres for SPDP: $\alpha=0.1$ (Dirichlet
hyper-parameter for words), $\gamma=0.1$ (Dirichetlet hyper-parameter
for topics), $a=0.7$ (discount parameter for SPDP), $b=100$ (concentration
parameter for SPDP). These parameters are suggested as optimal parameters
for blog data set by Chen \cite{chen2012spdp}, obtained by trial
and error. It is also possible to learn the concentration parameters
in the algorithm to improve the result, but these parameters are learned
outside the regular collapsed Gibbs sampling cycle. Since learning
these parameters is not related to the work in GPU speed improvement,
these techniques are not discussed in this thesis. }}

\paragraph*{\textmd{Perplexity will be evaluated based on the counts after 2000
Gibbs iterations. The common practice is to run less Gibbs iterations,
and to take the average output every few iterations after a burn-in
period to allow the counts to enter a relatively stable region. In
my experiments, this method is not used. Instead, the program runs
more iterations and reads out the output only once, because longer
iterations leads to better convergence, eliminates the error accumulated
by taking the average output from some states that may be far away
from the converging state. I observed that my improved algorithm,
even when it is only running on a single GPU in my test machine, could
be 50 times faster than the original SPDP implementation. When running
time is no longer a concern of SPDP algorithm, there is no reason
for not choosing to spend a bit more time on training in exchange
for a more accurate result. }}

\paragraph*{\textmd{I have two machines: the first machine is a homemade GPU
cluster configured with seven AMD HD5850 ``Evergreen'' GPUs, one
AMD FX8150 ``Bulldozer'' CPU, and 16GB Corsair DDR III 1667MHz RAM;
the second machine is a high-end laptop workstation configured with
one NVIDIA GTX460M ``Fermi'' GPU, one Intel Core i7 740QM CPU, and
8GB Kingston 1333MHz RAM. The difference is in RAM frequency will
not be an issue in benchmarking, since little computation is done
on CPU, and the transmission delay between CPU and main memory is
negligable. The AMD HD5850 is a reduced version of the HD5870 introduced
in the last chapter, and NVIDIA GTX460M is a reduced version of the
GTX480. They share the same architecture with their corresponding
full versions, the differences are the number of cores, core clocks,
memory rate, and memory frequency. Table \ref{tab:Experiment-Hardware-Platform}
shows some key parameters of the four computing devices mentioned
above, extracted from the vendor's official website.}}

\begin{table}
\begin{centering}
\begin{tabular}{|>{\centering}p{0.18\textwidth}|>{\centering}p{0.18\textwidth}|>{\centering}p{0.18\textwidth}|>{\centering}p{0.18\textwidth}|>{\centering}p{0.18\textwidth}|}
\hline 
Platform & GTX460M ``Fermi'' & HD5850 ``Cypress'' & AMD FX8150 ``Bulldozer'' & Intel Core i7 740QM\tabularnewline
\hline 
Core Frequency & 675MHz(Core) /1350MHz(Shader) & 725MHz & 3.6GHz\textasciitilde{}4.2GHz & 1.73GHz\textasciitilde{}2.93GHz\tabularnewline
\hline 
Memory Frequency & 2500MHz & 1000MHz & \textgreater{}1866MHz & 1066/1333MHz\tabularnewline
\hline 
Memory Bandwidth & 60GB/s & 128GB/s & N/A & N/A\tabularnewline
\hline 
Number of Cores & 192 & 1440 & 8 & 4 (8 Threads)\tabularnewline
\hline 
GFLOPs & 518.4 & 2088 & N/A & N/A\tabularnewline
\hline 
Max. Power Consumption & 50W & 151W & 125W & 45W\tabularnewline
\hline 
Local Memory Size & 48KB & 32KB & N/A & N/A\tabularnewline
\hline 
\end{tabular}
\par\end{centering}

\protect\caption{Experiment Hardware Platform Comparison\label{tab:Experiment-Hardware-Platform}}
\end{table}

\paragraph*{\textmd{Throughout the experiments, I will use the amount of time
elapsed for each Gibbs iteration (minus the amount of time spent on
the debugging code, such as unnecessary memory transfer between the
host and the GPUs) as the measure to compare the running time of different
experiments. The reason that a profiler is not used to measure the
amount of time allocated to processes and kernels is that it is also
desirable to to take the amount of time spent on memory operations
into account. I want to measure overall how fast my algorithm is compared
to the original SPDP algorithm. A fine-tuned profiler is not necessary
for the task because it is expected that the running time of MGPU-SPDP
is only a small fraction of the running time of the original SPDP
algorithm, and we only want to get some rough ratios. }}

\section{Experiments}

\FloatBarrier

\paragraph*{\textmd{The first experiment is performed on the original SPDP Java
implementation. The result is used as a benchmark to compare with
the results from the other experiments which are performed with different
versions of MGPU-DP-SPDP that enable part or all features in the proposed
three-level distributed parallelism. }}

\begin{figure}
\begin{centering}
\includegraphics[width=1\textwidth]{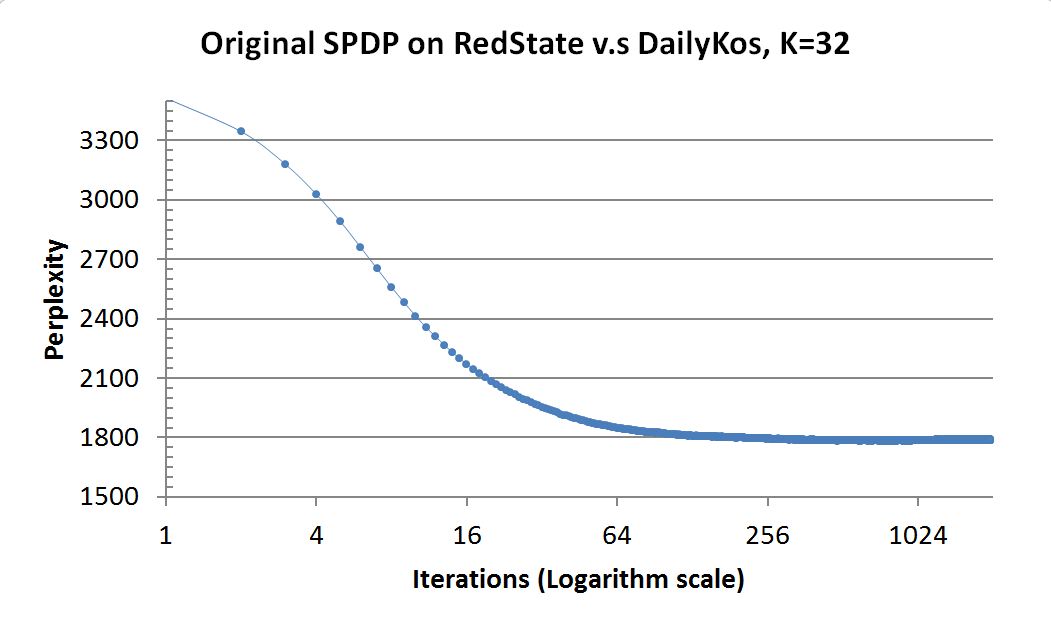}
\par\end{centering}

\protect\caption{SPDP perplexity through Gibbs iterations (logarithm scaled) with the
dataset ``RedState v.s DailyKos''\label{fig:SPDP-perplexity-through}}

\end{figure}

\paragraph*{\textmd{Table \ref{tab:SPDP-running-time} and Figure \ref{fig:SPDP-perplexity-through}
shows the perplexity of the original SPDP algorithm with dataset ``RedState
v.s DailyKos'' on both machines where number of topics is set to
be $K=32$. The algorithm converges to approximately the same perplexity
on both platforms at the same iterations, showing that although the
algorithm is stochastic, the convergence speed is stable and platform
independent, hence it is not necessary to test my algorithm on different
devices which belong to the same class. Although the perplexity changes
little after around 25 iterations, topic quality keeps getting better
as the algorithm progresses, as shown in table \ref{tab:SPDP-topic-quality-24}.
For each row in ``Iterations'' column, two rows in ``Top Words''
column corresponds to different group of documents, the first row
represents RedState group, second row represents DailyKos groups.
Topics are shared across groups but each group has its own ranking
of a particular topic.}}

\begin{table}
\begin{centering}
\begin{tabular}{|c|c|c|}
\hline 
Platform & AMD FX8150 ``Bulldozer'' & Intel Core i7 740QM\tabularnewline
\hline 
\hline 
Perplexity after 1 iteration & 3514 & 3514\tabularnewline
\hline 
Perplexity after 5 iterations & 2910 & 3028\tabularnewline
\hline 
Perplexity after 10 iterations & 2449 & 2412\tabularnewline
\hline 
Perplexity after 25 iterations & 2042 & 2015\tabularnewline
\hline 
Perplexity after 100 iterations & 1817 & 1819\tabularnewline
\hline 
Perplexity after 2000 iterations & 1790 & 1788\tabularnewline
\hline 
\end{tabular}
\par\end{centering}

\protect\caption{SPDP running time and perplexity with the dataset ``RedState v.s
DailyKos''\label{tab:SPDP-running-time}}

\end{table}

\begin{table}
\begin{centering}
\begin{tabular}{|c|c|>{\centering}p{0.5\textwidth}|}
\hline 
Iterations & Rank & Top Words (out of top 50) in Topic 24\tabularnewline
\hline 
\multirow{2}{*}{25} & 23 & state, money, democrats, way, right, thing, just, california, spending,
election, better, republicans, party, best, city, need, vote, really,
general, left, day, second, primary, governor, republican,\tabularnewline
\cline{2-3} 
 & 13 & republicans, democrats, republican, money, election, democratic, better,
need, just, thing, congress, primary, blue, majority, time, public,
looking, senate, right, brown, ballot, really, state, elections\tabularnewline
\hline 
\multirow{2}{*}{100} & 16 & money, spending, public, democrats, want, way, good, just, big, right,
problem, need, bad, important, better, budget, spend, congress, hard,
time, needs, republicans, best, able, effort, trying, fund, past\tabularnewline
\cline{2-3} 
 & 11 & republicans, democrats, money, know, congress, just, democratic, republican,
nation, big, better, left, got, way, time, right, trying, spending,
want, year, spend, hold, need, needs, public, finally, away,\tabularnewline
\hline 
\multirow{2}{*}{500} & 12 & tax, taxes, spending, money, government, economic, federal, economy,
business, american, year, cuts, democrats, income, congress, need,
raise, spend, increase, people, jobs, policy, dollars, cut, growth\tabularnewline
\cline{2-3} 
 & 19 & money, tax, bush, economic, dollars, government, republicans, pay,
spending, democrats, economy, people, iraq, run, taxes, year, corporate,
time, raise, fund, federal, cut, big, recession, american,\tabularnewline
\hline 
\multirow{2}{*}{2000} & 13 & tax, government, taxes, economy, money, spending, economic, federal,
jobs, business, increase, income, cuts, policy, year, american, growth,
need, people, pay, state, cut, dollars, americans, good\tabularnewline
\cline{2-3} 
 & 20 & money, economy, tax, economic, government, dollars, bush, companies,
taxes, crisis, need, spending, year, pay, american, credit, federal,
america, recession, wall, increase, cut, business, social, company\tabularnewline
\hline 
\end{tabular}
\par\end{centering}

\protect\caption{SPDP topic quality progression on topic 11 with the dataset ``RedState
v.s DailyKos'' \label{tab:SPDP-topic-quality-24}}

\end{table}

\paragraph*{\textmd{Table \ref{tab:SPDP-running-time-diff-topics} shows the
running time per iteration of the original SPDP algorithm with dataset
``RedState v.s DailyKos'' on both machines where the number of topics
varies. The experiment performed on the Intel Core i7 740QM has slightly
better running time due to the architectural advantage. The performance
sharply dropped when the number of topics is increased to 256, due
to fast expanding memory consumption (3GB) hence decreased CPU on-chip
cache performance. The AMD FX8150 outperformed the Intel Core i7 740QM
due to significantly larger level 2 cache (2$\times$4MB v.s. $4\times$256KB),
and larger level 3 cache (8MB v.s 6MB).}}

\begin{table}
\begin{centering}
\begin{tabular}{|c||c|c|}
\hline 
Number of Topics & AMD FX8150 ``Bulldozer'' & Intel Core i7 740QM\tabularnewline
\hline 
32 & 17.700 seconds & 16.510 seconds\tabularnewline
\hline 
64 & 35.024 seconds & 32.781 seconds\tabularnewline
\hline 
128 & 69.746 seconds & 66.230 seconds\tabularnewline
\hline 
\end{tabular}
\par\end{centering}

\protect\caption{SPDP running time for different number of topics on ``RedState v.s
DailyKos'' \label{tab:SPDP-running-time-diff-topics}}
\end{table}

\begin{figure}
\begin{centering}
\includegraphics[width=1\textwidth]{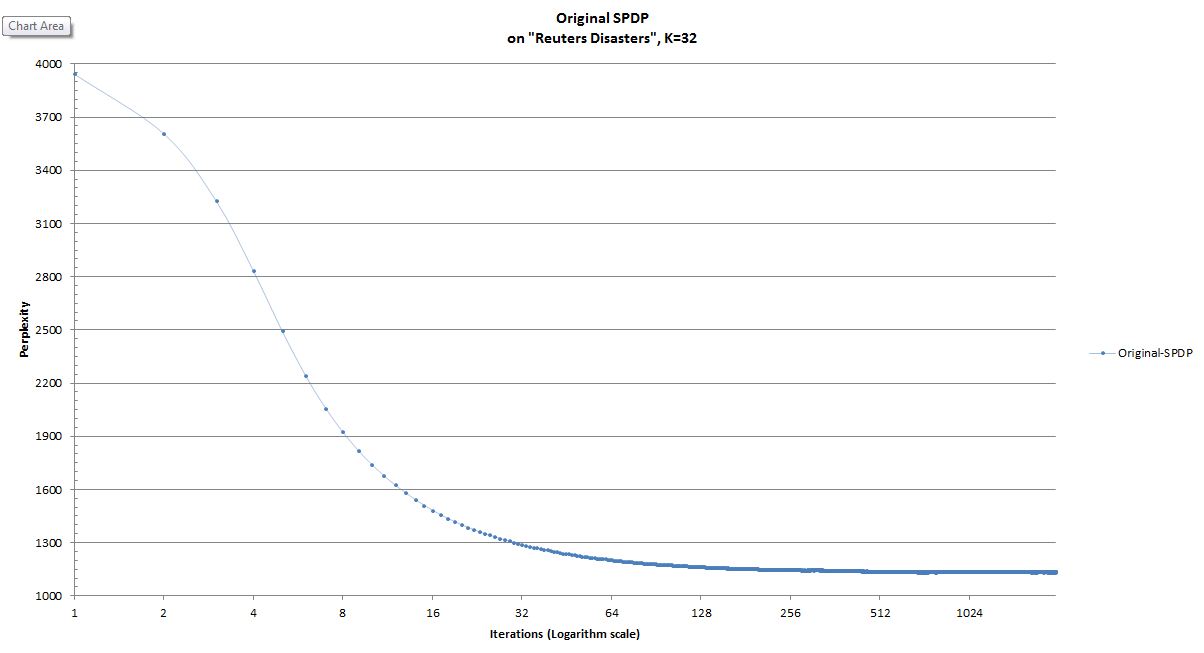}
\par\end{centering}

\protect\caption{SPDP perplexity through Gibbs iterations (logarithm scaled) with the
dataset ``Reuters Disasters''\label{fig:SPDP-perplexity-RD}}
\end{figure}

\begin{figure}
\begin{centering}
\includegraphics[width=1\textwidth]{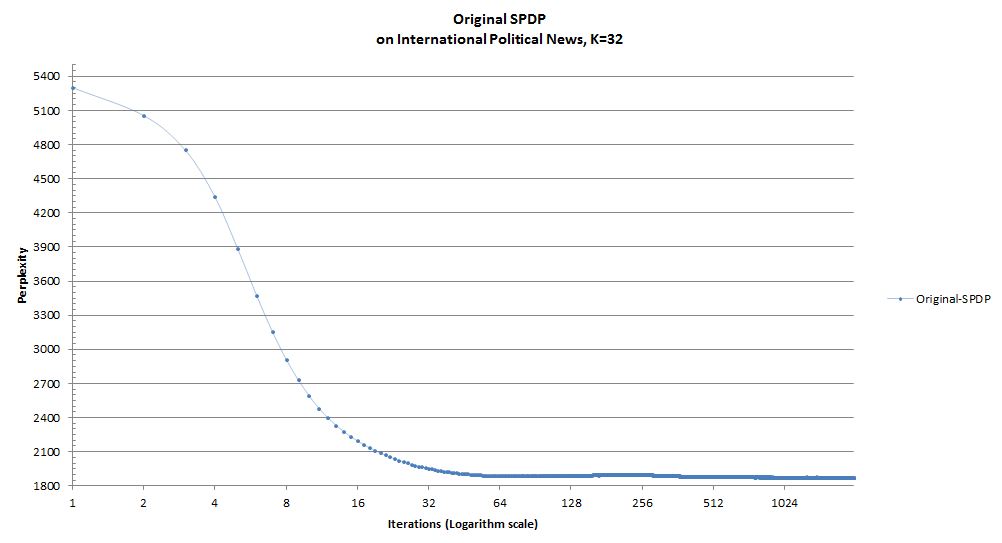}
\par\end{centering}

\protect\caption{SPDP perplexity through Gibbs iterations (logarithm scaled) with the
dataset ``Reuters Disasters''\label{fig:SPDP-perplexity-POL}}
\end{figure}

\paragraph*{\textmd{Table \ref{tab:SPDP-running-time-other-datasets} shows the
running time per iteration of original SPDP algorithm with the datasets
``Reuters Disaster'' and ``International Political News''. Figure
\ref{fig:SPDP-perplexity-RD} and \ref{fig:SPDP-perplexity-POL} show
the perplexity convergence of the original SPDP algorithm with these
two datasets. Table \ref{tab:SPDP-topic-quality-RvsD}, \ref{tab:SPDP-topic-quality-RD},
and \ref{tab:SPDP-topic-quality-POL} show a few converged topics
of original SPDP algorithm, on the dataset ``RedState v.s DailyKos'',
``Reuters Disaster'', and ``International Political News'' respectively.
The number of topics is set to 32 in these experiments.}}

\begin{table}
\begin{centering}
\begin{tabular}{|>{\centering}p{0.3\textwidth}||>{\centering}p{0.3\textwidth}|>{\centering}p{0.3\textwidth}|}
\hline 
Dataset & AMD FX8150 ``Bulldozer'' & Intel Core i7 740QM\tabularnewline
\hline 
``Reuters Disaster'' & 29.762 seconds & 26.621 seconds\tabularnewline
\hline 
``International Political News'' & 86.355 seconds & \textasciitilde{}192 seconds (memory issue)\tabularnewline
\hline 
\end{tabular}
\par\end{centering}

\protect\caption{SPDP running time with the datasets ``Reuters Disaster'' and ``International
Political News'' ($K=32$) \label{tab:SPDP-running-time-other-datasets}}
\end{table}

\FloatBarrier

\begin{table}
\begin{centering}
\begin{tabular}{|c|c|c|>{\centering}p{0.5\textwidth}|}
\hline 
Topic ID & Probability & Rank & Top Words (out of top 50)\tabularnewline
\hline 
\multirow{2}{*}{24} & 0.026321 & 13 & tax, government, taxes, economy, money, spending, economic, federal,
jobs, business, increase, income, cuts, policy, year, american, growth,
need, people, pay, state, cut, dollars, americans, good\tabularnewline
\cline{2-4} 
 & 0.017637 & 20 & money, economy, tax, economic, government, dollars, bush, companies,
taxes, crisis, need, spending, year, pay, american, credit, federal,
america, recession, wall, increase, cut, business, social, company\tabularnewline
\hline 
\multirow{2}{*}{30} & 0.023686 & 18 & iraq, war, troops, surge, qaeda, iraqi, afghanistan, security, american,
petraeus, military, strategy, forces, government, year, progress,
general, country, withdrawal, soldiers, success, sadr, army, \tabularnewline
\cline{2-4} 
 & 0.030925 & 12 & iraq, war, troops, military, bush, iraqi, american, surge, years,
government, americans, violence, forces, year, soldiers, iraqis, president,
cheney, invasion, months, progress, plan, security, success,\tabularnewline
\hline 
\multirow{2}{*}{31} & 0.042687 & 5 & romney, huckabee, mccain, thompson, rudy, mitt, fred, campaign, iowa,
candidate, republican, race, giuliani, think, john, hampshire, mike,
going, candidates, debate, immigration, south, reagan, win, huck,\tabularnewline
\cline{2-4} 
 & 0.027363 & 13 & romney, huckabee, mccain, iowa, republican, mitt, paul, gop, poll,
race, thompson, rudy, candidates, hampshire, win, mike, giuliani,
republicans, fred, john, ron, candidate, campaign, voters, field,
polls\tabularnewline
\hline 
\multirow{2}{*}{16} & 0.013617 & 23 & health, care, government, state, insurance, program, market, private,
coverage, plan, healthcare, san, reform, city, based, mandate, medical,
law, legislation, choice, pay, francisco, universal, benefits,\tabularnewline
\cline{2-4} 
 & 0.012852 & 25 & health, care, public, insurance, government, issues, schip, reform,
flu, pandemic, times, science, children, important, coverage, ny,
americans, emergency, disaster, sick, state, program, universal, safety\tabularnewline
\hline 
\end{tabular}
\par\end{centering}

\protect\caption{SPDP topic quality with dataset ``RedState v.s DailyKos''\label{tab:SPDP-topic-quality-RvsD}}
\end{table}

\begin{table}
\begin{centering}
\begin{tabular}{|c|c|c|>{\centering}p{0.5\textwidth}|}
\hline 
Topic ID & Probability & Rank & Top Words (out of top 50)\tabularnewline
\hline 
\multirow{4}{*}{31} & 0.081832 & 2 & iran, quake, earthquake, richter, scale, killed, iranian, irna, measuring,
hit, damage, area, relief, villages, injured, official, IRAN, region,
tehran, agency, province, earthquakes, news, ardabil, radio,\tabularnewline
\cline{2-4} 
 & 0.031095 & 8 & earthquake, scale, richter, quake, damage, tremor, iran, reported,
measuring, killed, felt, town, struck, area, epicentre, eastern, institute,
region, shook, gmt, injured, minor, hit, southern, miles\tabularnewline
\cline{2-4} 
 & 0.085058 & 1 & earthquake, scale, richter, damage, quake, measuring, tremor, region,
official, reports, gmt, km, area, casualties, jiashi, miles, earthquakes,
officials, china, epicentre, felt, shook, struck, seismological\tabularnewline
\cline{2-4} 
 & 0.018930 & 20 & earthquake, los, angeles, quake, damage, area, richter, USA, california,
miles, iran, scale, reports, san, earthquakes, km, volcano, ash, struck,
reported, alaska, francisco, survey, injuries, local, felt\tabularnewline
\hline 
\multirow{4}{*}{1} & 0.026527 & 15 & cyclone, state, andhra, pradesh, storm, coast, india, godavari, east,
hit, coastal, officials, winds, naidu, INDIA, miles, hyderabad, rice,
km, kakinada, indian, hectares, relief, port, rupees, fishermen,\tabularnewline
\cline{2-4} 
 & 0.020554 & 17 & poland, flood, wroclaw, floods, odra, polish, POL, water, warsaw,
crisis, city, committee, opole, villages, wave, pomes, towns, sq,
kwasniewski, communities, waters, czech, cimoszewicz, flooded, disaster,\tabularnewline
\cline{2-4} 
 & 0.029046 & 11 & mph, winds, typhoon, japan, storm, hours, shipping, expected, weather,
top, southern, flooding, east, USA, north, okinawa, wsc, moving, heavy,
tropical, threat, islands, sea, system, time, JAP, rains, \tabularnewline
\cline{2-4} 
 & 0.024342 & 11 & fire, firefighters, blaze, fires, acres, california, hectares, forest,
homes, contained, san, los, burning, national, wildfires, angeles,
officials, battling, burned, percent, area, miles, wildfire, malibu\tabularnewline
\hline 
\end{tabular}
\par\end{centering}

\protect\caption{SPDP topic quality with the dataset ``Reuters Disasters''\label{tab:SPDP-topic-quality-RD}}
\end{table}

\begin{table}
\begin{centering}
\begin{tabular}{|c|c|c|>{\centering}p{0.5\textwidth}|}
\hline 
Topic ID & Probability & Rank & Top Words (out of top 50)\tabularnewline
\hline 
\multirow{3}{*}{23} & 0.026381 & 10 & rights, human, report, u.n, group, international, amnesty, commission,
ogoni, groups, arrested, journalists, prison, visit, abuses, statement,
mission, detention, detained, nations, trial, alleged, \tabularnewline
\cline{2-4} 
 & 0.033964 & 6 & rights, human, china, u.n, resolution, commission, united, east, report,
nations, beijing, countries, timor, international, amnesty, states,
chinese, UN, torture, abuses, visit, geneva, indonesia, nigeria\tabularnewline
\cline{2-4} 
 & 0.026651 & 10 & rights, human, china, united, states, resolution, u.s, burma, beijing,
countries, u.n, nations, visit, chinese, foreign, international, record,
commission, european, asean, amnesty, world, state, year,\tabularnewline
\hline 
\multirow{3}{*}{15} & 0.037233 & 6 & vote, opposition, election, seats, results, polling, poll, percent,
voting, won, assembly, voters, elections, democratic, stations, candidates,
national, turnout, parties, parliament, ruling, voted, \tabularnewline
\cline{2-4} 
 & 0.041313 & 5 & mexico, pri, city, elections, prd, zedillo, state, pan, vote, mayor,
mexican, MEX, opposition, cardenas, congress, revolutionary, electoral,
institutional, ruling, election, july, revolution, democratic,\tabularnewline
\cline{2-4} 
 & 0.033671 & 6 & japan, ldp, hashimoto, coalition, prime, democratic, JAP, japanese,
house, parliament, u.s, okinawa, lower, ministry, yen, shinshinto,
liberal, tokyo, ryutaro, sakigake, parties, finance, bases, democrats\tabularnewline
\hline 
\end{tabular}
\par\end{centering}

\protect\caption{SPDP topic quality with the dataset ``International Political News''\label{tab:SPDP-topic-quality-POL}}
\end{table}

\FloatBarrier

\subsection{Basic Parallelization Over Topic\textmd{s}}

\paragraph*{\textmd{I implemented the proposal in Section \ref{sub:Basic-Parallelization:-Over},
a basic parallelism over topics and word-associations (recall that
only one word can be associated since the transformation matrices
$P^{i}$ are identity matrices). Threads are created before Gibbs
sampling initiates, and synchronized after topic and word-association
probabilities are computed. This experiment is only conducted on the
Intel i7 740QM CPU and the NVIDIA GTX460M GPU since the results are
sufficient to show that a single level of parallelism is not efficient.}}

\paragraph*{\textmd{Table \ref{tab:SPDP-Basic-Parallelism} shows the result
on the dataset ``RedState v.s DailyKos''. Just as we expected in
Section \ref{sub:Basic-Parallelization:-Over}, the cost of synchronization
surpasses the benefit of parallelism. The original SPDP running on
one thread is the consistent winner. The result shows that ``easy''
parallelism such as this simply does not work - it does not improve
the running speed of SPDP at all. }}

\begin{table}
\begin{centering}
\begin{tabular}{|c|c|c|}
\hline 
Number of Threads  & $K=32$ & $K=128$\tabularnewline
\hline 
1 (CPU) & 16.510 seconds & 66.230 seconds\tabularnewline
\hline 
2 (CPU) & 72.037 seconds & 109.563 seconds\tabularnewline
\hline 
4 (CPU) & 192.815 seconds & 198.165 seconds\tabularnewline
\hline 
64 (CPU) & \textgreater{} 600 seconds & \textgreater{}600 seconds\tabularnewline
\hline 
$2\times K$ (GPU) & 260 seconds & 334 seconds\tabularnewline
\hline 
\end{tabular}\protect\caption{SPDP basic parallelism result with the dataset ``RedState v.s DailyKos''\label{tab:SPDP-Basic-Parallelism}}

\par\end{centering}

\end{table}

\subsection{Parallelization Over Words}

\paragraph*{\textmd{I implemented the two-level parallel proposal in Section
\ref{sub:Parallelization-Over-Sampling}. The experiments are conducted
on both AMD Radeon HD5850 and NVIDIA GTX460M GPUs, with varying number
of topics. This version will be refered as GPU-SPDP from now on.\label{par:GPU-SPDP}}}

\paragraph*{\textmd{Table \ref{tab:GPU-SPDP-running-time} illustrates the running
time of GPU-SPDP with the dataset ``RedState v.s DailyKos''. Original
SPDP best running times on CPU from Table \ref{tab:SPDP-running-time}
are provided as a single column for comparison. A massive speedup
is obtained even though I did not optimize my code for the specific
GPUs, as explained in Section \ref{par:implementation-why-we-did-not-optimize}.
If the code is further optimized to device-specific parameters, we
may expect a further improvement a few times better than the current
result. With the implementation of Algorithm \ref{alg:Multi-GPU-SPDP-Kernel}
only a fraction of the maximum local memory provided by both devices
is used. Furthermore, the AMD HD5850, which has higher GFLOPs than
the NVIDIA GTX460M, is reporting longer running time. This evidence
shows that we have not used the full potential of these devices.}}

\paragraph*{\textmd{Table \ref{tab:GPU-SPDP-running-time-RD-and-POL} shows the
running time on other datasets with the number of topics set to 32.
Note this is not the optimal number of topics for current implementation
to get the maximum speedup. }}

\begin{table}
\begin{centering}
\begin{tabular}{|c|c|c|c|c|c|}
\hline 
 & \multicolumn{2}{c|}{Running Time Per Iteration} & Original & \multicolumn{2}{c|}{Speed Improvement }\tabularnewline
\hline 
Number of Topics & GTX460M & HD5850 & CPU (Fastest) & GTX460M & HD5850\tabularnewline
\hline 
32 & 0.518s & 0.618s & 16.510s & 3187\% & 2671\%\tabularnewline
\hline 
64 & 0.678s & 0.783s & 35.024s & 5165\% & 4473\%\tabularnewline
\hline 
128 & 1.22s & 2.164s & 66.230s & 5429\% & 3060\%\tabularnewline
\hline 
\end{tabular}
\par\end{centering}

\protect\caption{GPU-SPDP running time with the dataset ``RedState v.s DailyKos''
\label{tab:GPU-SPDP-running-time}}

\end{table}

\begin{table}
\begin{centering}
\begin{tabular}{|>{\centering}p{0.15\textwidth}|>{\centering}p{0.12\textwidth}|>{\centering}p{0.12\textwidth}|>{\centering}p{0.12\textwidth}|>{\centering}p{0.12\textwidth}|>{\centering}p{0.12\textwidth}|}
\hline 
 & \multicolumn{2}{c|}{Running Time Per Iteration} & Original & \multicolumn{2}{c|}{Speed Improvement }\tabularnewline
\hline 
Dataset & GTX460M & HD5850 & CPU (Fastest) & GTX460M & HD5850\tabularnewline
\hline 
Reuters Disasters & 0.940s & 1.146s & 26.621s & 2832\% & 2323\%\tabularnewline
\hline 
International Political News & 2.463 & 3.099s & 86.355s & 3506\% & 2787\%\tabularnewline
\hline 
\end{tabular}
\par\end{centering}

\protect\caption{GPU-SPDP running time with the datasets ``Reuters Disasters'' and
``International Political News'' when $K=32$ \label{tab:GPU-SPDP-running-time-RD-and-POL}}
\end{table}

\FloatBarrier

\begin{figure}
\begin{centering}
\includegraphics[width=1\textwidth]{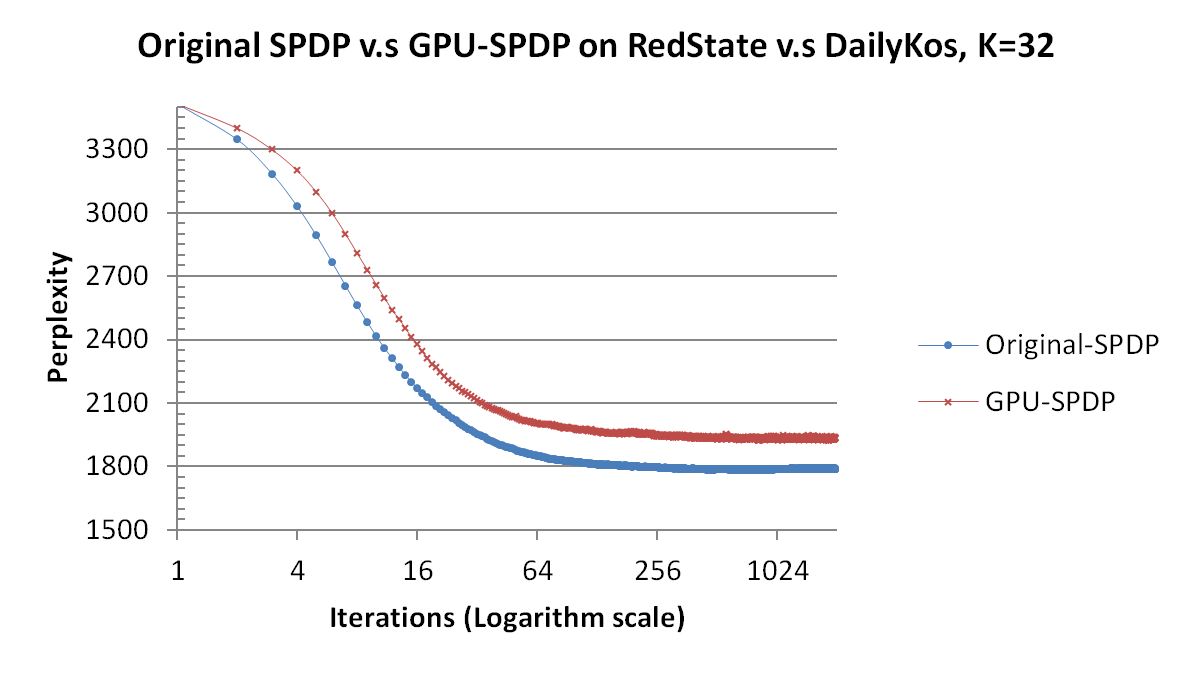}
\par\end{centering}

\protect\caption{GPU-SPDP perplexity v.s Original SPDP with the dataset ``RedState
v.s DailyKos''\label{fig:GPU-SPDP-perplexity-v.s-Original-SPDP}}
\end{figure}

\begin{figure}
\begin{centering}
\includegraphics[width=1\textwidth]{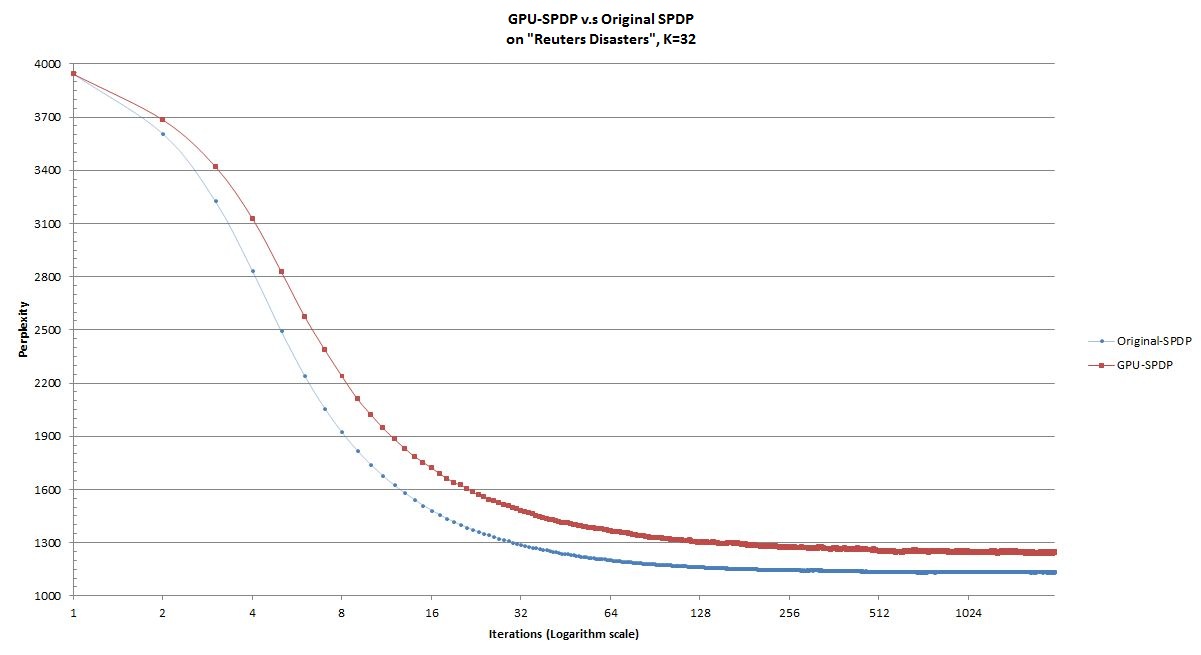}
\par\end{centering}

\protect\caption{GPU-SPDP perplexity v.s Original SPDP with the dataset ``Reuters
Disasters''\label{fig:GPU-SPDP-perplexity-v.s-Original-SPDP-RD}}
\end{figure}

\begin{figure}
\begin{centering}
\includegraphics[width=1\textwidth]{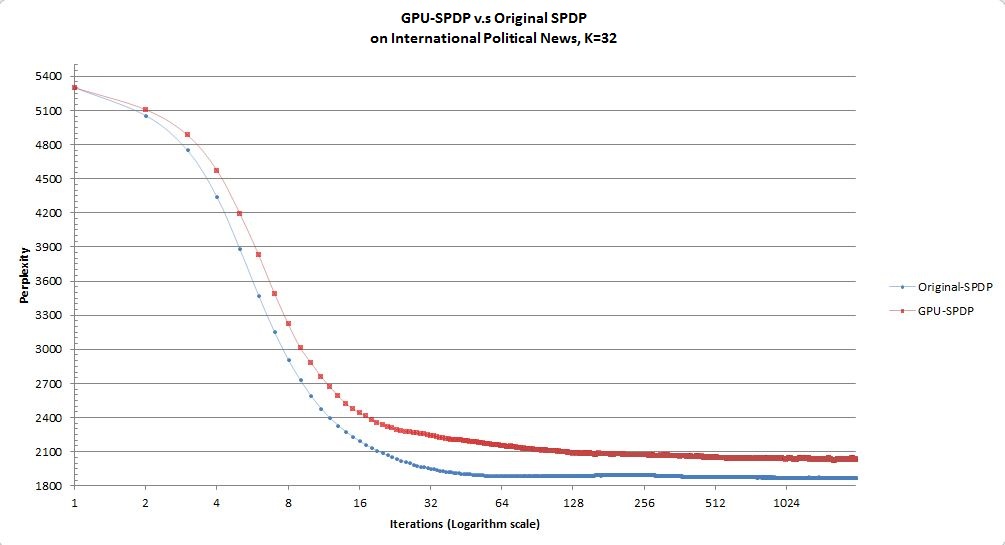}
\par\end{centering}

\protect\caption{GPU-SPDP perplexity v.s Original SPDP with the dataset ``International
Political News''\label{fig:GPU-SPDP-perplexity-v.s-Original-SPDP-POL}}
\end{figure}

\paragraph*{\textmd{Figure \ref{fig:GPU-SPDP-perplexity-v.s-Original-SPDP},
\ref{fig:GPU-SPDP-perplexity-v.s-Original-SPDP-RD}, and \ref{fig:GPU-SPDP-perplexity-v.s-Original-SPDP-POL}
show the comparisons between the perplexity of my algorithm and the
perplexity of the original algorithm for all three datasets as number
of iterations progresses, when number of topics is set to 32. It can
be seen from the figures that my algorithm is only slightly worse
than the original algorithm in terms of perplexity. Table \ref{tab:GPU-SPDP-topic-quality},
\ref{tab:GPU-SPDP-topic-quality-RD}, and \ref{tab:GPU-SPDP-topic-quality-POL}
show the sample outputs of a few topics produced by the algorithm
after 2000 iterations for all three datasets. From the human intepretability
point of view these topics possess high quality because an ordinary
human reader could understand and make sense of the words in each
topic. }}

\paragraph*{\textmd{For example, in the result of the experiment running on dataset
``RedState v.s DailyKos'':}}

\paragraph*{\textmd{Topic 26 shows RedState bloggers, who represent Republicans,
are trying to completely avoid mentioning the issue that US government
and CIA are torturing prisoners in Guantanamo detention camp who they
believe are terrorists. Instead they redirect the topic to international
issues in Pakistan and Israel, progress of peace negotiations in this
area, and terrorism events around the globe. }}

\paragraph*{\textmd{Topic 22 shows bloggers from both websites are addressing
the issue of global financial crisis in approximately the same fashion.
The subtle difference is DailyKos bloggers are looking at the issue
from the government and public point of view (as seen from words ``public,
matching, federal, etc.''), whereas RedState bloggers are more focused
on the market itself (as seen from words ``rate, credit, securities,
risk, bank, (wall) street''). This is coherent to public knowledge
that Democratic Party (that DailyKos represents) is running the government,
and republicans are known to be close to Wall Street and big finance
industry players. }}

\paragraph*{\textmd{Topic 4 is focused on racial and gender equality issues.
RedState bloggers seem to be quite interested by the fact that Barack
Obama, the current US president and leader of Democratic Party, is
an African American. On the other hand DailyKos bloggers cover a range
of issues in this area, treating them as equally important. The name
``Ferraro'' is mentioned because she is known to be a feminist who
has strong opinion and great influence. She is the first female Vice
Presidential candidate representing a major American political party.
The city ``San Francisco'' is mentioned because this is the major
city for racial and gender equality campaigns and protests. }}

\paragraph*{\textmd{Topic 12 is focused on political blogs and online media.
It is very interesting that each of their website names appears as
the top word in this topic. }}

\FloatBarrier

\begin{table}
\begin{centering}
\begin{tabular}{|c|c|c|>{\centering}p{0.5\textwidth}|}
\hline 
Topic ID & Probability & Rank & Top Words (out of top 50)\tabularnewline
\hline 
\multirow{2}{*}{26} & 0.007992 & 30 & israel, bush, president, hamas, north, israeli, state, administration,
korea, peace, rice, palestinian, terrorist, negotiations, east, knesset,
talks, negotiating, enemies, annapolis, south, terrorists\tabularnewline
\cline{2-4} 
 & 0.008642 & 29 & torture, prisoners, tapes, cia, bush, waterboarding, guant, namo,
held, tortured, evidence, destroyed, prisoner, prison, kiriakou, abu,
guantanamo, perino, given, means, detainees, interrogation, gitmo\tabularnewline
\hline 
\multirow{2}{*}{22} & 0.018667 & 12 & fed, market, financial, mortgage, markets, rate, credit, money, securities,
reserve, term, risk, federal, bear, bank, capital, crisis, treasury,
banks, stock, value, street, short, far, rates, long, mortgages\tabularnewline
\cline{2-4} 
 & 0.009010 & 28 & funds, money, matching, public, federal, financing, loan, financial,
bank, spending, fund, credit, mortgage, limits, spend, dollar, means,
market, rate, finance, commission, complaint, cash, opinion\tabularnewline
\hline 
\multirow{2}{*}{4} & 0.014814 & 19 & barack, black, race, left, racist, rich, america, americans, african,
american, chicago, king, racial, democrats, party, racism, flag, ayers,
liberals, wife, jackson, san, francisco, jim, civil, rights\tabularnewline
\cline{2-4} 
 & 0.012534 & 22 & black, white, americans, woman, san, american, african, women, ferraro,
francisco, king, racist, jackson, barack, racism, liberals, race,
radio, gibson, racial, fox, civil, reilly, rights, challenge\tabularnewline
\hline 
\multirow{2}{*}{12} & 0.012719 & 23 & redstate, post, book, blog, feith, online, bloggers, read, blogs,
community, readers, blogger, decision, milbank, discussion, blogging,
left, writing, kos, check, moveon, posts, wing, write, sites, reading\tabularnewline
\cline{2-4} 
 & 0.018648 & 13 & kos, daily, bloggers, blog, progressive, media, politics, blogs, read,
writers, diaries, diary, strike, writing, community, local, blogging,
netroots, write, traditional, blogger, lehane, fun, kossack\tabularnewline
\hline 
\end{tabular}
\par\end{centering}

\protect\caption{GPU-SPDP topic quality with the dataset ``RedState v.s DailyKos''\label{tab:GPU-SPDP-topic-quality}}

\end{table}

\begin{table}
\begin{centering}
\begin{tabular}{|c|c|c|>{\centering}p{0.5\textwidth}|}
\hline 
Topic ID & Probability & Rank & Top Words (out of top 50)\tabularnewline
\hline 
\multirow{4}{*}{24} & 0.008279 & 30 & sri, lanka, military, colombo, north, year, percent, tamil, rebels,
defence, SRILAN, rupees, war, refugees, growth, northern, fighting,
ltte, tigers, lankan, rebel, power, spending, jayawardena, liberation,\tabularnewline
\cline{2-4} 
 & 0.016090 & 15 & government, bank, flood, budget, percent, year, central, finance,
zlotys, belka, relief, reconstruction, cimoszewicz, tax, loan, 2.5,
credit, funds, inflation, economy, borrowing, deficit, borrow, zloty\tabularnewline
\cline{2-4} 
 & 0.014310 & 23 & fire, hotel, floor, building, rescue, smoke, thai, thailand, bangkok,
trapped, death, broke, blaze, THAIL, top, locked, firefighters, jumped,
royal, windows, toll, helicopters, started, escape, hours,\tabularnewline
\cline{2-4} 
 & 0.018745 & 15 & fire, firefighters, blaze, california, los, homes, angeles, san, contained,
acres, winds, battling, destroyed, burned, forest, hectares, wildfire,
malibu, burning, flames, diego, percent, southern,\tabularnewline
\hline 
\multirow{4}{*}{15} & 0.028165 & 8 & storm, coast, winds, cyclone, mph, hit, miles, expected, sea, east,
hours, heavy, km, weather, west, meteorological, shipping, typhoon,
tropical, coastal, damage, flooding, rain, month, moving, strong,\tabularnewline
\cline{2-4} 
 & 0.006866 & 31 & amsterdam, dakota, NETH, coastguard, north, senegal, tourists, african,
sea, television, den, bodies, storm, transport, helder, tambacouda,
dc-3, dakar, islands, schiphol, pleasure, national, centre, \tabularnewline
\cline{2-4} 
 & 0.052266 & 4 & winds, storm, mph, tropical, hurricane, typhoon, miles, expected,
kph, moving, coast, north, hours, cuba, east, weather, km, forecasters,
center, flooding, west, shipping, islands, USA, japan, heavy,\tabularnewline
\cline{2-4} 
 & 0.066367 & 4 & hurricane, storm, winds, mph, tropical, miles, north, coast, center,
kph, km, forecasters, expected, west, national, moving, gmt, atlantic,
florida, east, sustained, edt, warning, maximum, island, centre\tabularnewline
\hline 
\end{tabular}
\par\end{centering}

\protect\caption{GPU-SPDP topic quality with the dataset ``Reuters Disasters''\label{tab:GPU-SPDP-topic-quality-RD}}
\end{table}

\begin{table}
\begin{centering}
\begin{tabular}{|c|c|c|>{\centering}p{0.5\textwidth}|}
\hline 
Topic ID & Probability & Rank & Top Words (out of top 50)\tabularnewline
\hline 
\multirow{3}{*}{20} & 0.023948 & 9 & kabila, zaire, kinshasa, rebels, rebel, laurent, forces, capital,
sese, alliance, ZAIRE, zairean, city, democratic, seko, lubumbashi,
africa, kisangani, south, army, transitional, power, troops, liberation\tabularnewline
\cline{2-4} 
 & 0.015933 & 17 & colombia, samper, drug, colombian, COL, ernesto, extradition, campaign,
bogota, la, valdivieso, serpa, carlos, farc, office, banzer, bolivia,
lords, cartel, colombians, leftist, medellin, ban, charges,\tabularnewline
\cline{2-4} 
 & 0.020348 & 12 & deng, beijing, chinese, jiang, communist, china, xiaoping, CHINA,
death, mao, li, zemin, square, power, chen, tiananmen, died, paramount,
shanghai, reforms, funeral, central, army, leader, xinhua, chief,\tabularnewline
\hline 
\multirow{3}{*}{8} & 0.025943 & 7 & rights, human, commonwealth, u.n, group, international, report, britain,
activists, amnesty, visit, united, abuses, countries, commission,
ogoni, nations, mission, u.s, states, detainees, torture, team, \tabularnewline
\cline{2-4} 
 & 0.035477 & 3 & rights, human, china, u.n, resolution, commission, united, nations,
east, countries, international, amnesty, report, UN, timor, states,
geneva, beijing, torture, abuses, visit, foreign, indonesia, european\tabularnewline
\cline{2-4} 
 & 0.030808 & 5 & rights, human, china, chinese, beijing, united, tibet, resolution,
u.n, states, visit, u.s, international, lama, countries, nations,
dalai, commission, foreign, CHINA, record, france, world, french, \tabularnewline
\hline 
\end{tabular}
\par\end{centering}

\protect\caption{GPU-SPDP topic quality with the dataset ``International Political
News''\label{tab:GPU-SPDP-topic-quality-POL}}
\end{table}

\paragraph*{\textmd{Examining the result closely, it can be noticed that some
problems do exist in this version of approximation. The top topic
(which is not shown here) has excessive probability in both groups,
whereas the rest of the topics have much smaller probabilities compared
to the result from original SPDP algorithm. This can be explained
by the error accumulation due to parallel accessing same count variable
through many iterations of approximation. A solution for this through
word re-ordering is proposed in Section \ref{sub:Parallelization-With-Better}.
In the next experiment, this is implemented in conjunction with this
version of the approximation.}}

\FloatBarrier

\subsection{Effect of Re-ordering Words}

\begin{figure}
\begin{centering}
\includegraphics[width=1\textwidth]{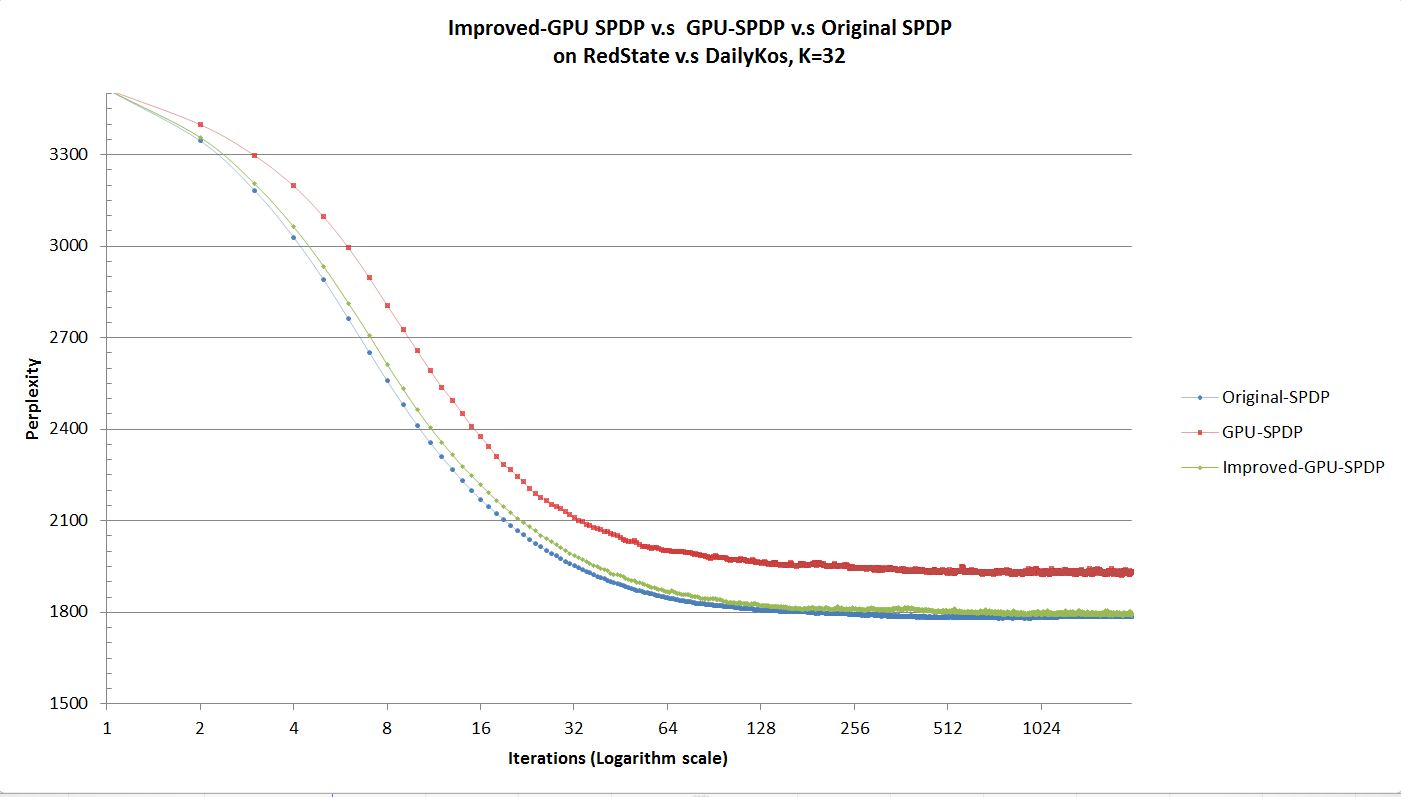}
\par\end{centering}

\protect\caption{Improved GPU-SPDP perplexity with the dataset ``RedState v.s DailyKos''\label{fig:Improved-GPU-SPDP-Perplexity}}

\end{figure}

\begin{figure}
\begin{centering}
\includegraphics[width=1\textwidth]{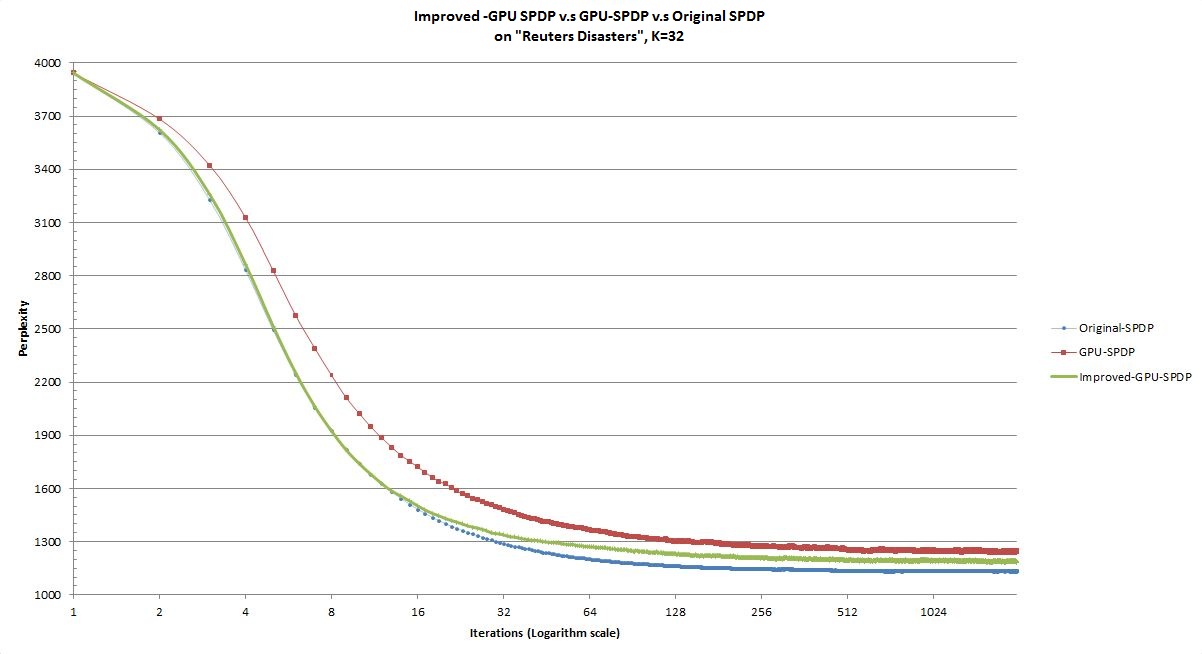}
\par\end{centering}

\protect\caption{Improved GPU-SPDP perplexity with the dataset ``RedState v.s DailyKos''\label{fig:Improved-GPU-SPDP-Perplexity-RD}}
\end{figure}

\begin{figure}
\begin{centering}
\includegraphics[width=1\textwidth]{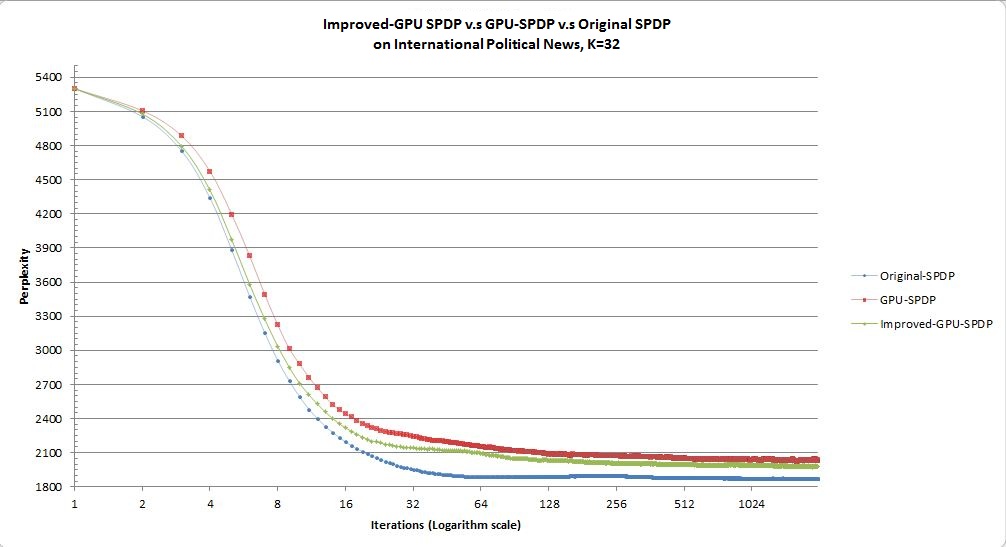}
\par\end{centering}

\protect\caption{Improved GPU-SPDP perplexity with the dataset ``RedState v.s DailyKos''\label{fig:Improved-GPU-SPDP-Perplexity-POL}}
\end{figure}

\paragraph*{\textmd{As proposed in Section \ref{sub:Parallelization-With-Better},
it is expected that the amount of error can be reduced by rearranging
the order of words before dispatching them to the GPU. In this experiment,
this feature is added to the implementation, while everything else
is as same as the previous experiment. Figure \ref{fig:Improved-GPU-SPDP-Perplexity},
\ref{fig:Improved-GPU-SPDP-Perplexity-RD}, and \ref{fig:Improved-GPU-SPDP-Perplexity-POL}
show the perplexity convergence of the improved version compared to
the perplexity convergence result from GPU-SPDP and the original SPDP.
Just as we did in the previous experiments, Table \ref{tab:Improved-GPU-SPDP-Topic-Quality},
\ref{tab:Improved-GPU-SPDP-Topic-Quality-RD}, and \ref{tab:Improved-GPU-SPDP-Topic-Quality-POL}
are included to show the quality of a few selected topics.}}

\paragraph*{\textmd{The perplexity is indeed reduced to a value almost as same
as the original SPDP algorithm. The topic probabilities are in the
correct range. The topic qualities are also slightly improved.}}

\paragraph*{\textmd{This means, now we have an algorithm which is almost as good
as the original SPDP, but it runs about 30 to 50 times faster on a
single consumer-grade cheap GPU which costs less than \$150 USD, before
any optimization, or using the power of multiple GPUs distribution
framework as proposed in Section \ref{sub:All-in-one:-Putting-everything}.}}

\FloatBarrier

\begin{table}
\begin{centering}
\begin{tabular}{|c|c|c|>{\centering}p{0.5\textwidth}|}
\hline 
Topic ID & Probability & Rank & Top Words (out of top 50)\tabularnewline
\hline 
\multirow{2}{*}{13} & 0.014364 & 22 & campaign, money, york, corruption, paterson, murtha, foster, franken,
culture, pay, democrat, party, david, office, mayor, giving, state,
lobbyist, fund, wife, organizations, governor, rnc, johnson\tabularnewline
\cline{2-4} 
 & 0.011071 & 30 & debbie, dccc, schultz, wasserman, florida, blue, fl, chair, joe, garcia,
red, trauner, south, gary, ileana, party, rahm, martinez, support,
campaign, candidates, democratic, races, majority, netroots,\tabularnewline
\hline 
\multirow{2}{*}{10} & 0.014269 & 23 & senate, senator, legislation, vote, sen, committee, house, voted,
floor, reid, number, democrats, mcconnell, votes, senators, passed,
act, resolution, fisa, dodd, minority, read, republicans, update,
mitch\tabularnewline
\cline{2-4} 
 & 0.040049 & 6 & senate, vote, senators, fisa, dodd, amnesty, house, republicans, senator,
reid, floor, votes, update, amendment, law, democrats, filibuster,
committee, telco, act, cloture, debate, retroactive, amendments\tabularnewline
\hline 
\multirow{2}{*}{27} & 0.013413 & 15 & oil, energy, global, gas, prices, food, change, production, gasoline,
nuclear, warming, fuel, price, going, cost, blackhedd, climate, domestic,
francis, year, companies, crude, ethanol, drive, cianfrocca\tabularnewline
\cline{2-4} 
 & 0.019349 & 17 & energy, global, climate, oil, warming, water, change, science, years,
california, earth, gas, environment, scientists, coal, environmental,
park, world, epa, action, year, land, clean, big, millions, power\tabularnewline
\hline 
\multirow{2}{*}{18} & 0.024284 & 11 & iraq, war, troops, qaeda, iraqi, military, surge, afghanistan, security,
american, petraeus, general, strategy, forces, progress, soldiers,
saddam, year, said, sadr, success, political, withdrawal, iraqis\tabularnewline
\cline{2-4} 
 & 0.031108 & 11 & iraq, war, troops, military, kristol, bush, iraqi, american, surge,
americans, violence, year, government, years, soldiers, forces, afghanistan,
invasion, iraqis, country, pakistan, progress, administration\tabularnewline
\hline 
\end{tabular}
\par\end{centering}

\protect\caption{Improved-GPU-SPDP topic quality with the dataset ``RedState v.s DailyKos''\label{tab:Improved-GPU-SPDP-Topic-Quality}}

\end{table}

\begin{table}
\begin{centering}
\begin{tabular}{|c|c|c|>{\centering}p{0.5\textwidth}|}
\hline 
Topic ID & Probability & Rank & Top Words (out of top 50)\tabularnewline
\hline 
\multirow{4}{*}{21} & 0.034472 & 9 & philippines, manila, storm, PHLNS, winds, philippine, coast, miles,
mph, south, expected, island, hours, sea, weather, heavy, luzon, strong,
typhoon, east, north, tropical, shipping, southern, flooding,\tabularnewline
\cline{2-4} 
 & 0.015062 & 24 & agency, news, turkish, anatolian, TURK, quoted, istanbul, state-run,
province, town, reported, crashed, southeast, fighter, dead, local,
western, further, km, crashes, military, northeast, central, evening\tabularnewline
\cline{2-4} 
 & 0.059273 & 4 & storm, mph, winds, hurricane, tropical, expected, kph, miles, weather,
hours, typhoon, coast, north, moving, km, cuba, flooding, forecasters,
central, east, south, lili, center, western, islands, west,\tabularnewline
\cline{2-4} 
 & 0.108872 & 2 & hurricane, storm, winds, mph, tropical, miles, expected, north, coast,
kph, center, hours, km, forecasters, west, moving, east, hortense,
islands, national, weather, florida, flooding, shipping, gmt,\tabularnewline
\hline 
\multirow{4}{*}{9} & 0.048960 & 6 & ship, crew, vessel, port, missing, boat, cargo, sea, carrying, sank,
rescued, coast, officials, ships, miles, rescue, shipping, capsized,
search, members, reported, official, island, reuters, tonnes, navy\tabularnewline
\cline{2-4} 
 & 0.058029 & 6 & ship, port, coast, crew, sea, vessel, missing, cargo, boat, carrying,
greek, island, sank, rescue, coastguard, carrier, board, officials,
service, capsized, rescued, guard, aground, vessels, shipping,\tabularnewline
\cline{2-4} 
 & 0.036614 & 8 & ship, vessel, sheep, crew, live, port, australia, missing, sank, coast,
australian, boat, trade, cargo, u.s, carrying, animal, fire, uniceb,
ocean, rescued, members, search, livestock, cattle, capsized\tabularnewline
\cline{2-4} 
 & 0.037560 & 6 & coast, guard, u.s, crew, ship, helicopter, search, air, navy, boat,
crashed, board, missing, members, marine, rescue, military, force,
miles, spokesman, vessel, sea, crash, island, km, helicopters, rescued\tabularnewline
\hline 
\end{tabular}
\par\end{centering}

\protect\caption{Improved-GPU-SPDP topic quality with the dataset ``Reuters Disasters''\label{tab:Improved-GPU-SPDP-Topic-Quality-RD}}
\end{table}

\begin{table}
\begin{centering}
\begin{tabular}{|c|c|c|>{\centering}p{0.5\textwidth}|}
\hline 
Topic ID & Probability & Rank & Top Words (out of top 50)\tabularnewline
\hline 
\multirow{3}{*}{12} & 0.013602 & 16 & rights, human, commonwealth, u.n, group, international, nigeria, amnesty,
activists, ogoni, report, abuses, groups, commission, saro-wiwa, mission,
detainees, visit, team, violations, detained, nations,\tabularnewline
\cline{2-4} 
 & 0.012239 & 24 & u.n, taleban, rights, women, human, kabul, afghanistan, nations, nigeria,
united, UN, afghan, islamic, special, mission, pakistan, visit, cambodia,
military, western, north, international, masood, militia\tabularnewline
\cline{2-4} 
 & 0.017282 & 14 & deng, china, jiang, xiaoping, chinese, beijing, communist, death,
mao, CHINA, zemin, square, died, army, paramount, economic, tiananmen,
health, funeral, power, military, state, reforms, shanghai, peng,\tabularnewline
\hline 
\multirow{3}{*}{25} & 0.056171 & 3 & percent, budget, economic, year, tax, bank, finance, economy, growth,
oil, industry, money, market, sector, foreign, service, policy, business,
financial, deficit, spending, public, development, state,\tabularnewline
\cline{2-4} 
 & 0.067658 & 2 & budget, economic, tax, economy, year, bank, deficit, fiscal, spending,
market, finance, growth, debt, plan, public, social, newsroom, companies,
sector, cut, congress, jobs, inflation, financial, fund,\tabularnewline
\cline{2-4} 
 & 0.073056 & 2 & budget, tax, bank, economic, policy, economy, fiscal, growth, financial,
spending, finance, cuts, business, rate, inflation, investment, market,
deficit, reform, interest, sector, rates, current, industry\tabularnewline
\hline 
\end{tabular}
\par\end{centering}

\protect\caption{Improved-GPU-SPDP topic quality with the dataset ``International
Political News''\label{tab:Improved-GPU-SPDP-Topic-Quality-POL}}
\end{table}

\FloatBarrier

\subsection{Multi-GPU Distributed Parallelism}

\paragraph*{\textmd{Finally, I implemented the ultimate algorithm MGPU-DP-SPDP,
and tested it using the same configuration as used in previous experiments
except all five Radeon HD5850 GPUs on the AMD machine are used this
time. }}

\paragraph*{\textmd{The running time analysis is shown in Table \ref{tab:MGPU-DP-SPDP-Running-Time}.
For the columns representing running time of the original SPDP and
GPU-SPDP, we used data from Table \ref{tab:GPU-SPDP-running-time}.
The speed improvement over GPU-SPDP is sublinear and sometimes much
lower than the maximum value (500\%) because of two reasons. First,
for safety reasons we used atomic locks for incrementing and decrementing
variables (i.e using OpenCL primitives). For some reasons, this is
necessary for AMD GPUs when multiple GPUs are used, otherwise the
program would crash. We believe atomic locks are not necessary, and
the result could be much more impressive if the experiment is conducted
on multiple professional NVIDIA GPUs, which do not have these problems.
However, we do not have the equipment at our disposal. Second, the
code is not optimized, and a substantial amount of operation is wasted
on memory transfer. Nonetheless, the experiment is sufficient to show
that MGPU-DP-SPDP is a scalable algorithm that could improve the running
speed of GPU-SPDP dramatically, sublinear to the number of GPUs available
to the machine. }}

\begin{table}
\begin{centering}
\begin{tabular}{|c|c|c|c|c|c|}
\hline 
 & \multicolumn{2}{c|}{Running Time Per Iteration} & Original & \multicolumn{2}{c|}{Speed Improvement }\tabularnewline
\hline 
Number of Topics & $5\times$HD5850 & $1\times$HD5850 & i7 740QM & To Original & To GPU-SPDP\tabularnewline
\hline 
32 & 0.183s & 0.618s & 16.510s & 9022\% & 3377\%\tabularnewline
\hline 
64 & 0.412s & 0.783s & 35.024s & 8501\% & 1900\%\tabularnewline
\hline 
128 & 0.95s & 2.164s & 66.230s & 6971\% & 2278\%\tabularnewline
\hline 
\end{tabular}
\par\end{centering}

\protect\caption{MGPU-DP-SPDP Running Time with the dataset ``RedState v.s DailyKos''
\label{tab:MGPU-DP-SPDP-Running-Time}}

\end{table}

\begin{table}
\begin{centering}
\begin{tabular}{|>{\centering}p{0.16\textwidth}|>{\centering}p{0.12\textwidth}|>{\centering}p{0.12\textwidth}|>{\centering}p{0.12\textwidth}|>{\centering}p{0.12\textwidth}|>{\centering}p{0.12\textwidth}|}
\hline 
 & \multicolumn{2}{c|}{Running Time Per Iteration} & Original & \multicolumn{2}{c|}{Speed Improvement }\tabularnewline
\hline 
Datasets & $5\times$HD5850 & $1\times$HD5850 & CPU (Fastest) & To Original & To GPU-SPDP\tabularnewline
\hline 
Reuters Disasters & 0.611s & 1.146s & 26.621s & 4356\% & 187\%\tabularnewline
\hline 
International Political News & 1.517 & 3.099s & 86.355s & 5692\% & 204\%\tabularnewline
\hline 
\end{tabular}
\par\end{centering}

\protect\caption{MGPU-DP-SPDP Running Time with the dataset ``Reuters Disasters''
and ``International Political News'' when $K=32$ \label{tab:MGPU-DP-SPDP-Running-Time-others}}
\end{table}

\FloatBarrier

\begin{figure}
\begin{centering}
\includegraphics[width=1\textwidth]{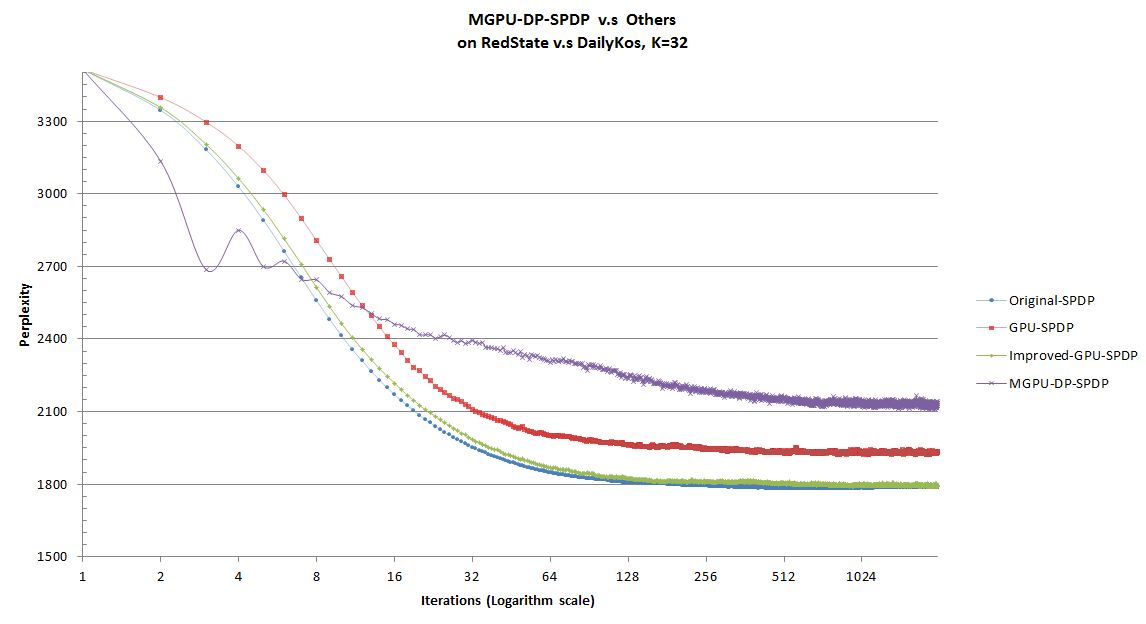}
\par\end{centering}

\protect\caption{MGPU-DP-SPDP perplexity convergence with the dataset ``RedState v.s
DailyKos'' \label{fig:MGPU-DP-SPDP-Perplexity-Converge}}

\end{figure}

\paragraph*{\textmd{The convergence in perplexity is shown in Figure \ref{fig:MGPU-DP-SPDP-Perplexity-Converge}.
The perplexity is not as good as the original algorithm, but it is
fairly comparable, and words for each topic still look sensible. This
is because some topic probabilities are off-scale, which is discussed
in the next section.}}

\begin{figure}
\begin{centering}
\includegraphics[width=1\textwidth]{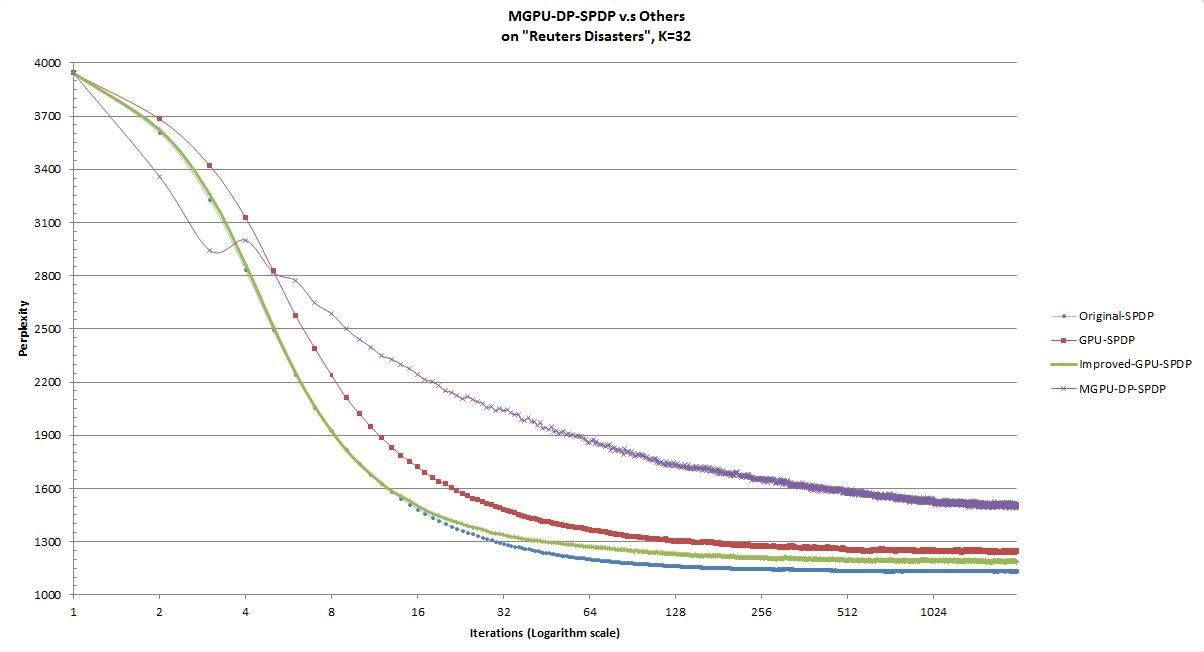}
\par\end{centering}

\protect\caption{MGPU-DP-SPDP perplexity convergence with the dataset ``Reuters Disasters''
\label{fig:MGPU-DP-SPDP-Perplexity-Converge-RD}}
\end{figure}

\begin{figure}
\begin{centering}
\includegraphics[width=1\textwidth]{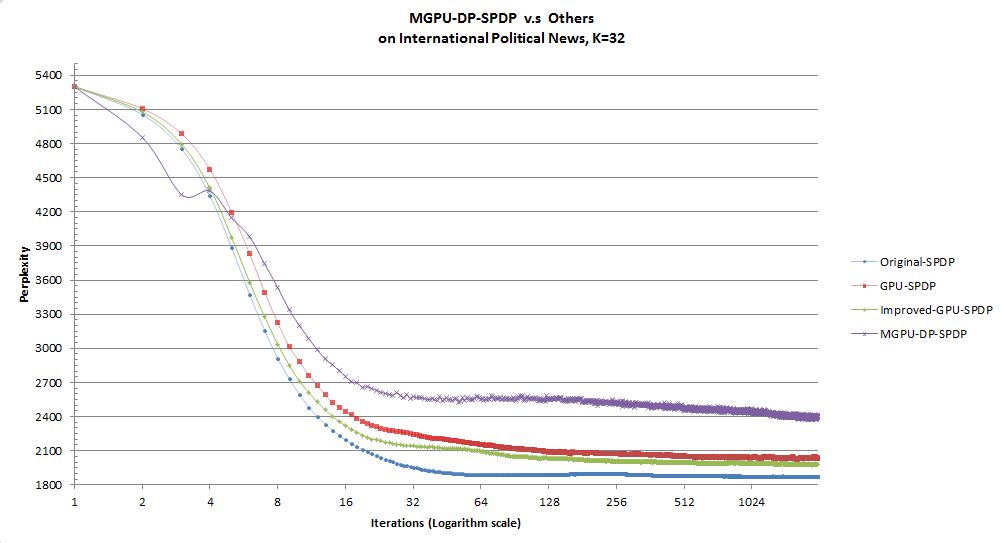}
\par\end{centering}

\protect\caption{MGPU-DP-SPDP perplexity convergence with the dataset ``International
Political News'' \label{fig:MGPU-DP-SPDP-Perplexity-Converge-POL}}
\end{figure}

\FloatBarrier

\paragraph*{\textmd{The topic quality is shown in Table \ref{tab:MGPU-DP-SPDP-Topic-Quality},
\ref{tab:Improved-GPU-SPDP-Topic-Quality-RD}, and \ref{tab:MGPU-DP-SPDP-Topic-Quality-POL}.
Occasionally strange words may appear, but overall the result is interesting,
informative, and sensible.}}

\begin{table}
\begin{centering}
\begin{tabular}{|c|c|c|>{\centering}p{0.5\textwidth}|}
\hline 
Topic ID & Probability & Rank & Top Words (out of top 50)\tabularnewline
\hline 
\multirow{2}{*}{21} & 0.009194 & 12 & iraq, iraqi, surge, petraeus, afghanistan, military, war, sadr, forces,
security, success, troops, withdrawal, baghdad, maliki, sunni, taliban,
strategy, groups, anbar, counterinsurgency, shi, progress\tabularnewline
\cline{2-4} 
 & 0.014949 & 7 & iraq, troops, iraqi, kristol, military, war, surge, forces, violence,
occupation, soldiers, invasion, iraqis, basra, success, lives, killed,
cheney, sadr, security, mission, british, maliki, army, saddam,\tabularnewline
\hline 
\multirow{2}{*}{30} & 0.009467 & 11 & tax, care, taxes, insurance, health, state, private, economic, reform,
federal, jobs, businesses, plan, san, costs, schwarzenegger, francisco,
government, increases, revenue, taxpayers, amt, spending, city,\tabularnewline
\cline{2-4} 
 & 0.007422 & 23 & health, care, tax, insurance, reform, economy, taxes, coverage, cut,
legislation, companies, government, state, private, bills, safety,
fund, harold, programs, profit, corporations, liveblogging, plan,\tabularnewline
\hline 
\multirow{2}{*}{19} & 0.007533 & 22 & energy, oil, prices, gas, tax, gasoline, price, global, jobs, economic,
cianfrocca, carbon, crude, company, companies, wheat, francis, warming,
farmers, profits, costs, methanol, industrial, motor,\tabularnewline
\cline{2-4} 
 & 0.008073 & 17 & oil, park, epa, coal, safety, gas, global, wildlife, stickler, energy,
industry, stimulus, administration, hrc, interior, environmental,
plants, emissions, companies, waiver, environmentalists, economy,\tabularnewline
\hline 
\multirow{2}{*}{22} & 0.008227 & 17 & israel, iran, north, bush, israeli, world, weapons, military, palestinian,
rice, nuclear, nations, foreign, president, terrorist, negotiating,
peace, talks, korea, administration, knesset, borders, \tabularnewline
\cline{2-4} 
 & 0.007767 & 19 & bush, iran, nuclear, pakistan, bhutto, musharraf, minister, weapons,
assassination, military, policy, terrorists, war, arms, blah, nato,
prime, iranian, arabia, pakistani, resolution, oil, interview,\tabularnewline
\hline 
\end{tabular}
\par\end{centering}

\protect\caption{MGPU-DP-SPDP topic quality with the dataset ``RedState v.s DailyKos''
\label{tab:MGPU-DP-SPDP-Topic-Quality}}
\end{table}

\begin{table}
\begin{centering}
\begin{tabular}{|c|c|c|>{\centering}p{0.5\textwidth}|}
\hline 
Topic ID & Probability & Rank & Top Words (out of top 50)\tabularnewline
\hline 
\multirow{4}{*}{19} & 0.015744 & 10 & fire, oil, pipeline, occidental, drilling, blaze, refinery, forest,
bangladesh, ongc, crude, refineries, rig, energy, assam, petroleum,
ltd, sylhet, extinguish, burst, gas, bhandari, production, bazar,\tabularnewline
\cline{2-4} 
 & 0.018931 & 9 & fire, refinery, oil, company, plant, irish, explosion, ireland, production,
leak, newsroom, gas, chemical, unit, pipeline, blaze, tonnes, rafinerska,
japanese, cork, shut, illegal, trawler, a.s, hoechst,\tabularnewline
\cline{2-4} 
 & 0.029501 & 3 & fire, oil, refinery, plant, company, production, singapore, toyota,
gas, corp, petroleum, mobil, irish, operations, taiwan, ltd, plants,
leak, fishing, unit, spokesman, aisin, pipeline, taoyuan, explosion\tabularnewline
\cline{2-4} 
 & 0.028317 & 4 & fire, refinery, plant, oil, gas, company, unit, production, explosion,
corp, pipeline, gasoline, energy, shut, tosco, barrels, bpd, crude,
natural, facility, rig, drilling, texaco, units, platform, toyota\tabularnewline
\hline 
\multirow{4}{*}{5} & 0.010241 & 19 & norwegian, spitzbergen, plane, miners, longyear, arctic, mining, mountain,
norway, russian, crashed, barentsburg, airliner, tu-154, NORW, land,
tupolev, ntb, recorder, ukrainian, treaty, russians,\tabularnewline
\cline{2-4} 
 & 0.009434 & 23 & insurance, shroud, montserrat, turin, cathedral, cloth, london, blaze,
caribbean, pzu, insured, claims, volcano, re, glass, cost, guarini,
trematore, lloyd, christ, roman, royal, miracle, pay, image,\tabularnewline
\cline{2-4} 
 & 0.012325 & 19 & vietnam, insurance, storms, mekong, hanoi, claims, cambodia, rice,
houses, ho, cross, delta, chi, losses, flooding, cost, year, maoming,
july, compensation, zhanjiang, laos, company, picc, winbond, \tabularnewline
\cline{2-4} 
 & 0.012907 & 16 & earthquake, losses, authority, market, cea, insurers, insured, homeowners,
claims, companies, industry, property, coverage, pay, insurance, bonds,
risk, farm, butler, plan, allstate, quackenbush, premiums,\tabularnewline
\hline 
\end{tabular}
\par\end{centering}

\protect\caption{MGPU-DP-SPDP topic quality with the dataset ``Reuters Disasters''\label{tab:MGPU-DP-SPDP-Topic-Quality-RD}}
\end{table}

\begin{table}
\begin{centering}
\begin{tabular}{|c|c|c|>{\centering}p{0.5\textwidth}|}
\hline 
Topic ID & Probability & Rank & Top Words (out of top 50)\tabularnewline
\hline 
\multirow{3}{*}{13} & 0.005928 & 27 & burundi, hutu, buyoya, army, rebels, coup, regional, killed, ntibantunganya,
frodebu, pierre, bujumbura, BURUN, talks, tanzania, imposed, embargo,
cndd, tutsi, civilians, 150,000, tutsis, nyangoma, rpf,\tabularnewline
\cline{2-4} 
 & 0.009565 & 17 & army, rebels, rebel, epr, talks, guatemala, zapatista, guerrilla,
guerrero, mexican, leftist, zapatistas, southern, marcos, guatemalan,
urng, chiapas, peace, popular, zedillo, arzu, MEX, city, oaxaca,\tabularnewline
\cline{2-4} 
 & 0.014908 & 5 & deng, jiang, chinese, communist, xiaoping, death, mao, CHINA, zemin,
square, paramount, tiananmen, shanghai, funeral, power, corruption,
zedong, peng, yuan, chen, leadership, zhao, body, source, mourning\tabularnewline
\hline 
\multirow{3}{*}{26} & 0.034677 & 3 & budget, percent, economic, year, finance, tax, manuel, economy, market,
sector, deficit, oil, spending, revenue, exchange, development, service,
privatisation, expenditure, rate, investment, programme,\tabularnewline
\cline{2-4} 
 & 0.033753 & 2 & budget, tax, economic, deficit, spending, fiscal, economy, year, cut,
debt, reforms, growth, measures, sector, jobs, social, newsroom, taxes,
rate, bank, increase, oil, rates, fund, cuts, expected, finance,\tabularnewline
\cline{2-4} 
 & 0.034065 & 2 & fiscal, spending, percent, economy, policy, cuts, inflation, rates,
budget, growth, rate, market, year, bank, cut, economic, reserve,
deficit, savings, economists, markets, 1996/97, investment, current, \tabularnewline
\hline 
\end{tabular}
\par\end{centering}

\protect\caption{MGPU-DP-SPDP topic quality with the dataset ``International Political
News'' \label{tab:MGPU-DP-SPDP-Topic-Quality-POL}}
\end{table}

\FloatBarrier

\subsection{Multi-GPU Distributed Parallelism: Effect of Duplicating Training
Data}

\begin{figure}
\begin{centering}
\includegraphics[width=1\textwidth]{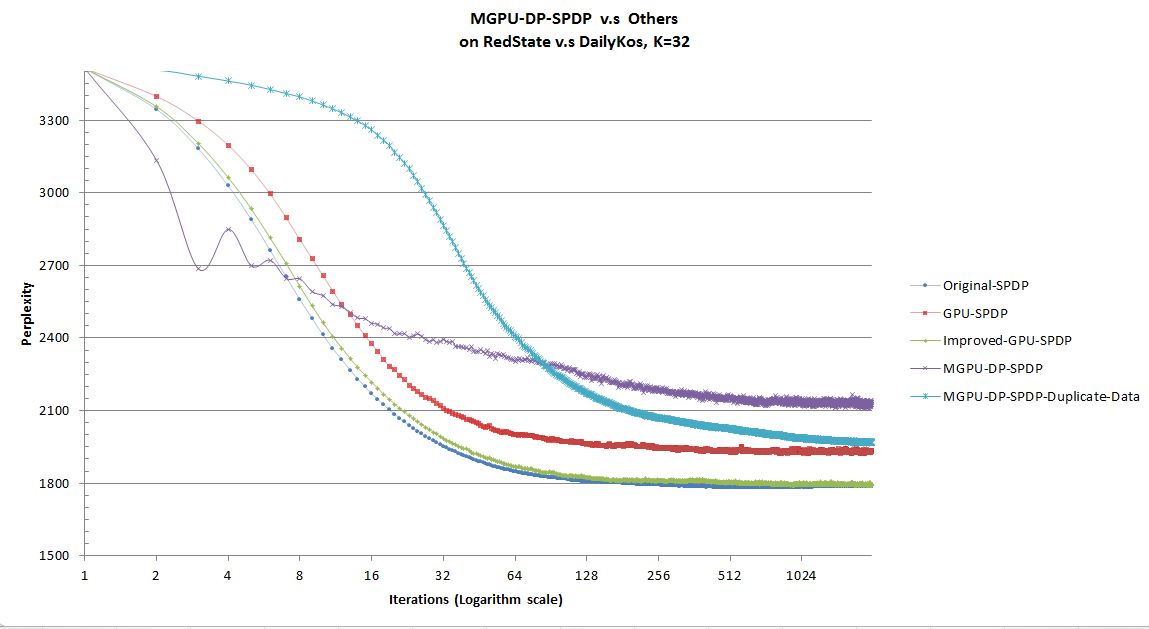}
\par\end{centering}

\protect\caption{MGPU-DP-SPDP perplexity convergence with the duplicated dataset ``RedState
v.s DailyKos'' \label{fig:MGPU-DP-SPDP-Perplexity-Converge-dup}}
\end{figure}

\paragraph*{\textmd{It was found in one of my early implementations of MGPU-DP-SPDP
(which has slightly worse error correction mechanism) that duplicating
training data is sometimes crucial to get topic probabilities which
do not look ridiculous. In my current version, duplicating the training
data is no longer necessary. I found duplicating the data once could
slightly improve the correctness of topic probabilities and the topic
qualities particularly in a small document collection, which unsurprisingly
results a better perplexity. Due to hardware limitations, I am unable
to verify if more than one duplicate is required to improve the perplexity
for a large number of GPUs, though I believe with good implementation
of error correction and optimization, the need of improving the perplexity
from duplicating the training data can be completely eliminated. Figure
\ref{fig:MGPU-DP-SPDP-Perplexity-Converge-dup} shows the perplexity
comparison after duplicating the data once for the ''RedState v.s
DailyKos'' dataset. Table \ref{tab:MGPU-DP-SPDP-Topic-Quality-dup}
shows the topic qualities.}}

\FloatBarrier

\begin{table}
\begin{centering}
\begin{tabular}{|c|c|c|>{\centering}p{0.5\textwidth}|}
\hline 
Topic ID & Probability & Rank & Top Words (out of top 50)\tabularnewline
\hline 
\multirow{2}{*}{4} & 0.029230 & 17 & trade, china, chavez, states, united, economic, policy, world, chinese,
comes, russia, foreign, power, know, concerning, american, colombia,
read, putin, thanks, hugo, regime, freedom, position, castro,\tabularnewline
\cline{2-4} 
 & 0.007415 & 32 & trade, economic, united, china, chavez, american, tremendous, know,
states, policy, world, weber, nader, murray, safety, miner, neo, chinese,
jones, foreign, comes, company, power, msha, canyon\tabularnewline
\hline 
\multirow{2}{*}{30} & 0.023349 & 24 & speech, human, rights, religious, religion, conservatives, jesus,
liberals, faith, god, men, moral, society, constitution, question,
evidence, research, certain, truth, man, merely,\tabularnewline
\cline{2-4} 
 & 0.015697 & 27 & religious, rights, religion, human, speech, question, faith, jesus,
constitution, god, cell, stem, democracy, word, social, conservatives,
church, research, progress, death, society, truth, merely, moral,\tabularnewline
\hline 
\multirow{2}{*}{19} & 0.041756 & 6 & romney, mccain, huckabee, thompson, rudy, fred, mitt, think, campaign,
paul, iowa, republican, mike, john, ron, conservative, giuliani, gop,
candidate, guy, going, state, said, debate, hampshire\tabularnewline
\cline{2-4} 
 & 0.021821 & 21 & romney, mccain, huckabee, mitt, rudy, paul, thompson, think, fred,
republican, campaign, mike, gop, giuliani, iowa, john, state, going,
ron, candidate, guy, candidates, conservative, said, lot, immigration\tabularnewline
\hline 
\multirow{2}{*}{31} & 0.038127 & 10 & tax, health, government, state, care, money, taxes, people, economy,
private, need, spending, federal, economic, business, pay, insurance,
increase, plan, market, jobs, income, program, read, americans, \tabularnewline
\cline{2-4} 
 & 0.019285 & 23 & health, tax, care, government, state, need, economy, people, private,
pay, taxes, dollars, insurance, money, business, reform, economic,
spending, federal, companies, plan, american, americans, increase,\tabularnewline
\hline 
\end{tabular}
\par\end{centering}

\protect\caption{MGPU-DP-SPDP topic quality with the duplicated dataset ``RedState
v.s DailyKos'' \label{tab:MGPU-DP-SPDP-Topic-Quality-dup}}
\end{table}

\FloatBarrier

\subsection{Hellinger Distance}

\begin{figure}
\begin{centering}
\includegraphics[width=1\textwidth]{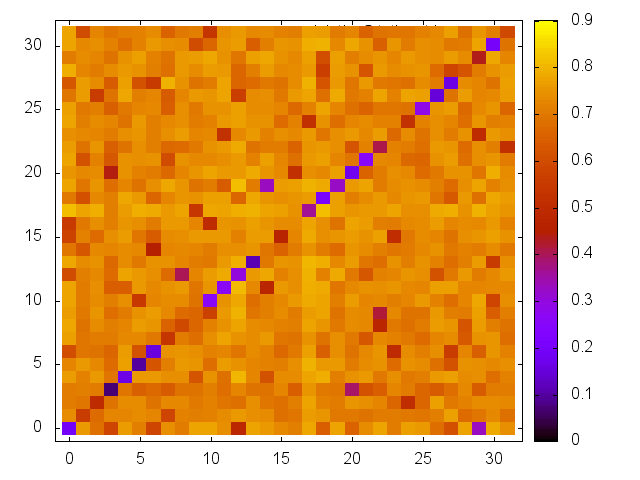}
\par\end{centering}

\protect\caption{Hellinger distance between the original SPDP and Improved-GPU-SPDP
with the dataset ``Reuters Disasters'' \label{fig:Hellinger-distance-RD-IvsS} }

\end{figure}

\begin{figure}
\begin{centering}
\includegraphics[width=1\textwidth]{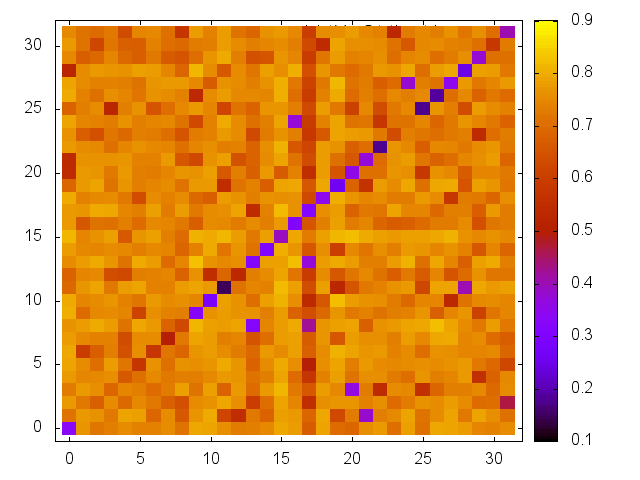}
\par\end{centering}

\protect\caption{Hellinger distance between the original SPDP and MGPU-DP-SPDP with
the dataset ``Reuters Disasters'' \label{fig:Hellinger-distance-RD-MvsS} }
\end{figure}

\paragraph*{\textmd{Hellinger distance \cite{nicta_5776} is a popular measure
of the similarity between two topic models. Here two heatmaps in Figure
\ref{fig:Hellinger-distance-RD-IvsS} and \ref{fig:Hellinger-distance-RD-MvsS}
are provided to show the Hellinger distance between the original SPDP
and Improved-GPU-SPDP, and the Hellinger distance between the original
SPDP and MGPU-DP-SPDP, both with dataset ``Reuters Disasters'' The
x-axis is the topics as appeared in Improved-GPU-SPDP, and y-axis
is the topics as appeared in the original SPDP, re-ordered to align
with the topics in x-axis. Lower values mean better match in topics.
The results shown here look almost as good as comparing two runs of
the original SPDP.}}

\chapter{Conclusion}

\section{Conclusions and Key Contributions}

\paragraph*{\textmd{I transformed the state-of-art algorithm SPDP into a distributed
parallel approximation, where the running speed is substantially improved
on a single GPU when only one GPU is used, and sublinearly scalable
to number of GPUs available when multiple GPUs are used.}}
\begin{itemize}
\item The single GPU algorithm improved the speed of the differential topic
modelling algorithm SPDP by about \textit{50 times on a single, cheap,
medium range laptop GPU}.
\item The multi-GPU algorithm MGPU-DP-SPDP leverages the latest modern multi-GPU
architecture, improves the running speed of SPDP and topic modelling
on a homemade small GPU cluster, with very small sacrifice in topic
quality, and only slightly worse or fairly comparable perplexity. 
\item SPDP is a representative of perhaps another hundred extensions of
LDA. My algorithm is designed in a generalized way, leaving mathematical
derivations of SPDP intact. With little modification, the algorithm
can be applied to other LDA extensions.
\end{itemize}

\section{Future Work}

\paragraph*{\textmd{Although the experiments have been quite successful in my
experiments, there are a few other improvements that could be done:}}
\begin{itemize}
\item automated optimization tailored to GPU specifications and parameters;
\item scalability and robustness of multi-GPU sampling;
\item full SPDP parallelization with non-identity transformation matrices;
\item large scale implementation;
\item adapting the three-level distributed parallel framework to other LDA
extensions.
\end{itemize}

\subsection{Automated Optimization}

\paragraph*{\textmd{The OpenCL framework provides APIs for users to get information
about the capacity and specification of available devices. It is possible
to determine the preferred local workgroup size from these information
and tailor my algorithm to maximize the performance. The number of
threads per workgroup should be made as close to the preferred size
as possible. Each thread should be allowed to compute multiple sample
probabilities, and the value should be determined by $\lceil\frac{N_{words}}{N_{threads}}\frac{K(S+1)}{T_{preferred}}\rceil$,
where $N_{words}$ is the number of words that can be sampled in parallel
with at most $C$ conflicts, $N_{threads}$ is the maximum number
of parallel hardware threads the device supports, $K$ is number of
topics, $S$ is number of word associations ($S=1$ if transformation
matrix $P$ is identity), $T_{preferred}$ is the preferred local
workgroup size, $C$ is a constant value to be determined. The formula
simply states that we should sample as many as possible words per
wave in parallel with at most $C$ conflicts. A conflict is defined
as multiple access to one count variable, which could happen when
two workgroups represent the same word and topic, or same word and
document, or same document and topic.}}

\paragraph*{\textmd{To have this idea implemented efficiently, kernel code must
be generated dynamically and complicated change needs to be done in
host code. With this properly implemented, we may expect a significant
speed improvement to the algorithm due to better occupancy hence more
efficient use of GPU device.}}

\subsection{Scalability And Robustness of Multi-GPU Sampling}

\paragraph*{\textmd{The results of the experiments imply that the current error
correction mechanism works well with word probabilities, but not as
well with topic probabilities, hence increasing the perplexity. A
better error correction mechanism is required for multi-GPU sampling.
In addition, sending and reading data from multiple GPUs can be inefficient
especially when the number of GPUs is large. It is possible to leave
data on the GPUs, perform the error correction, random number generation,
and count regeneration on GPUs. Complicated issues may arise, such
as looking for an efficient way to generate random numbers and propagate
to all GPU devices, and how not to make error while correcting error
in parallel. We may also consider using the model introduced by Smola
et al.$\ $in \cite{journals/pvldb/SmolaN10}, having a dedicated
GPU to update counts for all devices. However, the implementation
of this model would be very complicated, and may require approximation
as some features supported by the CPU and distributed networks are
not supported by the GPU and OpenCL. }}

\subsection{Full SPDP Parallelization With Non-identity Transformation Matrices}

\paragraph*{\textmd{Chen has showed in \cite{chen2012spdp} that appropriate
transformation matrices could uniformly improve the perplexity and
the topic quality. For simplicity we implemented MGPU-DP-SPDP in Algorithm
\ref{alg:Multi-GPU-SPDP-Host} and \ref{alg:Multi-GPU-SPDP-Kernel}
assuming transformation matrices $P^{i}$ are identity matrices. As
mentioned in Section \ref{sub:Parallelization-Over-Sampling}, the
difficulty in dealing with non-identity transformation matrices is
in implementing a linked list $v$ that can be modified in parallel
with high performance. One easy solution is to store the modifications
in a separate array, which is to be merged into linked list $v$ after
the GPU kernel finishes execution. However, because a massive number
of words are scheduled to be sampled on the GPU, this may causes substantial
delay in updating the linked list variable $v$ and the count variable
$q$, which may cost the algorithm more than the benefit of non-identity
transformation matrices could bring. Another approach is to allocate
a fixed amount of global memory and local memory for each workgroup
to store the linked list corresponding to the words they are assigned
to, and merge the linked lists at the start of execution of each workgroup
into the local memory. This allows any update to $v$ and $q$ to
be reflected immediately, but consequently the algorithm may be slowed
down substantially due to significantly increased amount of global
memory access.}}

\subsection{Large Scale Implementation}

\paragraph*{\textmd{With a GPU cluster supercomputer, my algorithm has the potential
to improve the running speed of SPDP and topic modelling to an unparallel
level of magnitude, potentially sublinear to number of GPUs times
number of cores available on a single GPU. Take this medium sized
GPU cluster supercomputer \cite{CSIROSuperComputer} in CSIRO for
example, a GPU cluster with 64 Tesla S2050s containing 256 GPUs with
114688 streaming processors. Assuming each GPU can only improve the
running speed of SPDP by 100 times (which is a highly conservative
under-estimate, consider these GPUs are all the highest range professional
scientific computing GPUs, and even my cheap laptop GPU is powerful
enough to improve the running speed of SPDP by }\textmd{\textit{54
times before optimization}}\textmd{). Based on the evidence in my
experiments it is reasonable to estimate that my algorithm running
on this GPU cluster could speed up SPDP by}\textmd{\textit{ thousands
of times before optimization,}}\textmd{ if the document collection
is large enough and a few issues faced in my experiments are solved
through better implmentation. This value is much greater than the
speedup achieved by Smola et al$.\ $did for LDA\cite{journals/pvldb/SmolaN10},
which is the best speedup known so far to LDA. Of course, this is
only an estimate, it may cause the algorithm to produce more error
when the algorithm is running at a large scale, and it may take large
amount of effort to make my algorithm scale to the level similar to
CSIRO GPU cluster, since there could be many synchronization and memory
issues. }}

\subsection{Adapting the Three-level Distributed Parallel Framework to Other
LDA Extensions}

\paragraph*{\textmd{Although my three-level distributed parallel framework is
proposed and implemented for SPDP, it does not use many specific properties
of SPDP. In particular, the mathematical derivation of SPDP is not
modified. This enables the algorithm to be generalized and adapted
to other LDA extensions, or LDA itself. It requires few modification
to the framework, and a small amount of work for implementing error
correction and other related parts. The framework is especially beneficial
to those extensions with complex structures with many variables. Furthermore,
the same approach may be taken to parallelize and approximate Markov
chains and solve a wider range of problems. }}

\paragraph*{\bibliographystyle{apalike2}
\bibliography{IJ01}
}
\end{document}